\documentclass[9pt,5p,twocolumn,authoryear]{elsarticle}

\usepackage[hidelinks,colorlinks,linkcolor=blue]{hyperref}
\usepackage{amssymb}
\usepackage{amsmath}
\usepackage{algorithm}
\usepackage{algorithmicx}
\usepackage{algpseudocode}
\usepackage{bookmark}
\usepackage{makecell}
\usepackage{multirow}
\usepackage{tabularx}
\usepackage{array}
\usepackage{orcidlink}

\journal{Engineering Applications of Artificial Intelligence}

\newcolumntype{L}{>{\raggedright\arraybackslash}X}

\begin{document}

\begin{frontmatter}

\title{Visual Swarm Navigation via Deep Reinforcement Learning and Evolutionary Hybrid Design}

\author[a]{\'{A}lvaro D\'{i}ez\corref{cor1}\,\orcidlink{0009-0004-6036-7733}}
\ead{alvaro.diez@ua.es}

\author[a]{Fidel Aznar\,\orcidlink{0000-0003-4521-956X}}
\ead{fidel@ua.es}

\cortext[cor1]{Corresponding author}

\affiliation[a]{organization={Department of Computer Science and Artificial Intelligence, University of Alicante},
	addressline={Carretera San Vicente del Raspeig},
	city={San Vicente del Raspeig},
	postcode={03690},
	state={Alicante},
	country={Spain}}

\begin{abstract}
Swarm robotics presents a robust and cost-effective paradigm for advanced automation in complex, dynamic environments, such as those encountered in search and rescue or environmental monitoring. A fundamental challenge for this field is the data-driven design of decentralized controllers capable of generating emergent collective behaviors. This paper proposes a novel, Artificial Intelligence (AI)-driven hybrid methodology for the automatic synthesis of swarm robotic controllers for autonomous visual navigation. This approach synergistically combines multi-agent reinforcement learning with neuro-evolutionary strategies, specifically leveraging implementations of the cross-entropy method and the covariance matrix adaptation evolution strategy to optimize a pre-trained individual navigation policy. The underlying deep architecture is engineered for low-cost, resource-constrained platforms, utilizing a compact neural network that relies exclusively on monocular camera imagery. This vision-based design emphasizes computational and energy efficiency, a critical requirement for practical swarm deployments. Experiments, performed in a high-fidelity physics simulator, demonstrate that the resulting controllers enable robust and scalable collective exploration of diverse indoor environments. The controller trained using our cross-entropy method achieves superior exploration coverage, visiting 36.20\% more regions compared to the covariance matrix adaptation evolution strategy. Critically, our best vision-based policy achieves exploration performance statistically comparable to traditional methods relying on more expensive distance sensors, while delivering a significant 31.40\% average reduction in energy consumption. These findings validate an effective and economically viable autonomous control system, establishing a path for deploying highly efficient collective intelligence in real-world engineering applications.   
\end{abstract}


\end{frontmatter}

\section{Introduction}\label{sec:introduction}

In recent decades, extensive research in robotics has focused on developing control methods for autonomous and intelligent agents. Roboticists have implemented strategies aimed at solving tasks traditionally performed by humans in complex environments \citep{Mukhlish_2018}, characterized by high uncertainty and dynamism. Multi-agent systems, particularly swarm robotics, have emerged as a promising and cost-effective paradigm for addressing these challenges. Drawing inspiration from the collective behaviors observed in social animals, swarm robotics aims to develop decentralized control methods for multiple simple, autonomous agents that rely on local sensing and communication.

This approach offers a valuable alternative to traditional strategies that depend on a single, sophisticated, and expensive robot, which may be inefficient or prone to complete failure in high-risk operational environments \citep{Mukhlish_2018}. For a swarm to function effectively, its members must self-organize through coordination and cooperation, enabling the emergence of robust, flexible, and scalable collective behaviors. This allows the system to maintain operational integrity despite agent loss \citep{Mukhlish_2018}, adapt to changing environmental configurations \citep{Brambilla_2013}, and perform diverse tasks effectively regardless of the swarm's size \citep{Cheraghi_2021}.

The design of controllers that facilitate such emergent behaviors is a significant challenge. While traditional behavior-based design methods have been employed, they are often labor-intensive and result in solutions tailored to specific problems in static environments \citep{Mukhlish_2018}. This approach involves manually modeling, implementing, analyzing, and refining individual robot behaviors through iterative trial and error until satisfactory collective behavior emerges \citep{Brambilla_2013}. Consequently, automatic design methods, primarily based on Reinforcement Learning (RL) \citep{Kaelbling_1996} and Evolutionary Computation (EC) \citep{Back_1996}, are strongly advocated for their ability to generate more adaptive solutions without explicit modeling of the problem or the desired collective behaviors \citep{Mukhlish_2018}.

Recent advancements in Artificial Intelligence (AI) have introduced powerful, large-scale generative models that are setting new benchmarks in robotics. Among the most prominent are transformers, such as Generative Pre-trained Transformer (GPT), which leverage attention mechanisms to process complex data sequences \citep{Feuerriegel_2023}. However, this power comes at a significant cost. These models are exceptionally demanding, requiring massive datasets for training and consisting of an enormous number of parameters. Their complexity in terms of memory, processing, and energy consumption makes them unsuitable for the low-cost, resource-constrained platforms typically used in swarm robotics. These limitations underscore the need for more efficient and pragmatic architectures specifically designed for the constraints of swarm robotics.

This paper charts a different course, diverging from the trend of developing computationally expensive models. We propose a hybrid automatic design method for swarm robotics that uses Evolution Strategies (ES) to address Deep Reinforcement Learning (DRL) \citep{Ha_2022} problems in autonomous navigation. Compared to traditional DRL algorithms, evolution strategies are highly parallelizable, scalable, and demonstrate greater robustness against the common challenges of temporal credit assignment and solution space exploration \citep{Salimans_2017}, making them particularly suitable for training multi-agent systems.

Our approach is built on an efficient deep architecture, demonstrating that robust, flexible, and scalable collective navigation is achievable without massive computational overhead, an indispensable requirement for the deployment of swarm systems in the real world. This method is specifically engineered to learn collective visual navigation policies, enabling homogeneous swarm agents to perform collision-free exploration in uncertain and dynamic environments using only low-cost monocular cameras \citep{Kolar_2020}.

While recent centralized visual-inertial Simultaneous Localization and Mapping (SLAM) systems have achieved remarkable mapping richness and localization accuracy through cloud-based multi-robot feature fusion \citep{LiuX_2025}, such advanced paradigms inherently require substantial computational and communicational bandwidth. For swarm applications prioritizing hardware simplicity, our decentralized, map-less \citep{Guzel_2013} strategy offers a complementary approach that significantly reduces implementation costs and energy requirements, making it highly viable for resource-constrained scenarios.

To validate our approach, this work addresses five fundamental research questions:

\begin{enumerate}[1.]
	
	\item How effectively can a hybrid automatic design method combining deep reinforcement learning with evolution strategies generate functional swarm robotic controllers for autonomous visual navigation?
	
	\item What level of stability do these automatically generated controllers maintain when subjected to sensory perturbations?
	
	\item How do the controllers respond to unforeseen variations in environmental conditions not encountered during training?
	
	\item What impact does varying the size of the robotic swarm have on the efficacy of the developed controllers?
	
	\item Can vision-based swarm controllers achieve performance comparable to traditional methods that employ more expensive and heavier distance sensors?
	
\end{enumerate}

The remainder of this article is organized as follows. Section \ref{sec:literature} reviews the most relevant research directions in the automatic design of swarm robotic controllers. Section \ref{sec:problem} presents the formal specification of our deep reinforcement learning problem. Section \ref{sec:methodology} details our training methodology for learning collective visual navigation behaviors, including two specific implementations of different evolution strategies. Section \ref{sec:experiments} compares and discusses the experimental results achieved by our two swarm controllers. Section \ref{sec:analysis} evaluates and analyzes the robustness, flexibility, and scalability of our controllers. Finally, Section \ref{sec:conclusions} synthesizes our conclusions, addresses the limitations of this work, and outlines future research directions.

\section{Related Work}\label{sec:literature}

The automatic design of swarm robotic controllers is a pivotal research area dedicated to generating the robust, flexible, and scalable collective behaviors required for real-world applications. This paradigm seeks to automate the complex process of engineering emergent behaviors, significantly reducing the manual effort traditionally required by designers \citep{Brambilla_2013}. Our research contributes to this field by proposing a hybrid methodology that synergizes learning-based and evolution-based techniques. Specifically, we present a soft computing approach that employs Multi-Agent Reinforcement Learning (MARL) and neuro-evolution to produce swarm controllers for autonomous visual navigation applications.

Multi-agent reinforcement learning comprises techniques commonly used for the automatic, learning-based design of swarm controllers. This paradigm can be conceptualized as a reward optimization process, where a specific function grants rewards to individuals within a multi-agent system to guide their learning \citep{Kuckling_2023}. We explore this variant of reinforcement learning in greater detail in Subsection \ref{subsec:learning}.

In parallel, neuro-evolution represents one of the earliest and most studied techniques for automatic controller design in swarm robotics. This approach involves optimizing the parameters, and potentially the topology, of one or more artificial neural networks with respect to a given objective function \citep{Kuckling_2023}. We examine this evolutionary computation approach in Subsection \ref{subsec:evolution}.

\subsection{Multi-Agent Reinforcement Learning}\label{subsec:learning}

Multi-Agent reinforcement learning extends traditional RL principles to systems comprising multiple interacting individuals \citep{Mukhlish_2018}. Whereas a conventional reinforcement learning agent learns behavior through trial-and-error interactions with its environment \citep{Brambilla_2013}, the multi-agent variant involves agents that interact with both the environment and each other \citep{Mukhlish_2018}. In recent years, deep reinforcement learning has become a prominent technique for addressing collective autonomous navigation problems in swarm robotics.

\citet{Huttenrauch_2019} proposed a MARL approach based on the Trust Region Policy Optimization (TRPO) training algorithm \citep{Schulman_2015}. Their method was validated in two classic multi-robot scenarios: a rendezvous task requiring agents to gather at an arbitrary location, and a pursuit-evasion task where a group of agents must cooperatively capture one or more targets. Similar to our system, their swarm was homogeneous; however, their agents relied on local communication to exchange task-relevant information. This is a key distinction from our approach, which employs a communication-free framework to enhance system simplicity and robustness.

More recently, \citet{Batra_2021} employed the popular Proximal Policy Optimization (PPO) algorithm \citep{Schulman_2017} to train decentralized control policies for drone swarms. Their MARL process, conducted in realistic simulated environments, successfully generated coordinated flocking behaviors that enabled the drones to maintain various formations while avoiding collisions with each other and with dynamic obstacles. Critically, the collective behaviors learned in simulation were successfully transferred to multiple small, real-world drones with limited computational resources, demonstrating the feasibility of deploying learned policies on hardware-constrained platforms, a central objective of our own research.

A study by \citet{Bloom_2022} is also of particular relevance for its development of a MARL method for collective object transport, one of the three principal navigation behaviors in swarm robotics alongside coordinated movement and collective exploration \citep{Brambilla_2013}. Consistent with our approach, their method utilized neural networks with few fully connected layers to learn collision-free navigation policies for a swarm of wheeled robots with differential drive \citep{Stefek_2020}. A key difference, however, is that their agents perceived the environment through distance sensors, whereas our work addresses the more challenging task of navigation using only monocular camera imagery.

Recent literature continues to expand the capabilities of PPO-based algorithms for complex spatial coordination. For instance, \citet{Bezerra_2025} proposed a decentralized framework that extends the Multi-Agent Proximal Policy Optimization (MAPPO) algorithm to address dynamic coalition formation in multi-robot task allocation. Although their primary focus lies in task assignment rather than explicitly in swarm robotic control, their framework intrinsically governs multi-robot spatial navigation, utilizing an end-to-end Convolutional Neural Network (CNN) \citep{LeCun_2015} based on a U-Net architecture to process local grid observations and plan collision-free paths. While this MAPPO-based approach is highly effective in dynamic environments, it relies on explicit intention sharing among agents and uses sophisticated convolutional architectures to process spatial action maps.

In a similar vein, other recent advancements have sought to enhance the adaptability of learning-based models by integrating them with classical rule-based algorithms. For example, the adaptive cooperative path planning method proposed by \citet{LiuH_2025} successfully utilized MARL to dynamically optimize rotational artificial potential fields, achieving efficient, collision-free navigation and effectively overcoming local minima in continuous action spaces.

Building upon these broad advances in multi-agent learning and hybrid optimization, our methodology explores an alternative paradigm: achieving robust collective navigation through a strictly communication-free approach, utilizing a lightweight, purely reactive perception and control architecture engineered specifically for severely resource-constrained platforms where heavy computational and communication overheads might be prohibitive.

The automatic design methods discussed in this subsection typically generate swarm behaviors through online reinforcement learning, which relies on numerous direct interactions between the agents and their environment. In contrast, deep architectures based on modern generative models, such as the generative pre-trained transformer, are often trained using offline reinforcement learning to mitigate the computational complexity of the online training phase. This offline paradigm utilizes data collected from a previously learned behavioral policy to find a new, optimal policy \citep{Boige_2023}. Notable examples of such architectures include Decision Transformers (DTs) and Control Transformers (CTs).

A decision transformer solves RL problems through the generative modeling of trajectories, which consist of sequences of states, actions, and rewards \citep{Meng_2023}, and can be adapted for multi-agent autonomous navigation applications. The control transformer, a variant of the former, seeks to improve generalization capabilities in tasks such as robotic navigation \citep{Lawson_2023}. However, both decision and control transformers currently feature architectures that demand significantly greater computational resources compared to the models typically used in automatic swarm controller design, and particularly in comparison to the low-cost solution proposed in this study.

\subsection{Neuro-Evolution}\label{subsec:evolution}

Neuro-evolution draws inspiration from Darwinian principles of natural selection and evolution, combining EC techniques with artificial neural networks to optimize network parameters \citep{Brambilla_2013}. In robotics, this approach treats neural networks as black-box controllers and generally requires no heuristic domain-specific information about the problem to be solved \citep{Kuckling_2023}. Neuro-evolution has proven highly effective for generating a wide array of swarm behaviors, including pattern formation, collective transport, task allocation, and foraging \citep{Mukhlish_2018}.

Within the broader field of evolutionary computation, our research focuses on evolution strategies, a family of black-box optimization algorithms that function as heuristic search processes inspired by natural evolution \citep{Salimans_2017}. For this study, we selected two prominent ES algorithms, the Cross-Entropy Method (CEM) \citep{Rubinstein_1999} and the Covariance Matrix Adaptation Evolution Strategy (CMA-ES) \citep{Hansen_2001}, valued for their straightforward implementation and minimal hyperparameter tuning requirements.

The cross-entropy method is a stochastic, adaptive optimization algorithm that can be classified as an evolution strategy due to its operational similarities with such techniques \citep{Nguyen_2021}. CEM has been applied to various single-agent \citep{Nguyen_2021} and multi-agent robotic control systems. For instance, \citet{Fan_2018} utilized this EC technique to evaluate a swarm of low-cost drones in two complex scenarios: a cooperative attack on a defended enemy base and a confrontation between two autonomous aerial teams. Similar to our methodology, they employed a neural network controller whose parameters were optimized using an ES-based method in a realistic simulation featuring agent interactions, varying initial conditions, and Gaussian noise \citep{Yogesh_2017}. However, their network contained three hidden layers, whereas our work demonstrates the efficacy of an even more compact single-layer model.

CMA-ES, one of today's most effective evolution strategies, is notable for its advanced capabilities in searching the solution space \citep{Hasselmann_2021}. Its effectiveness in designing swarm robotic controllers was recently validated by \citet{Hasselmann_2021}, whose work involved generating behaviors like aggregation and foraging and evaluating them in both simulation and on physical robots. Their comparative analysis confirmed the competitive performance of the covariance matrix adaptation evolution strategy against other neuro-evolutionary methods, including the Exponential Natural Evolution Strategy (xNES) \citep{Glasmachers_2010}, the Neuro-Evolution of Augmenting Topologies (NEAT) algorithm \citep{Stanley_2002}, and the basic EvoStick approach \citep{Francesca_2014}, reinforcing its suitability as a powerful tool for the automatic design of collective behaviors.

Building upon the foundational principles of algorithms like NEAT, which seek to optimize both the structural architecture and the synaptic weights of neural networks, recent advancements have continued to explore this dual optimization paradigm for decentralized control. For example, a recent study by \citet{Hiraga_2026} proposed a topology and weight neuro-evolutionary method designed to foster social learning within robotic swarms. By utilizing a mutation-based evolving artificial neural network, their approach allows robots to dynamically alter their controller's architecture through local interactions.

While evolving the topology offers theoretical flexibility and mitigates the need for manual network design, it inherently expands the search space and introduces significant computational overhead. In stark contrast, our methodology deliberately limits the evolutionary search space to the synaptic weights of an ultra-compact, fixed-topology controller. By coupling this static, highly efficient control policy with a deeply compressed representation of the visual input space, our approach bypasses the severe computational demands of topology mutation, guaranteeing rapid convergence and strict architectural efficiency for resource-constrained, vision-based navigation.

\section{Problem Specification}\label{sec:problem}

This section provides a formal specification for the challenge of learning swarm robotic controllers designed for autonomous navigation, framing it as a problem of deep reinforcement learning. To properly address the complexities of a multi-agent system, this problem is modeled not as a standard Markov Decision Process (MDP) \citep{Kaelbling_1996}, but as a Decentralized Partially Observable Markov Decision Process (Dec-POMDP), following similar approaches in the literature \citep{Amato_2013, Bloom_2022}. This more advanced framework is necessary due to the inherent limitations of the traditional MDP formulation in multi-agent scenarios, particularly its inability to adequately handle partial observability and the curse of dimensionality, which are defining challenges in the automatic design of swarm robotic controllers \citep{Bloom_2022, Kuckling_2023}.
	
We therefore define this process as the tuple $\langle \mathcal{E}, \mathcal{S}, \mathcal{O}, \mathcal{A}, \mathcal{P}, \mathcal{R} \rangle$. Within this formal structure, $\mathcal{E}$ represents the finite set of embodied agents comprising the swarm; $\mathcal{S}$ is the finite set of global system states; $\mathcal{O}$ constitutes the finite set of local observations available to each agent; $\mathcal{A}$ is the finite set of individual actions an agent can perform; $\mathcal{P}$ corresponds to the probabilistic state transition function governing the system's dynamics; and $\mathcal{R}$ is the reward function that guides the learning process. Each of these components is formally specified in the subsections that follow.

\subsection{Embodied Agents}\label{subsec:agents}

To facilitate the learning of effective collective navigation behaviors within a controlled and repeatable setting, we first generate a set of $J$ virtual agents, denoted as $\mathcal{E} = \{e_{j}\}_{j = 1}^{J}$. This set constitutes a multi-agent system, specifically designed to function as a swarm robotic system. Utilizing a proprietary high-fidelity physics simulator, each agent in the system is modeled as a low-cost differential drive vehicle, based on the physical specifications of a Makeblock mBot Ranger\footnote{https://www.makeblock.com/pages/mbot-ranger-robot-building-kit} robot of reduced dimensions.

Our virtual agents are considered embodied, a term signifying that they exist within a defined environment and interact with it through a physical body equipped with sensors for perception and actuators for movement. For perception, each agent is equipped with a single embedded monocular camera, which captures local observations of the environment in the form of Red, Green, Blue (RGB) images. For actuation, each agent executes navigation actions by independently controlling two motors, each driving one of the two wheels.

\subsection{State Space}\label{subsec:states}

In our reinforcement learning formulation, the state space, denoted as $\mathcal{S} = \{s_{t}\}_{t = 1}^{T}$, provides a complete characterization of the global state of the entire system at any given time instant $t$, up to a maximum episode duration $T$. This global state encompasses all relevant information about both the environment and the individual states of every agent within it. A critical feature of the Dec-POMDP framework is that individual agents do not have access to this complete state space; their knowledge is limited to their own local perceptions. Instead, the global state space $\mathcal{S}$ contains variables that are essential for managing the learning process and the simulation itself.

One of the most critical variables in our state space is the count of agent collisions at any given time, as it plays a fundamental role in defining the termination conditions for each learning episode. An episode is concluded either when the maximum time $T$ is elapsed or when the total number of agent collisions surpasses a predefined threshold $C$. Other variables vital for the simulation include the precise positions and orientations of all mobile agents and the locations of static obstacles within the environment.

The environment itself is a realistic digital twin of an indoor corridor at a real-world university, rendered using the same high-fidelity physics simulator employed for the agents. This synthetic environment is rich with realistic details, including random variations in lighting conditions, objects with diverse surface textures, and multiple static obstacles such as columns, walls, doors, and waste bins. Since this is a multi-agent reinforcement learning problem, the environment is inherently non-stationary \citep{Bloom_2022, Kuckling_2023}, as the agents themselves are considered dynamic obstacles from the perspective of their peers.

\subsection{Observation Space}\label{subsec:observations}

As established in Section \ref{sec:introduction}, each robot in the swarm operates with only local sensing capabilities and without any form of inter-agent communication. Within our Dec-POMDP formulation, this means that each mobile agent $e_{j}$ perceives only a partial observation $o_{t}^{j}$ of the global state $s_{t}$ at every time step $t$. Consequently, each agent operates based on its own individual subset of observations. The complete observation space for the system, $\mathcal{O}$, is therefore the union of all these individual subsets, which can be formally expressed as $\mathcal{O} = \{\mathcal{O}_{j}\}_{j = 1}^{J} = \{o_{t}^{j}\}_{t = 1}^{T}$. This decentralized approach to observation is crucial, as it effectively overcomes two major challenges inherent to applying conventional MDPs to multi-robot systems: the inability to model the partially observable nature of the environment from an individual agent's perspective \citep{Bloom_2022} and the curse of dimensionality \citep{Kuckling_2023}, which would cause the state space to grow intractably with the number of agents.

Discovering effective behaviors through multi-agent reinforcement learning typically requires a vast number of interactions between the agents and their environment, which renders the training phase computationally intensive \citep{Ye_2023}. To mitigate this challenge and optimize for both training and inference efficiency, our solution incorporates a Variational Autoencoder (VAE) \citep{Kingma_2013}. This deep neural network, denoted as $G$ and detailed in Fig. \ref{fig:microscopic}, is responsible for perception. Its primary function is to compress each high-dimensional local observation $o_{t}^{j}$ into a low-dimensional visual representation $z_{t}^{j}$. Specifically, the VAE's encoder, through the function $z_{t}^{j} = g(x_{t}^{j})$, transforms a preprocessed 64x64 pixel RGB image $x_{t}^{j}$ into a dense latent vector containing just 32 real-valued elements.

This aggressive compression is a cornerstone of our low-cost architectural design, as it enables the use of an exceptionally compact control module with very few trainable parameters. This design choice stands in stark contrast to the current trend of employing large-scale generative models like transformers, which would necessitate processing high-dimensional image data directly, thereby imposing a computational complexity orders of magnitude higher and making them unsuitable for resource-constrained swarm agents. In addition to compression, the autoencoder's decoder can also perform the inverse operation, reconstructing the input image $x_{t}^{j}$ from its latent vector $z_{t}^{j}$ via $x_{t}^{\prime j} = g^{\prime}(z_{t}^{j})$.

\subsection{Action Space}\label{subsec:actions}

Following the same decentralized principle applied to the observation space, the system's action space is partitioned into multiple subsets of individual actions $a_{t}^{j}$. Each action is executed independently by its respective agent $e_{j}$ at each time instant $t$ of an episode. This partitioning, formally expressed as $\mathcal{A} = \{\mathcal{A}_{j}\}_{j = 1}^{J} = \{a_{t}^{j}\}_{t = 1}^{T}$, is a crucial strategy to prevent the combinatorial explosion of the joint action space that would otherwise occur in a multi-agent system.

Given that our problem domain is autonomous robotic navigation, each action is selected from a discrete set of four fundamental navigation commands: move forward, turn right, turn left, and stop. Thus, any action $a_{t}^{j}$ performed by an agent at any given time must correspond to one of these commands, formally expressed as $a_{t}^{j} \in \{a_{forward}, a_{right}, a_{left}, a_{stop}\}$.

\subsection{Transition Function}\label{subsec:transitions}

In the context of multi-agent reinforcement learning, a transition from the current system state $s_{t}$ to the subsequent state $s_{t + 1}$ is driven by the collective actions of the individuals within the environment. In our system, each robot $e_{j}$ independently performs an action $a_{t}^{j}$ based on the compressed visual representation $z_{t}^{j}$ derived from its local observation $o_{t}^{j}$. This dynamic can be modeled by the probabilistic transition function $\mathcal{P} = (s_{t + 1} \vert s_{t}, a_{t}^{j})$, which maps the current state and an agent's action to a distribution over the next possible states. We implement this function using a Multi-Layer Perceptron (MLP) \citep{Pierson_2017}, denoted as $H$, which serves as the autonomous navigation controller. This neural network is tasked with automatically mapping the compressed perceptual inputs (\textit{i.e.}, $z_{t}^{j}$) to the appropriate control outputs (\textit{i.e.}, $a_{t}^{j}$).

\begin{figure*}[h!]
	\centering
	\includegraphics[height=150mm, width=\textwidth]{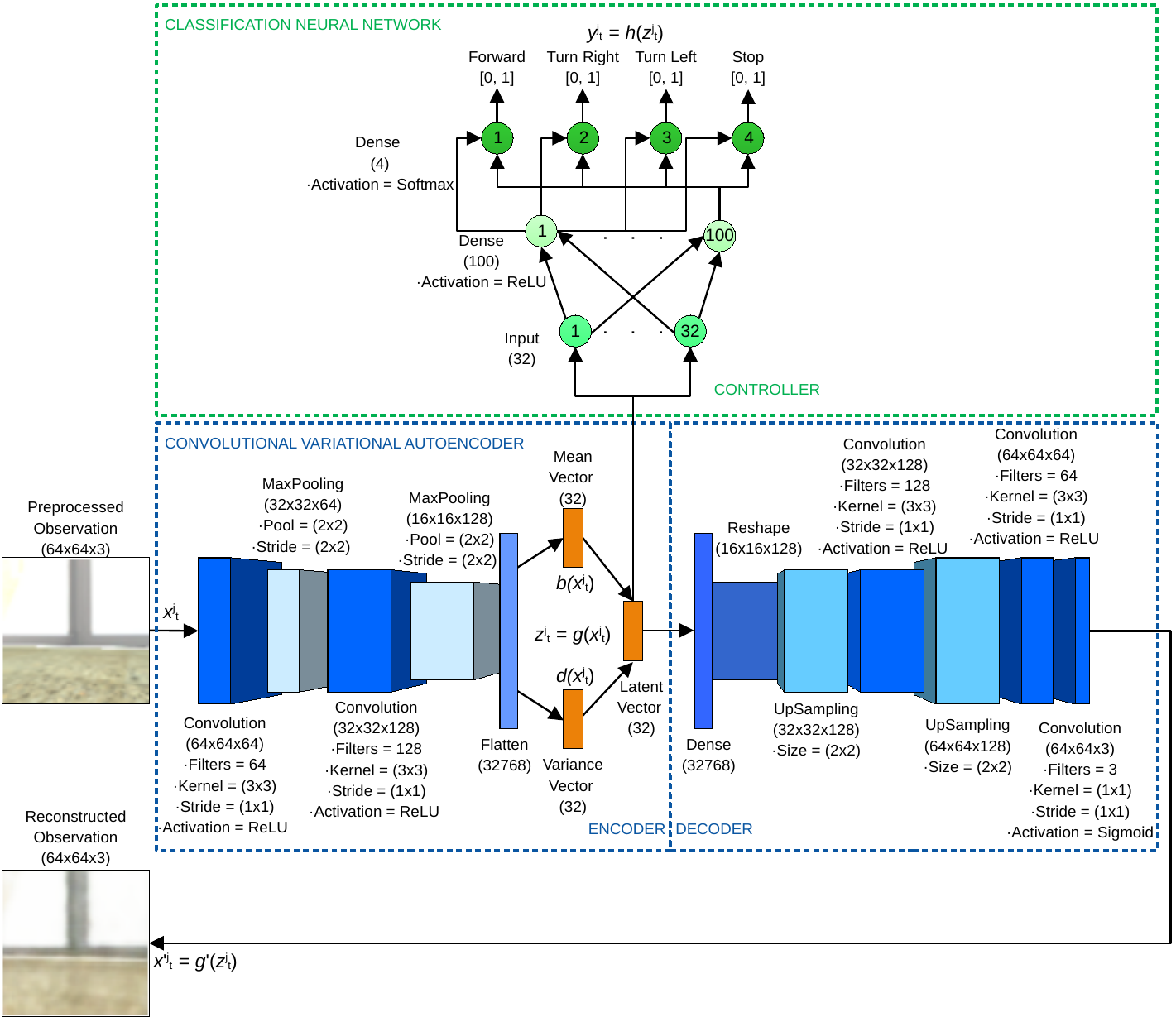}
	\caption{Microscopic model architecture. This figure details the layer structure of the convolutional variational autoencoder (bottom), which handles perception by compressing high-dimensional images, and the lightweight classification neural network (top), which serves as the controller for action selection.}\label{fig:microscopic}
\end{figure*}

The MLP network $H$ is the core component that we aim to optimize through our DRL process. As illustrated in Fig. \ref{fig:microscopic}, the controller receives the compact 32-element feature map $z_{t}^{j}$ generated by the VAE $G$. This efficient input representation allows the navigation controller to be exceptionally compact and lightweight. The design, which features a single hidden layer, results in a model with 3704 trainable parameters. This makes the online MARL process computationally viable, a stark contrast to the demands of larger architectures.

For context, recent research in reinforcement learning has utilized relatively small transformers for simpler control tasks, such as the popular Half Cheetah environment \citep{Wawrzynski_2009}, which uses a flat vector of only 18 real values as its observation space. Even in that less complex scenario, the transformer model comprised approximately 7 million parameters \citep{Boige_2023}. Attempting to apply such an architecture directly to our visual navigation problem, where observations are high-dimensional 64x64 RGB images and the system involves multiple interacting robots, would necessitate a vastly larger and more computationally intensive model, rendering an online training approach intractable. Our deliberate choice of a lightweight VAE-MLP architecture is therefore a pragmatic solution tailored to the specific constraints of swarm robotics.

\begin{figure*}[h!]
	\centering
	\includegraphics[scale=0.65]{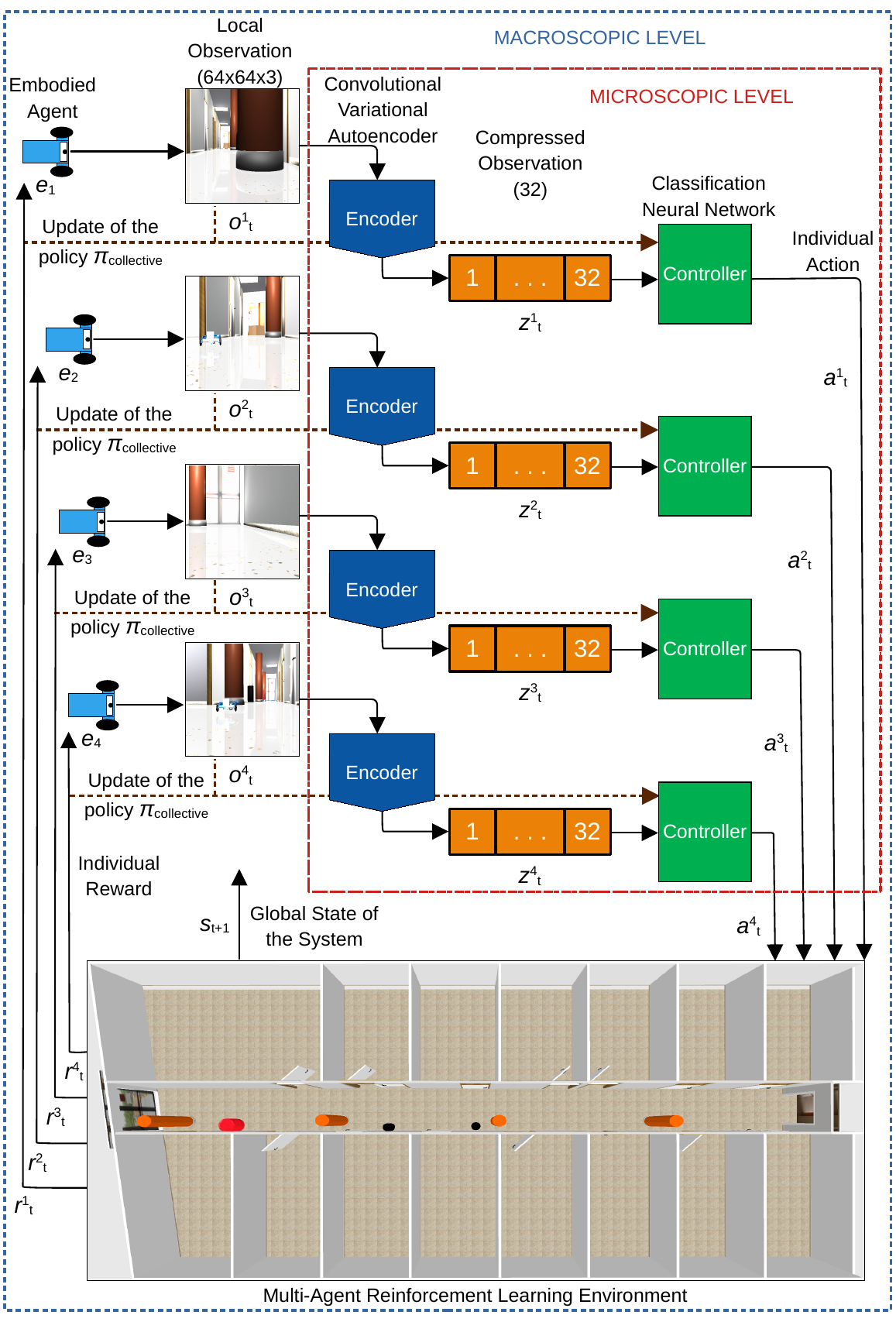}
	\caption{Relationship between the microscopic and macroscopic models of the proposed swarm robotic system. The parameters of the collective navigation policy are updated based on the individual rewards assigned to agents for actions they execute in response to their partial observations of the environment.}\label{fig:macroscopic}
\end{figure*}

The MLP's architecture includes an output layer that calculates the membership probabilities, $y_{t}^{j} = (y_{forward}, y_{right}, y_{left}, y_{stop})_{t}^{j}$, for each of the four possible navigation commands based on the input latent vector $z_{t}^{j}$. Consequently, the MLP $H$ functions as a classification neural network, with its operation summarized by the function $y_{t}^{j} = h(z_{t}^{j})$. This reactive navigation controller \citep{Xiao_2022}, in conjunction with the perceptual VAE model $G$, forms the microscopic model \citep{Brambilla_2013} of our swarm robotic system. The interplay between this microscopic decision-making level and the emergent macroscopic system behavior, denoted generically as $\pi_{collective}$, is illustrated in Fig. \ref{fig:macroscopic}.

\subsection{Reward Function}\label{subsec:rewards}

In MARL processes designed for swarm robotics, agents typically receive specific reward values at the individual level. These individual rewards are then aggregated to calculate a global reward, which serves as an evaluation of the swarm's overall collective performance. As noted by \citet{Kuckling_2023}, to be effective, learning in swarm robotics must occur at the individual level. Accordingly, our reward function $\mathcal{R}$ is structured to divide the global reward into distinct individual rewards, $r_{t}^{j}$, which are distributed to each agent $e_{j}$ based on the specific action $a_{t}^{j}$ it executed at time $t$. This approach directly addresses the spatial credit assignment problem, a common challenge in multi-agent learning and swarm robotics where it can be difficult to determine which agent's action contributed to a particular outcome \citep{Kuckling_2023}.

Our automatic design method is flexible and can accommodate various reward functions, allowing for subtle modifications to be made through manual credit assignment to promote different collective behaviors. The desired swarm behavior is contingent on the specific task the robots must learn to solve; for instance, aggregation behavior is crucial for target tracking, whereas dispersion is valuable for collective exploration. As the focus of this paper is on autonomous exploration, our goal is to learn an effective policy for map-less visual exploration of a dynamic environment, based on reactive detection and avoidance of obstacles. The reward function designed to elicit this policy is defined as follows:
\begin{equation}
	\mathcal{R} =\begin{cases}
		5, & \text{if } a_{t}^{j} = a_{forward},\\
		0.5, & \text{if } a_{t}^{j} = a_{right} \text{ or } a_{t}^{j} = a_{left},\\
		-1, & \text{if } a_{t}^{j} = a_{stop},\\
		-10, & \text{if } collision.\label{eqn:reward}
	\end{cases}
\end{equation}

It is crucial to clarify the semantic definition of\text{ }\textquotedblleft{exploration}\textquotedblright\text{ }within the context of our proposed task. The objective is not optimal, single-pass spatial coverage (\textit{e.g.}, area mapping), where revisiting a previously explored sector would be considered an algorithmic inefficiency. Instead, the learned collective policy is intended to operate fundamentally as an emergent, continuous wandering behavior. This behavior is driven by reactive open-space detection and dynamic obstacle avoidance. 

Because the swarm must autonomously explore without a predefined terminal destination, introducing a goal-oriented reward would inherently contradict the fundamental paradigm of this decentralized application. Consequently, the true effectiveness of the learned policies is not demonstrated by reaching a specific target, but rather by maximizing long-term spatial coverage and overall system survivability.

The specific coefficients and relative weights of our reward function were derived through an empirical tuning process during a preliminary pilot phase. In this phase, alternative reward formulations were systematically evaluated to assess the sensitivity of the policy. We observed that the MARL process is highly sensitive to the balance between collision and stop penalties. For instance, relaxing the collision penalty reliably resulted in reckless exploration. Conversely, minimizing the stop penalty induced overly stagnant behaviors, as agents preferred to remain stationary rather than risk navigating unfamiliar areas.

Because these alternative formulations failed to meet the minimum viability criteria for the emergence of effective collision-free collective behaviors, they were not advanced to the full, computationally intensive evolutionary training and evaluation phases. Furthermore, turning actions are rewarded to a significantly lesser extent than forward advancement to prevent the emergence of unintended policies. Although turning commands are vital for collision avoidance, our preliminary studies demonstrated that rewarding them excessively causes the robots to rotate continuously on their own axis, leading to severe oscillatory behaviors and drastically reducing the spatial regions visited by the swarm.

The slight penalty for stopping serves to break stagnation and encourage continuous movement, while still providing a viable emergency action to avoid the catastrophic collision penalty. The presented configuration encapsulates the optimal heuristic balance found to successfully solve the exploration-exploitation trade-off, facilitating a robust policy that, as demonstrated in our experiments, generalizes effectively across different swarm sizes and environmental conditions.

To conclude the formal problem specification, Table \ref{tab:problem} consolidates not only the defined reward structure but all foundational components comprising our Dec-POMDP framework. By systematically summarizing the state, observation, and action spaces alongside the transition dynamics, this explicit delineation establishes the rigorous mathematical baseline required before transitioning to the evolutionary training methodology in the subsequent section.

\begin{table}[h!]
	\begin{minipage}{\columnwidth}
		\begin{center}
			\caption{Systematic summary of the Dec-POMDP formulation for our swarm exploration task.}\label{tab:problem}
			\begin{tabularx}{\columnwidth}{@{}ccL@{}}
				\hline
				\textbf{Component} & \textbf{Definition} & \textbf{Implementation}\\
				\hline
				\hline
				Agents & $\mathcal{E}$ & Homogeneous swarm of $J$ robots operating under a fully decentralized, map-less paradigm.\\
				States & $\mathcal{S}$ & Unobservable global variables (\textit{e.g.}, time steps $T$ per episode and collision termination threshold $C$).\\
				Observations & $\mathcal{O}$ & Local partial images: 64x64 RGB camera feed per agent, processed into a 32-element latent vector by the VAE $G$.\\
				Actions & $\mathcal{A}$ & Discrete reactive navigation space: $a_{t}^{j} \in \{a_{forward}, a_{right}, a_{left}, a_{stop}\}$.\\
				Transition & $\mathcal{P}$ & Mapping of actions $a_{t}^{j}$ and system states $s_{t}$ to new states $s_{t + 1}$ through the MLP $H$.\\
				Reward & $\mathcal{R}$ & Emergent heuristic scalar evaluating the transition: move forward ($+5$), turn ($+0.5$), stop ($-1$), and collision ($-10$).\\
				\hline
			\end{tabularx}
		\end{center}
	\end{minipage}
\end{table}

\section{Training Methodology}\label{sec:methodology}

This section details the two-tiered training methodology employed to develop the swarm robotic system, addressing both the microscopic and macroscopic levels. First, in Subsection \ref{subsec:individual}, we describe the process for learning a robust individual navigation policy, which forms the foundation for each agent's autonomous capabilities. This microscopic model, comprising a perceptual module and a control module, is trained using established machine learning paradigms: unsupervised and supervised learning \citep{Shanthamallu_2017}.

Our approach, similar to that of \citet{Ha_2018}, synergizes reinforcement learning and evolution strategies, utilizing a multi-agent reinforcement learning reward function as an objective function for a neuro-evolutionary algorithm. Additionally, we leverage the pre-trained individual policy as a starting point for yielding emergent swarm behaviors. This critical transition from isolated navigation to the evolutionary optimization of emergent collective behaviors is formalized and comprehensively detailed in Subsection \ref{subsec:collective}.

\subsection{Individual Navigation Policy}\label{subsec:individual}

The initial phase focuses on training the microscopic model, which consists of the perception and control modules responsible for each individual agent's reactive navigation. This process yields a foundational policy, $\pi_{individual}$, capable of collision-free navigation based on visual input.

\subsubsection{Convolutional Variational Autoencoder}\label{subsubsec:autoencoder}

The perceptual module of our architecture is a convolutional variational autoencoder, denoted as $G$. We train this model through a combination of unsupervised learning and Bayesian variational inference \citep{Kingma_2013}. This paradigm was deliberately chosen for its capacity to learn a smooth, well-structured latent space that captures salient environmental features while promoting generalization. The probabilistic nature of the VAE allows for an explicit representation of uncertainty, yielding more robust visual features that can adapt to the variations and unpredictability of real-world environments.

The training objective is twofold: to learn an encoder function, $g$, that compresses high-dimensional observations, $\{x_{i}\}_{i = 1}^{M}$, into low-dimensional latent vectors, $\{z_{i}\}_{i = 1}^{M}$, and a decoder function, $g^{\prime}$, that reconstructs the original images from these latent vectors. To achieve this, we minimize a specialized loss function based on \citep{Bank_2023}:
\begin{equation}
	\mathbb{L}_{G} = \sqrt{\frac{1}{M}\sum\limits_{i = 1}^{M}(x_{i} - g^{\prime}(z_{i}))^{2}} + KL,\label{eqn:perception}
\end{equation}
where the Kullback-Leibler (KL) divergence \citep{Bank_2023} term is defined as:
\begin{equation}
	KL = \text{divergence}(N(b(x_{i}), d(x_{i})), N(0, I)).\label{eqn:divergence}
\end{equation}

The loss function $\mathbb{L}_{G}$ consists of two critical components. The first term is the reconstruction loss, quantified as the root mean squared error between the original images $x_{i}$ and the decoded images $g^{\prime}(z_{i})$. The second term, the Kullback-Leibler divergence, acts as a regularization mechanism. It measures the discrepancy between the distribution learned by the encoder, parameterized by mean $b(x_{i})$ and variance $d(x_{i})$ vectors, and a standard normal distribution $N(0, I)$. By penalizing deviations from this prior, the KL divergence encourages the model to learn a continuous and organized latent space, which is instrumental in preventing overfitting and enhancing the generalization capabilities of the learned policy. The training dataset $X$ consists of 45225 synthetic 64x64 pixel RGB images (\textit{i.e.}, $M = 45225$), acquired from the monocular camera of a single virtual agent within our simulated corridor.

The selection of a variational autoencoder over alternative lightweight feature extractors, such as standard deterministic convolutional neural networks, is fundamentally driven by the specific mathematical and structural requirements of neuro-evolutionary control tasks. While a standard CNN provides efficient dimensionality reduction, its deterministic mapping often yields a highly non-linear or fragmented latent space. In contrast, the probabilistic nature of the VAE constructs a continuous, regularized, and smooth latent manifold. This probabilistic representation offers a crucial advantage for the subsequent evolutionary optimization phase: a smooth state space ensures that the stochastic weight mutations introduced by our evolution strategies into the controller translate into predictable, continuous variations in the agent's action space. Consequently, this mitigates the risk of catastrophic behavioral jumps and instability that frequently occur when evolving control policies over irregular deterministic representations.

Additionally, the characteristic lack of sharpness in VAE outputs provides a substantial advantage for embodied navigation by acting as an intrinsic regularization mechanism. The mapping of complex, high-dimensional visual data into a compact 32-element latent representation inherently discards high-frequency, domain-specific noise, like detailed surface textures. At the same time, the process guarantees the retention of critical low-frequency structural cues, such as corridor layouts and obstacle boundaries. Ultimately, prioritizing these invariant geometric features over superficial visual details plays a crucial role in bridging the sim-to-real domain gap.

\subsubsection{Classification Neural Network}\label{subsubsec:classifier}
 
The control module, a lightweight multi-layer perceptron denoted as $H$, is trained via supervised learning to form the individual reactive navigation policy, $\pi_{individual}$. The training process utilizes the compact 32-element latent vectors $\{z_{i}\}_{i = 1}^{M}$ generated by the VAE's pre-trained encoder as inputs. The corresponding ground-truth labels, $L = \{l_{i}\}_{i = 1}^{M}$, are one-hot encoded vectors \citep{Dahouda_2021} representing one of the four discrete navigation commands: move forward, turn right, turn left, or stop.

The objective of this training stage is to learn the function $y_{i} = h(z_{i})$, which maps a given latent representation $z_{i}$ to a probability distribution over the possible actions, $y_{i} = (y_{forward}, y_{right}, y_{left}, y_{stop})_{i}$. To this end, we minimize the categorical cross-entropy loss function \citep{Dufourq_2017}, defined as:
\begin{align}
	\mathbb{L}_{H} &= -\sum\limits_{i = 1}^{M}(l_{forward}, l_{right}, l_{left}, l_{stop})_{i} \ln h(z_{i}),\label{eqn:control}\\
	h(z_{i}) &= (y_{forward}, y_{right}, y_{left}, y_{stop})_{i}.\nonumber
\end{align}

This loss function quantifies the dissimilarity between the predicted action probabilities and the ground-truth labels, guiding the network to learn an accurate mapping from perception to action. The efficiency of this stage is greatly enhanced by the VAE's dimensionality reduction, which enables the use of an exceptionally compact MLP. This design is not only crucial for rapid training but also a prerequisite for the computationally intensive evolutionary optimization phase that follows.

\subsection{Collective Navigation Policies}\label{subsec:collective}

This subsection details our hybrid method for automatically designing collective behaviors. We frame the challenge of learning a collective policy, $\pi_{collective}$, as a deep reinforcement learning problem. However, instead of employing traditional gradient-based DRL algorithms, which can be sensitive to hyperparameter tuning \citep{Nguyen_2021} and struggle with exploration in sparse-reward settings, we utilize evolution strategies.

To realize these decentralized behaviors, the choice of the optimization algorithm is critical. While gradient-based deep reinforcement learning algorithms, such as PPO or TRPO, are prevalent in continuous control, they present significant practical limitations in swarm robotics. Specifically, gradient-based MARL methods struggle with the temporal credit assignment problem when dealing with sparse rewards in reinforcement learning (\textit{e.g.}, delayed binary penalties upon collision in complex environments).

Recent literature has demonstrated highly effective solutions to this reward sparsity challenge; for example, employing transfer-learning-based meta-Q-learning to construct implicit advantage functions has proven remarkably robust for path planning under dynamic uncertainties \citep{Liu_2026}. While such sophisticated value-based frameworks excel in environments where rich state information can be leveraged, our architecture operates under extreme partial observability by compressing visual inputs into a 32-variable latent vector. In this strictly constrained context, gradient-based methods become highly susceptible to vanishing gradients and instability \citep{Nguyen_2021}.

In contrast, evolution strategies provide a robust alternative. Beyond their general parallelizability, ES methods offer distinct advantages for reinforcement learning: they are inherently invariant to delayed reward distributions, do not rely on backpropagation through time, and bypass the need for value function approximations \citep{Salimans_2017}, making them exceptionally stable in non-stationary multi-agent settings. However, this approach introduces a critical practical trade-off: ES algorithms are notoriously sample-inefficient when tasked with optimizing high-dimensional neural network architectures \citep{Nguyen_2021} directly from raw pixels.

To mitigate this architectural trade-off, our methodology draws inspiration from the generative representation techniques proposed by \citet{Ha_2018}. By integrating a pre-trained variational autoencoder into the macroscopic model, high-dimensional visual inputs are effectively compressed into a 32-dimensional latent bottleneck. Consequently, our individual navigation policy, $\pi_{individual}$, can be parameterized by an ultra-compact multi-layer perceptron. Within this drastically reduced parameter space, the traditional sample inefficiency of ES algorithms is bypassed. Neuro-evolutionary techniques such as CEM and CMA-ES thus become remarkably efficient at escaping local optima and discovering robust decentralized policies, outperforming gradient-based alternatives.

Our neuro-evolutionary approach uses fitness values calculated from individual evaluations in each training iteration $k$. We employ the MARL reward function from Eq. (\ref{eqn:reward}) to calculate the fitness values, introducing spatial and temporal knowledge of the environment into the objective function. Thus, we mitigate the problems of deception and bootstrapping \citep{Kuckling_2023}. These fitness values guide the optimization process of the individual reactive navigation behavior, previously learned by means of the compressed image classifier $H$. Its design offers two additional advantages: reduced evolutionary process search space dimensions and focused optimization on model parameters, specifically their weights. These advantages help mitigate the high dimensionality problem \citep{Brambilla_2013} and reduce training computational costs.

To clarify the explicit interaction between the perception and control modules during the evolutionary RL phase, it is vital to note that the parameters of our variational autoencoder are strictly frozen. The VAE operates as an inference-only feature extractor that continuously provides the state representation to the mobile agents. The reinforcement learning signal (\textit{i.e.}, the fitness score) derived from environment interactions is exclusively used to mutate and optimize the synaptic weights of the MLP controller. This explicit decoupling ensures that the stable, pre-learned visual representations are not catastrophically forgotten or disrupted by the high variance of the evolutionary exploration.

Within our hybrid framework, it is necessary to conceptually separate fitness evaluation from policy optimization. In our pipeline, the reinforcement learning paradigm is confined to the fitness evaluation phase: agents interact with the environment over episodes, accumulating rewards. Once an episode concludes, the RL loop yields a scalar fitness score. Conversely, policy optimization is entirely governed by the evolution strategies. The ES algorithms do not perform backpropagation or use temporal gradients derived from the RL interactions; instead, they treat the RL fitness scores as a black-box evaluation metric to rank, select, and mutate the underlying synaptic weights of the macroscopic model.

The transition from individual policy learning to collective policy evolution represents a critical juncture in our training pipeline. The success of the final emergent collective behavior is heavily dependent on the quality of the pre-trained individual policy, $\pi_{individual}$. Because learning robust obstacle avoidance from scratch in a multi-agent environment typically suffers from sparse reward signals and severe penalties for critical failures, starting the evolutionary process from randomly initialized neural network weights would consistently result in poor initial performance. This would yield a predominantly flat fitness landscape where the ES algorithms struggle to find a meaningful direction for improvement. Therefore, the individual policy provides the essential foundational competence (\textit{i.e.}, valuable prior knowledge about reactive static obstacle avoidance), acting as a mandatory stepping stone.

The operational principle of this transition relies on direct weight transfer to seed the evolutionary search distributions. The pre-trained weights of the individual MLP classifier $H$, learned via supervised learning in the first stage, are extracted and flattened into a single one-dimensional vector. This vector acts as the foundational blueprint for the multi-agent evolutionary loop by defining the central point of the initial probability distributions. By injecting a predefined initial variance or step size, the ES algorithms generate a starting population of candidate policies distributed precisely around this competent baseline.

The subsequent evolutionary process refines these baseline weights to account for dynamic obstacles (\textit{i.e.}, the other swarm agents), thus bridging the gap between isolated navigation and coordinated multi-agent exploration. For this purpose, we also fine-tune the variational autoencoder $G$ to improve perception and avoidance of dynamic obstacles. This involves large-scale image capture in our simulated environment using multiple mobile robots simultaneously. The acquired observations create an extensive training dataset for unsupervised autoencoder adjustment using the loss function from Eq. (\ref{eqn:perception}).

Our training protocol incorporates two key enhancements:

\begin{itemize}
	
	\item \textbf{Parameter Sharing}: All agents in the swarm share a single policy's parameters during a fitness evaluation (\textit{i.e.}, a MARL episode). This protocol efficiently optimizes collaborative learning in homogeneous robotic swarms while addressing the non-stationarity of the multi-agent environment \citep{Gupta_2017}.
	
	\item \textbf{Curriculum Learning}: We incrementally increase the maximum MARL episode duration, $T_{k}$, at each generation $k$, starting from an initial duration $T_{0}$ and up to a maximum number of generations $\kappa$:
	\begin{equation}
		T_{k} = T_{0} + k * \text{integer}(4500 / \kappa).\label{eqn:curriculum}
	\end{equation}

	This strategy accelerates the initial training iterations to encourage broad exploration of the solution space and prevents premature convergence. As the evolutionary process advances and candidate solutions improve, longer episodes allow for more effective exploitation of the most promising navigation controllers.
	
\end{itemize}

\subsubsection{Cross-Entropy Method}\label{subsubsec:cem}

Our first evolution strategy is a specialized implementation of the cross-entropy method, a type of Estimation of Distribution Algorithm (EDA). The core principle of this family of algorithms involves iteratively managing and updating a search distribution to sample potential solutions, which are then evaluated \citep{Wierstra_2014}. At each iteration $k$, the CEM algorithm samples a population of candidate controllers, $W_{k}$, from a multivariate Gaussian distribution $N(\mu_{k}, V_{k})$. In our neuro-evolutionary approach, each candidate solution, $w_{p}^{k}$, represents a complete set of weights for the MLP controller and is sampled around the mean vector $\mu_{k}$, which represents the mean solution of population $W_{k}$. The distribution is parameterized by both the mean vector and a diagonal covariance matrix $V_{k}$ \citep{Nguyen_2021}, computed using a variance vector $\sigma_{k}$.

\begin{figure}[h!]
	\centering
	\includegraphics[scale=0.65]{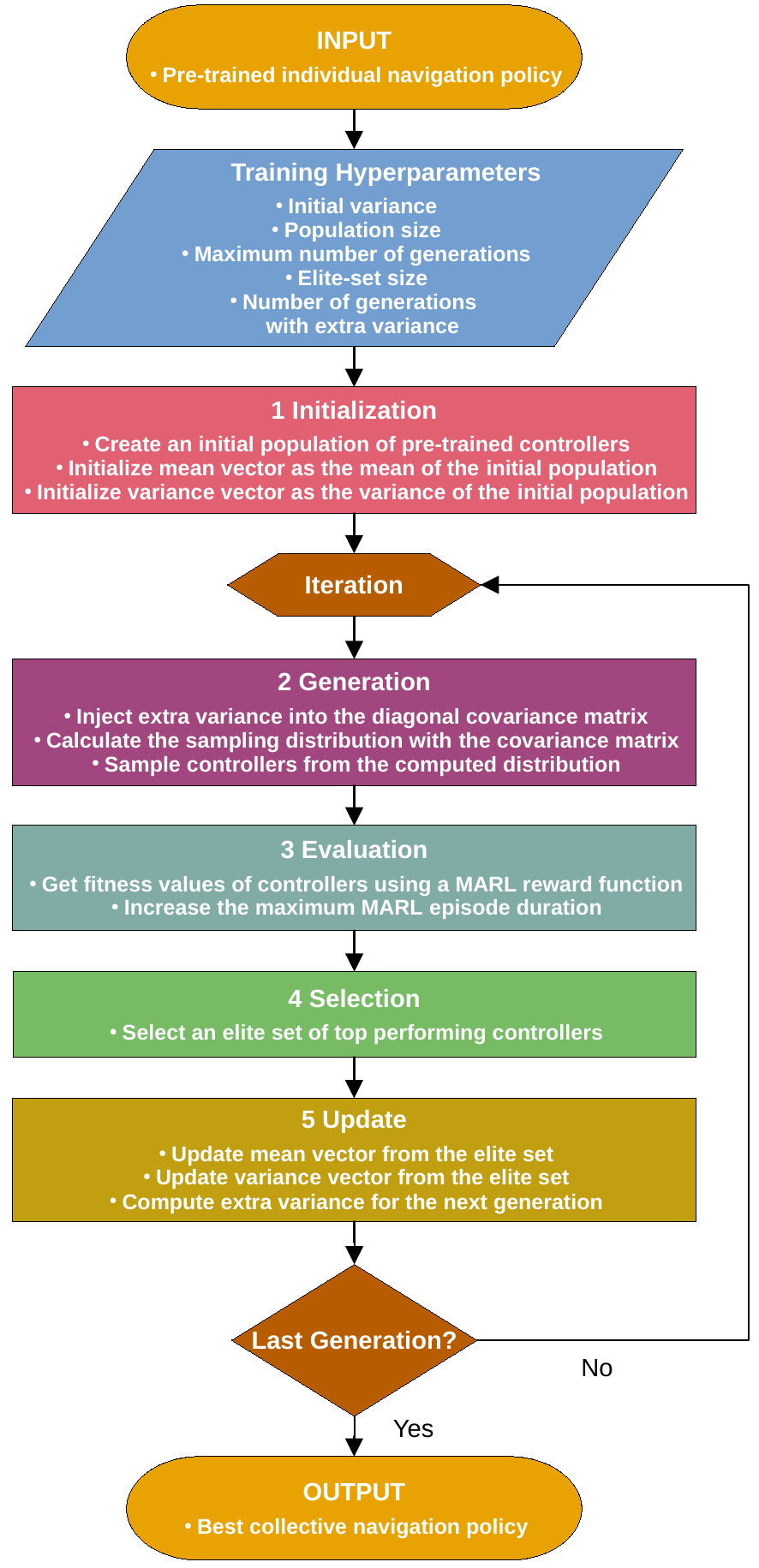}
	\caption{Flowchart of the proposed CEM algorithm.}\label{fig:cem}
\end{figure}

\begin{algorithm}[h!]
	\caption{Pseudo-code implementation of the proposed CEM algorithm.}\label{alg:cem}
	\begin{algorithmic}[1]
		
		\Require{
			\\
			Initial variance $\sigma_{0}$, population size $\rho$, maximum number of generations $\kappa$, individual navigation policy $\pi_{individual}$, elite-set size $\epsilon$, number of generations $\gamma$ with extra variance.	
		}
		\Ensure{
			\\
			Collective navigation policy $\pi_{collective}$.\\
		}
		
		\For{$p = 1$,...,$\rho$}
			\State Initialize an individual of $W_{0}$ with $w_{p}^{0} = \pi_{individual}$.
		\EndFor
		
		\State $k = 1$.
		\State Initialize mean vector $\mu_{k}$ as the mean of $W_{0}$.
		\State Initialize variance vector $\sigma_{k}$ as the standard deviation of $W_{0}$.
		
		\While{$k <= \kappa$}
		
			\For{$p = 1$,...,$\rho$}
				\State Create a new individual for $W_{k}$ with Eq. (\ref{eqn:cem}).
			\EndFor
			
			\For{$p = 1$,...,$\rho$}
				\State Get the fitness value of an individual of $W_{k}$ with Eq. (\ref{eqn:reward}).
			\EndFor
			
			\State Select $\epsilon$ individuals with the highest fitness from $W_{k}$.
			\State Update $\mu_{k + 1}$ according to the mean of the $\epsilon$ selected individuals.
			\State Update $\sigma_{k + 1}$ according to the standard deviation of the $\epsilon$ selected individuals.
			\State Compute extra variance with $\sigma_{k + 1}^{extra} = \max(1 - (k + 1 / \gamma), 0) * \sigma_{0}^{2}$.
			\State Update covariance matrix with $V_{k + 1} = diagonal(\sigma_{k + 1}^{2} + \sigma_{k + 1}^{extra})$.
			\State $k = k + 1$.
		
		\EndWhile    
		
	\end{algorithmic}
\end{algorithm}

As previously discussed, our EC approach begins with an initial solution, represented by population $W_{0}$. In the first iteration of our CEM algorithm, we initialize the mean vector $\mu_{k}$ and variance vector $\sigma_{k}$ as the mean and standard deviation of this initial population. In subsequent iterations, we update these vectors using an elite set comprising a fixed number $\epsilon$ of individuals. These elite individuals represent the candidate solutions from population $W_{k}$ that achieve the highest accumulated reward values during a multi-agent reinforcement learning episode. This elitist approach promotes the exploitation of promising solutions based on their fitness values.

The selection of elitist individuals is a technique employed by numerous researchers using the cross-entropy method, including \citet{Botev_2004} and \citet{Nguyen_2021}. Their approach introduces an additional enhancement to the basic CEM implementation by injecting supplementary variance into the sampling distribution during a predetermined number of iterations. At each generation $k$, we calculate the extra variance magnitude $\sigma_{k}^{extra}$ using two hyperparameters: the number $\gamma$ of iterations for variance injection and a specified initial variance $\sigma_{0}$. This additional variance is incorporated into the diagonal covariance matrix $V_{k}$ used to generate each individual in the population $W_{k}$ of candidate solutions. The generation of a new individual $w_{p}^{k}$ follows the following formulation:
\begin{align}
	w_{p}^{k} &= N(\mu_{k}, V_{k}),\label{eqn:cem}\\
	V_{k} &= \text{diagonal}(\sigma_{k}^{2} + \sigma_{k}^{extra}),\nonumber\\
	\sigma_{k}^{extra} &= \max(1 - (k / \gamma), 0) * \sigma_{0}^{2}.\nonumber
\end{align}

The hyperparameter $\gamma$ plays a crucial role, as injecting extra variance over more generations enables more thorough exploration of potential solutions. This mechanism functions as a form of mutation \citep{Back_1996}, allowing the training algorithm to produce more diverse individuals, particularly during the initial iterations of the evolutionary process. The complete workflow of our CEM implementation is illustrated in Fig. \ref{fig:cem} and detailed in Algorithm \ref{alg:cem}.

\subsubsection{Covariance Matrix Adaptation Evolution Strategy}\label{subsubsec:cma}

Our second ES algorithm is the covariance matrix adaptation evolution strategy, a powerful black-box optimization algorithm renowned for its performance in optimizing synaptic weights of neural networks with limited parameters. CMA-ES employs a sophisticated mechanism for adapting the search distribution. Its key advantage lies in its ability to dynamically adapt the covariance matrix of the search distribution \citep{Hansen_2001}, which enables efficient exploration in highly correlated search spaces. This characteristic offers significant advantages when optimizing synaptic weights for neural robotic controllers \citep{Yu_2015}.

\begin{figure}[h!]
	\centering
	\includegraphics[scale=0.65]{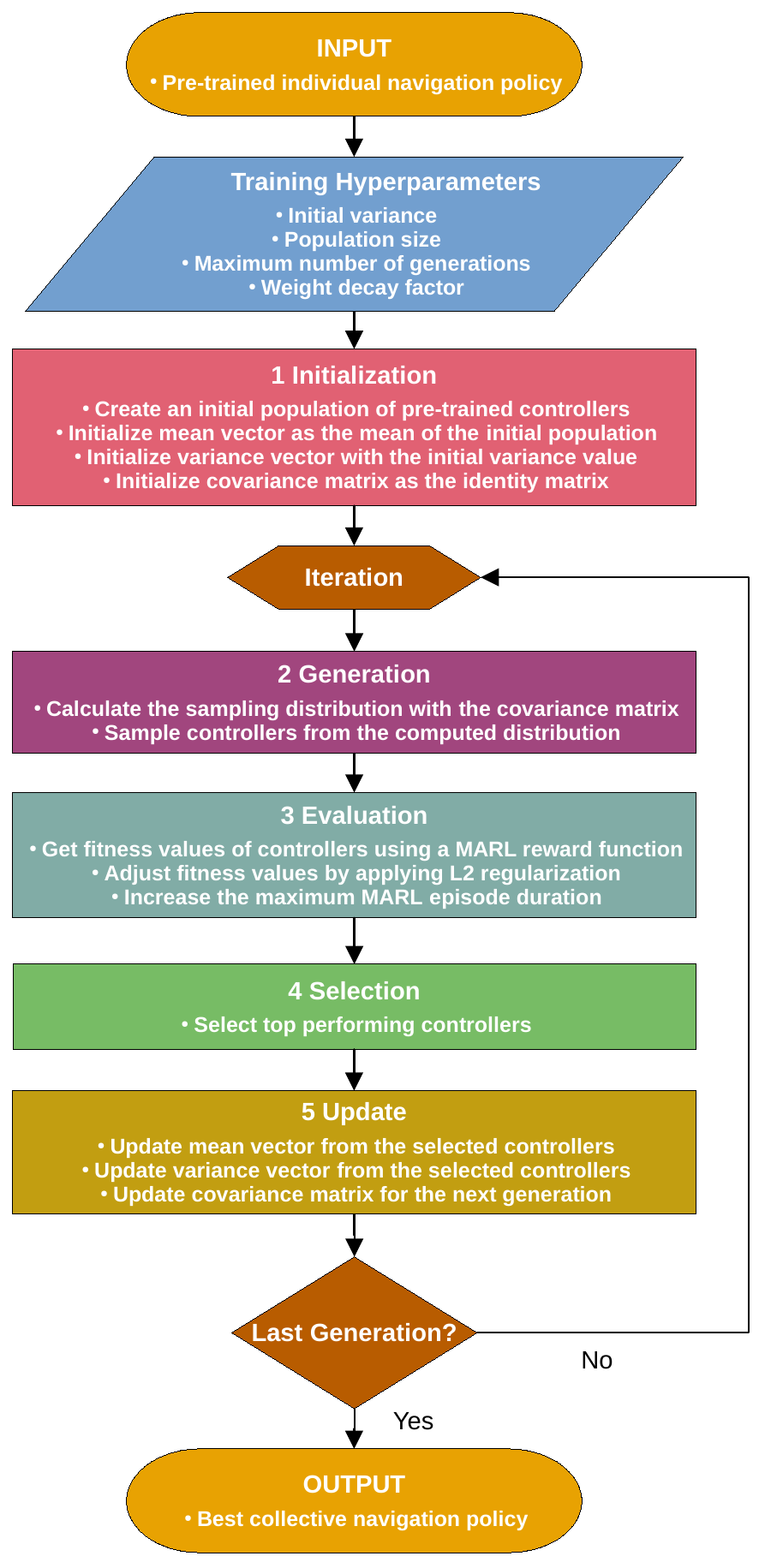}
	\caption{Flowchart of the proposed CMA-ES algorithm.}\label{fig:cma}
\end{figure}

\begin{algorithm}[h!]
	\caption{Pseudo-code implementation of the proposed CMA-ES algorithm.}\label{alg:cma}
	\begin{algorithmic}[1]
		
		\Require{
			\\
			Initial variance $\sigma_{0}$, population size $\rho$, maximum number of generations $\kappa$, individual navigation policy $\pi_{individual}$, weight decay factor $\lambda$.
		}
		\Ensure{
			\\
			Collective navigation policy $\pi_{collective}$.\\
		}
		
		\For{$p = 1$,...,$\rho$}
			\State Initialize an individual of $W_{0}$ with $w_{p}^{0} = \pi_{individual}$.
		\EndFor
		
		\State $k = 1$.
		\State Initialize mean vector $\mu_{k}$ as the mean of $W_{0}$.
		\State Initialize variance vector $\sigma_{k}$ with $\sigma_{0}$.
		\State Initialize covariance matrix with $V_{k} = I$.
		
		\While{$k <= \kappa$}
		
			\For{$p = 1$,...,$\rho$}
				\State Create a new individual for $W_{k}$ with Eq. (\ref{eqn:cma}).
			\EndFor
			
			\For{$p = 1$,...,$\rho$}
				\State Get the fitness value of an individual of $W_{k}$ with Eq. (\ref{eqn:reward}).
			\EndFor
			
			\For{$p = 1$,...,$\rho$}
				\State Adjust the fitness value of an individual of $W_{k}$ with $\lambda$ and Eq. (\ref{eqn:decay}).
			\EndFor
			
			\State Select individuals with the highest fitness from $W_{k}$.
			\State Update $\mu_{k + 1}$ according to the mean of the selected individuals.
			\State Update $\sigma_{k + 1}$ according to the standard deviation of the selected individuals.
			\State Update covariance matrix $V_{k + 1}$ for the next generation.
			\State $k = k + 1$.
			
		\EndWhile    
		
	\end{algorithmic}
\end{algorithm}

While CMA-ES shares operational principles with the cross-entropy method, the generation of individuals at each iteration $k$ differs significantly. For CMA-ES, each candidate solution $w_{p}^{k}$ is produced by combining the current generation's mean $\mu_{k}$ with a multivariate Gaussian distribution $N(0, \sigma_{k}^{2} V_{k})$ centered at $0$. This distribution incorporates the variance vector $\sigma_{k}$ and current covariance matrix $V_{k}$. Thus, we can express the generation of new potential solutions as:
\begin{equation}
	w_{p}^{k} = \mu_{k} + N(0, \sigma_{k}^{2} V_{k}).\label{eqn:cma}
\end{equation}

Our CMA-ES implementation initializes population $W_{0}$ using the individual navigation controller $\pi_{individual}$ weights. To jump-start the exploration around this pre-trained baseline, a global initial step size (\textit{i.e.}, the variance $\sigma_{0}$) is defined as a hyperparameter, and the covariance matrix $V_{k}$ is initialized as the identity matrix, $I$. Unlike our CEM implementation, this evolution strategy does not maintain a fixed number of high-performing individuals. Instead, it selects solutions achieving higher accumulated rewards, which can undergo recombination and mutation \citep{Hansen_2001}. These selected individuals inform the updates of the mean vector $\mu_{k + 1}$, variance vector $\sigma_{k + 1}$, and covariance matrix $V_{k + 1}$ for the subsequent generation.

Inspired by related work in neuro-evolution \citep{Ha_2017}, we incorporate an L2 regularization technique into the fitness function (\textit{i.e.}, the proposed multi-agent reinforcement learning reward function). This method penalizes large network weights and promotes simpler, more generalizable solutions \citep{Moradi_2019}. The adjusted fitness $\mathcal{R}^{\prime}(W_{k})$ for a population is calculated as:
\begin{equation}
	\mathcal{R}^{\prime}(W_{k}) = \mathcal{R}(W_{k}) - \lambda \frac{1}{\rho}\sum\limits_{p = 1}^{\rho}(w_{p}^{k} * w_{p}^{k}),\label{eqn:decay}
\end{equation}
where $\lambda$ is the weight decay factor and $\rho$ is the population size. As noted by \citet{Moradi_2019}, this regularization helps prevent overfitting to the training scenarios and enhances the robustness of the final evolved policy. Thereby, our approach aligns with recent research trends highlighted by scholars such as \citet{Hasselmann_2021}. They suggest that incorporating deep learning regularization techniques represents the most promising direction for automatically designing swarm robotic controllers through neuro-evolution. The complete workflow of our CMA-ES implementation is depicted in Fig. \ref{fig:cma} and detailed in Algorithm \ref{alg:cma}.

\section{Collective Policy Comparison}\label{sec:experiments}

This section provides a rigorous empirical evaluation of the proposed hybrid AI-driven methodology. We conduct a multi-stage analysis centered on collective visual exploration within a simulated, dynamic environment using a low-cost robotic swarm. The mobile agents are tasked with executing collision-free navigation using only monocular camera RGB images as their sensory input, and critically, without any form of inter-agent communication, thereby ensuring system simplicity and robustness. The virtual vision device mounted on each robot was modeled after the specifications of the Raspberry Pi Camera Module 2\footnote{https://www.raspberrypi.com/products/camera-module-v2/}.

To establish a robust benchmark, the performance of our most promising evolved neural controllers is quantitatively and qualitatively compared against a conventional Light Detection and Ranging (LiDAR)-based navigation approach \citep{Kolar_2020}. This reactive control method draws inspiration from classical bio-inspired navigation paradigms. Specifically, it is based on Braitenberg vehicles and the concept of negative tropotaxis, where agents exhibit an aversion response to certain stimuli \citep{Shaikh_2020}. In our work, each robot's LiDAR readings of nearby obstacles serve as the aversive stimulus, generating appropriate motor responses that direct the mobile agents away from detected obstacles.

LiDAR technology is widely adopted in mobile robotics due to its high precision in distance measurement. In this baseline scenario, each virtual agent was equipped with a simulated distance sensor equivalent to the Slamtec RPLIDAR A1 laser scanner\footnote{https://www.slamtec.com/en/lidar/a1}. Comparing our learned vision-based controllers against this traditional, sensor-heavy method allows us to assess their relative performance, comprehensively highlighting the potential advantages in efficiency, cost reduction, and adaptability of the proposed monocular approach.

While evaluating our proposed hybrid methodology, we intentionally selected this traditional LiDAR-based reactive approach as our primary baseline rather than gradient-based deep reinforcement learning algorithms, such as PPO or TRPO, or alternative neuro-evolutionary approaches. Although PPO and TRPO represent the state of the art in continuous control, conducting a direct comparison with these methodologies within our specific evaluation framework would present a fundamental paradigm mismatch.

Representative DL-based MARL architectures often assume the existence of explicit multi-agent collaboration mechanisms (\textit{e.g.}, local communication networks to share state information), or demand precise spatial data (\textit{e.g.}, global coordinates) to mitigate the challenges of sparse rewards. In contrast, our approach fundamentally champions a fully communication-free, decentralized paradigm to maximize system simplicity and scalability on severely resource-constrained hardware.

Forcing such gradient-based algorithms to operate under the strict constraints of our swarm architecture (\textit{i.e.}, purely decentralized execution, zero inter-agent communication, and raw monocular vision compressed into a 32-variable latent vector) would severely degrade their convergence capabilities, leading to an unfair and uninformative comparison. Therefore, the conventional LiDAR-based navigation paradigm provides a far more rigorous and pragmatic proof of the proposed system's viability for real-world swarm applications than artificial comparisons with heavily constrained DRL algorithms.

The experiments were conducted using our proprietary high-fidelity physics simulator. This simulator provides a controlled yet realistic testing ground, facilitating the accurate modeling of diverse indoor environments and commercial robot platforms. It utilizes the Planck.js\footnote{https://piqnt.com/planck.js/} 2D physics engine, which is built upon the widely recognized Box2D\footnote{https://box2d.org/} engine, and is seamlessly integrated with our deep learning, reinforcement learning, and evolutionary computation frameworks.

This evaluation process is structured as follows: first, in Subsection \ref{subsec:hyperparameters}, a sensitivity analysis of evolutionary hyperparameters is conducted to justify the selection of the primary collective policies. Subsequently, in Subsections \ref{subsec:performance} and \ref{subsec:statistical}, the chosen policies are benchmarked against the LiDAR baseline using a comprehensive suite of energy and navigation metrics.

\subsection{Hyperparameter Sensitivity Analysis}\label{subsec:hyperparameters}

The automatic synthesis of decentralized swarm controllers through neuro-evolutionary strategies is highly contingent upon the precise calibration of the search space parameters. While an exhaustive grid search is the standard approach for hyperparameter tuning in many deep learning contexts, executing a full-factorial grid search over all possible hyperparameter combinations is computationally prohibitive in our specific framework. This limitation arises from the online nature of our MARL process, where each fitness evaluation requires a full multi-agent physics simulation within a non-stationary environment subject to continuous lighting variations. Consequently, our sensitivity analysis is structured as a focused, partial search, deliberately fixing certain parameters to preserve pre-trained knowledge while varying the hyperparameters directly responsible for managing the exploration-exploitation trade-off.

The evolutionary algorithms begin with an initial population seeded by a pre-trained individual map-less navigation policy (\textit{i.e.}, $\pi_{individual}$). This policy possesses valuable reactive static obstacle avoidance capabilities, learned via supervised learning from a synthetic dataset, as explained in Subsection \ref{subsec:individual}. To transition from individual static avoidance to collective dynamic avoidance, we fine-tuned the underlying variational autoencoder using unsupervised learning, enabling the extraction of visual features associated with dynamic obstacles (\textit{i.e.}, other mobile agents).

Crucially, to prevent the catastrophic forgetting of these pre-trained visual features and foundational navigation skills, we constrained the initial evolutionary search space. Specifically, we intentionally fixed the population size to a relatively small number of 32 candidate controllers per generation. We also bounded the initial variance hyperparameters to low values: 1.5 for the cross-entropy method and 1.0 for the covariance matrix adaptation evolution strategy. Allowing massive populations or high initial variances would induce an overly diffusive search, destroying the established collision-free navigation parameters before beneficial emergent collective behaviors could consolidate.

Furthermore, a curriculum learning strategy was implemented not merely as an optional enhancement, but as a critical acceleration heuristic to balance exploration and exploitation. The maximum episode duration, $T_{k}$, progressively increases over the training iterations according to Eq. (\ref{eqn:curriculum}). Starting from a moderate initial value of $T_{0} = 500$ simulation steps, the duration was gradually extended to a maximum of 4500 time steps by the final iteration. Given that each simulation step equals $1/3$ of a second, episode durations scale from approximately 2.5 minutes in the first generation to 25 minutes in the last. Because any collision strictly and immediately terminates an episode during the training phase, starting optimization directly with 4500-step episodes would severely penalize early exploratory actions, drastically slowing down convergence and increasing the likelihood of premature overfitting to sub-optimal, overly conservative stop-and-wait behaviors.

The evolutionary process itself was bounded to a maximum of 2000 generations. This termination threshold was not arbitrarily selected; preliminary empirical tests indicated that the learning curves for both the CEM and CMA-ES algorithms consistently reached a convergence plateau near this generation mark. Extending the online multi-agent simulations beyond 2000 generations yielded only marginal performance gains that simply did not justify the exponentially compounding computational costs.

Building upon these mathematically and computationally justified constraints, we systematically evaluated three distinct configurations for the remaining key hyperparameters of both the CEM and CMA-ES algorithms. The results of this evaluation process allows us to justify the selection of the final policies based on convergence stability and spatial coverage effectiveness.

\subsubsection{Experimental Setup}
\label{subsubsec:setup}

\begin{figure*}[h!]
	\centering
	\includegraphics[width=\textwidth]{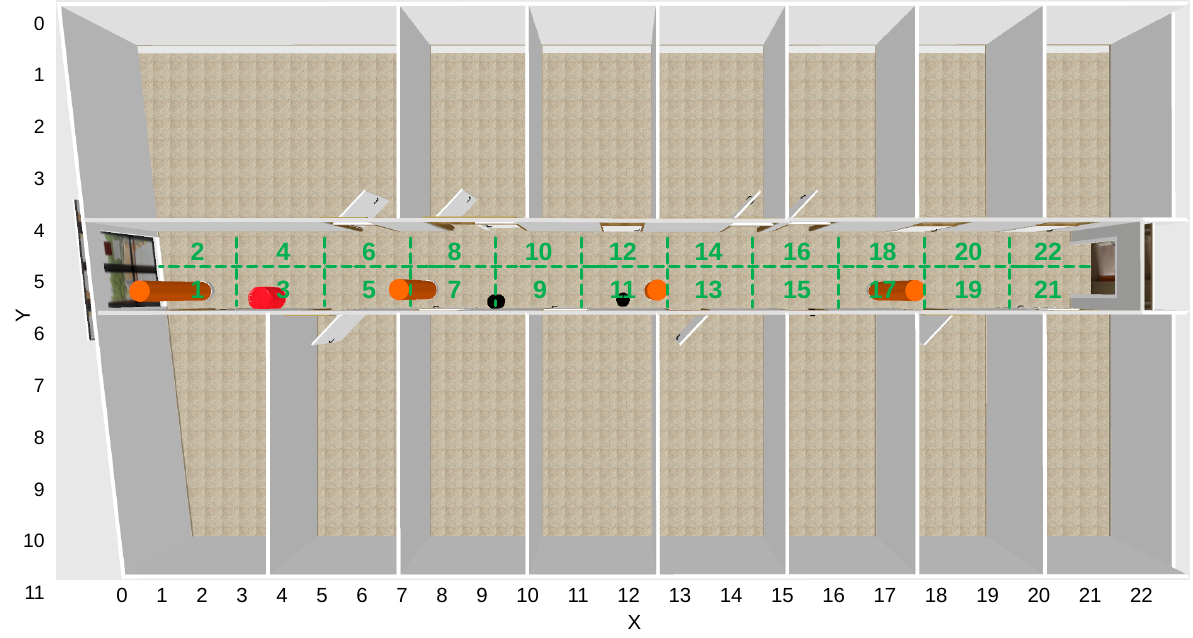}
	\caption{Division of the simulated environment into 22 rectangular sectors. This map shows the partitioned indoor corridor (in green) used for quantitative evaluation of spatial coverage. The corridor has a navigable surface approximately 20 meters long and 2 meters wide.}\label{fig:environment}
\end{figure*}

The training and evaluation of the swarm controllers were conducted in a simulated indoor corridor environment. To provide a granular assessment of collective behavior, the environment was explicitly partitioned into 22 rectangular sectors, as illustrated in Fig. \ref{fig:environment}.

To quantify the efficacy of the evolved policies, two primary metrics were employed:

\begin{itemize}
	
	\item \textbf{Accumulated Reward}: The total number of reward points garnered by the system during a multi-agent reinforcement learning episode, calculated using Eq. (\ref{eqn:reward}). This serves as a global measure of task effectiveness, actively incentivizing exploration while heavily penalizing collisions.
	
	\item \textbf{Explored Sectors}: The absolute count of distinct navigable regions in the environment visited at least once by any robot. This metric circumvents the abstraction of reward points to provide a direct, tangible measure of spatial coverage.
	
\end{itemize}

\subsubsection{Cross-Entropy Method Analysis}\label{subsubsec:cem_hyperparameters}

Our CEM implementation depends heavily on the elite-set size, $\epsilon$, and the duration of the supplementary variance injection, $\gamma$. The elite set fosters exploitation by propagating the most promising candidate solutions, while the extra variance injection acts as a targeted mutation mechanism to prevent premature convergence in the high-dimensional weight space. We evaluated three distinct configurations to isolate the impact of these factors: elitism rates of 20\%, 30\%, and 40\%, corresponding to elite sets of 6, 9, and 12 individuals from the population of 32, combined with extra variance injections spanning the first 100, 500, and 1000 generations, respectively. Table \ref{tab:cem_hyperparameters} formally details these experiments.

\begin{figure*}[h!]
	\centering
	\includegraphics[width=\textwidth]{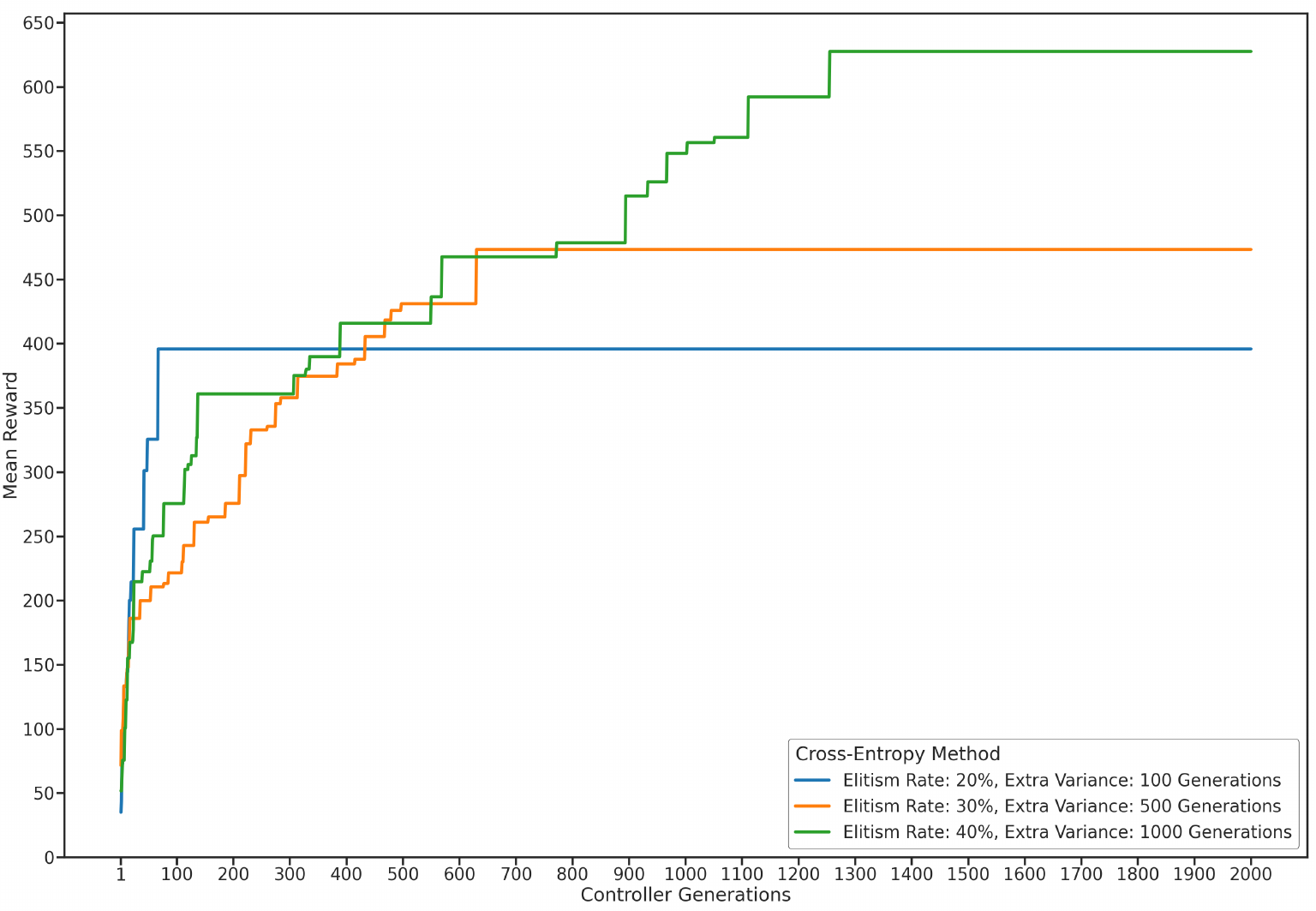}
	\caption{Evolution of the mean accumulated reward during 2000 generations of CEM training. The curves represent the filtered improvement progress for hyperparameter configurations (1) (in blue), (2) (in orange), and (3) (in green).}\label{fig:cem_training}
\end{figure*}

\begin{table*}[h!]
	\begin{minipage}{\textwidth}
		\begin{center}
			\caption{Hyperparameter configurations for the CEM algorithm sensitivity analysis.}\label{tab:cem_hyperparameters}
			\begin{tabular}{@{}ccccc@{}}
				\hline
				\multirow{2}{*}{\textbf{Hyperparameter}} & \multirow{2}{*}{\textbf{Description}} & \multicolumn{3}{c}{\textbf{Configuration}}\\
				& & \textbf{(1)} & \textbf{(2)} & \textbf{(3)}\\
				\hline
				\hline
				$J$ & Agents in the system & 4 & 4 & 4\\
				$T$ & Time steps per episode & 500-4500 & 500-4500 & 500-4500\\
				$C$ & Collision termination threshold & 0 & 0 & 0\\
				$\kappa$ & Total generations & 2000 & 2000 & 2000\\
				$\rho$ & Population size & 32 & 32 & 32\\
				$\sigma_{0}$ & Initial variance & 1.5 & 1.5 & 1.5\\
				$\epsilon$ & Elite-set size & 6 & 9 & 12\\
				$\gamma$ & Generations with extra variance & 100 & 500 & 1000\\
				\hline
			\end{tabular}
		\end{center}
	\end{minipage}
\end{table*}

\begin{figure*}[h!]
	\centering
	\includegraphics[scale=0.60]{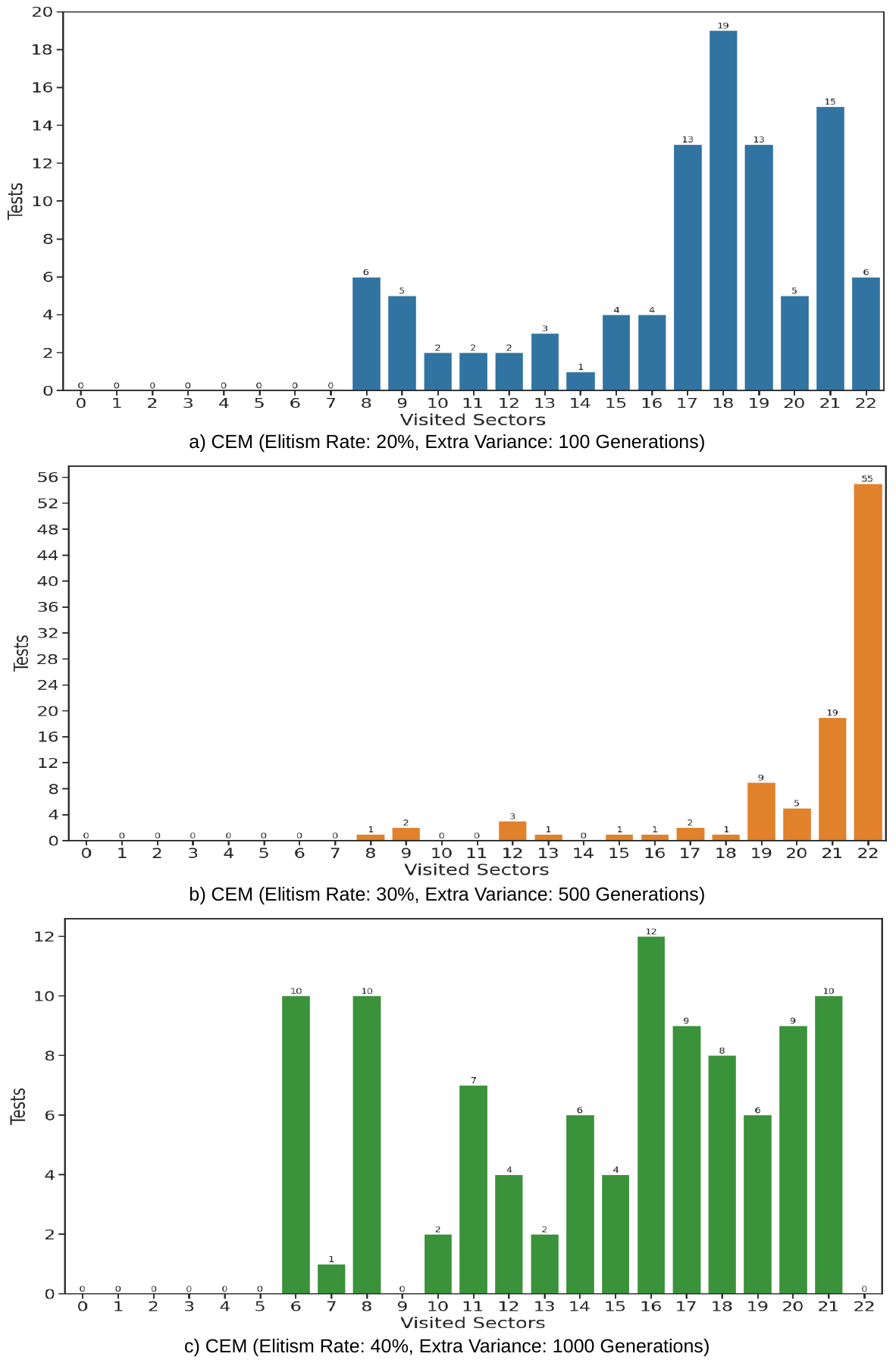}
	\caption{Number of tests achieving specific numbers of visited sectors. The diagrams compare the ability of our three CEM hyperparameter configurations to explore the indoor corridor across 100 validation trials.}\label{fig:cem_results}
\end{figure*}

The training progression curves depicted in Fig. \ref{fig:cem_training} explicitly reveal the impact of these hyperparameter trade-offs. Configuration (1) converges rapidly to a local optimum, achieving a mean reward of 396.03 points as early as generation 67, due to an insufficient variance injection period that stifles exploration. Conversely, Configuration (3) displays an overly diffusive search behavior, postponing its highest mean reward, 627.58 points, until generation 1255, hindering the early consolidation of high-performing policies.

Configuration (2), with a 30\% elitism rate and 500 generations of extra variance, exhibits the most favorable and stable learning trajectory. This is rigorously confirmed by our post-training validation phase, spanning 100 independent trials summarized in Table \ref{tab:cem_results}. In these validation tests, Configuration (2) achieves a superior mean spatial coverage of 20.38 sectors and an average reward of 4673.10 points, significantly outperforming Configuration (1), and specially, Configuration (3). This latter milestone proves that excessively high values for elitism and variance injection do not necessarily yield superior performance.

Beyond achieving the highest mean spatial coverage, Configuration (2) also demonstrates superior policy stability. By extracting the distribution of visited sectors from Fig. \ref{fig:cem_results}, we calculate a standard deviation of 3.05 for Configuration (2), compared to 4.03 and 4.88 for Configurations (1) and (3), respectively. This lower dispersion quantitatively confirms that the selected hyperparameter balance not only maximizes exploration but also ensures highly consistent collective behavior across independent evaluations.

\begin{table}[h!]
	\begin{minipage}{\columnwidth}
		\begin{center}
			\caption{Quantitative performance comparison of CEM configurations across 100 validation trials.}\label{tab:cem_results}
			\begin{tabular}{@{}ccc@{}}
				\hline
				\textbf{Configuration} & \textbf{Sectors} & \textbf{Reward}\\
				\hline
				\hline
				(1) & 16.93 & 3573.56\\
				(2) & \textbf{20.38} & \textbf{4673.10}\\
				(3) & 14.55 & 2483.32\\
				\hline
			\end{tabular}
		\end{center}
	\end{minipage}
\end{table}

Reflecting on these results, the superiority of Configuration (2), hereafter denoted as $\pi_{collective}^{CEM}$, stems from its ability to maintain swarm coherence while visiting all 22 sectors of the environment in 55\% of trials, as shown in Fig. \ref{fig:cem_results}. This configuration effectively balances the risk of collision with the imperative of forward exploration, satisfying the robustness requirements for autonomous navigation.

\subsubsection{Covariance Matrix Adaptation Evolution Strategy Analysis}\label{subsubsec:cma_hyperparameters}

For the CMA-ES algorithm, our sensitivity study focused on the weight decay factor, $\lambda$, used for L2 regularization. This parameter is crucial in deep neuro-evolution; it actively penalizes large network weights to prevent overfitting to the specific training scenarios and mitigates the vanishing gradient problem \citep{Pierson_2017}, thereby directly enhancing the generalization robustness of the final evolved policy. Three decay factors were systematically tested: 0.01, 0.05, and 0.1, representing low, moderate, and high regularization constraints. Table \ref{tab:cma_hyperparameters} presents the precise configurations evaluated.

\begin{figure*}[h!]
	\centering
	\includegraphics[width=\textwidth]{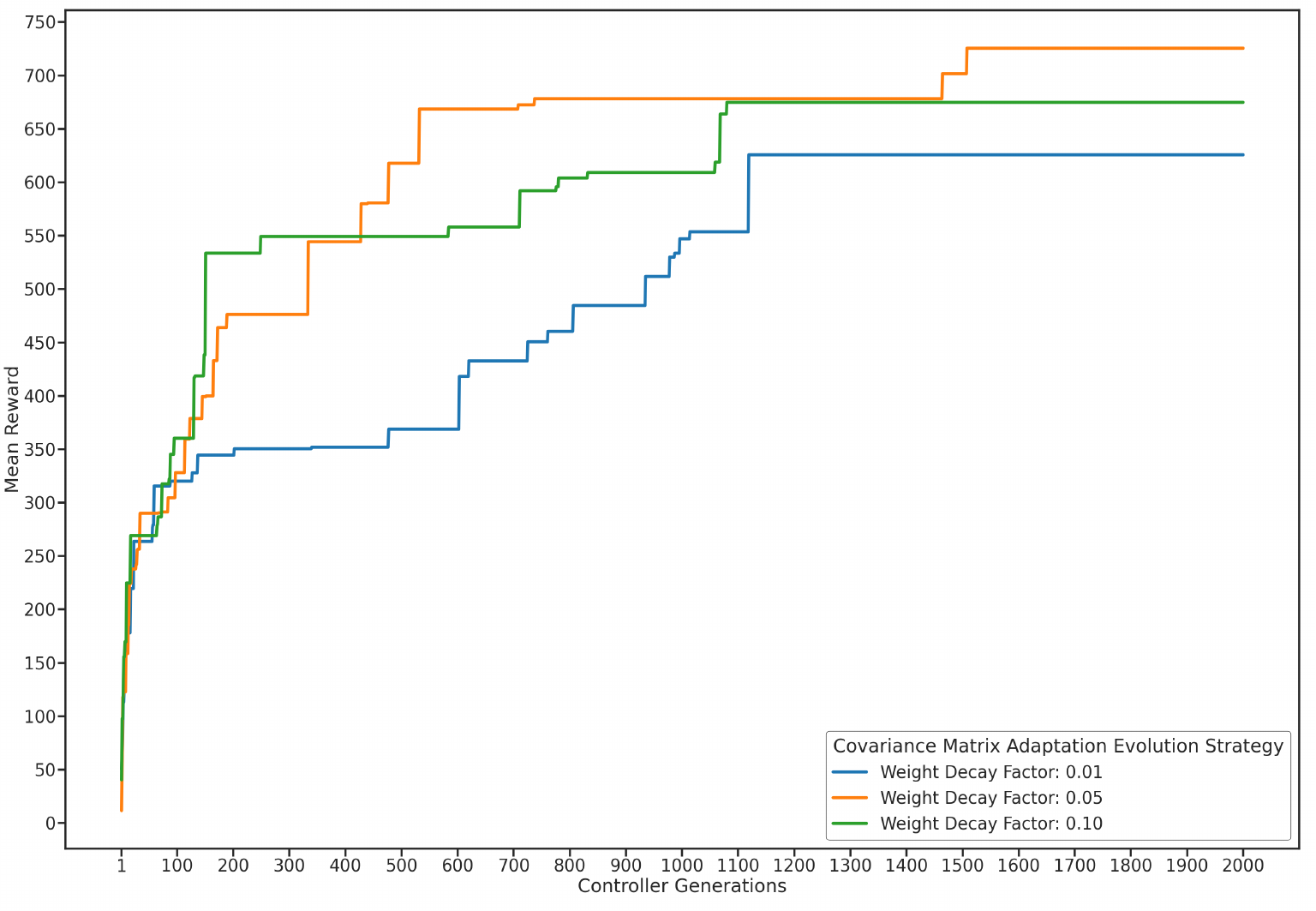}
	\caption{Evolution of the mean accumulated reward during 2000 generations of CMA-ES training. The curves represent the filtered improvement progress for hyperparameter configurations (4) (in blue), (5) (in orange), and (6) (in green).}\label{fig:cma_training}
\end{figure*}

\begin{table*}[h!]
	\begin{minipage}{\textwidth}
		\begin{center}
			\caption{Hyperparameter configurations for the CMA-ES algorithm sensitivity analysis.}\label{tab:cma_hyperparameters}
			\begin{tabular}{@{}ccccc@{}}
				\hline
				\multirow{2}{*}{\textbf{Hyperparameter}} & \multirow{2}{*}{\textbf{Description}} & \multicolumn{3}{c}{\textbf{Configuration}}\\
				& & \textbf{(4)} & \textbf{(5)} & \textbf{(6)}\\
				\hline
				\hline
				$J$ & Agents in the system & 4 & 4 & 4\\
				$T$ & Time steps per episode & 500-4500 & 500-4500 & 500-4500\\
				$C$ & Collision termination threshold & 0 & 0 & 0\\
				$\kappa$ & Total generations & 2000 & 2000 & 2000\\
				$\rho$ & Population size & 32 & 32 & 32\\
				$\sigma_{0}$ & Initial variance & 1.0 & 1.0 & 1.0\\
				$\lambda$ & Weight decay factor & 0.01 & 0.05 & 0.1 \\
				\hline
			\end{tabular}
		\end{center}
	\end{minipage}
\end{table*}

\begin{figure*}[h!]
	\centering
	\includegraphics[scale=0.60]{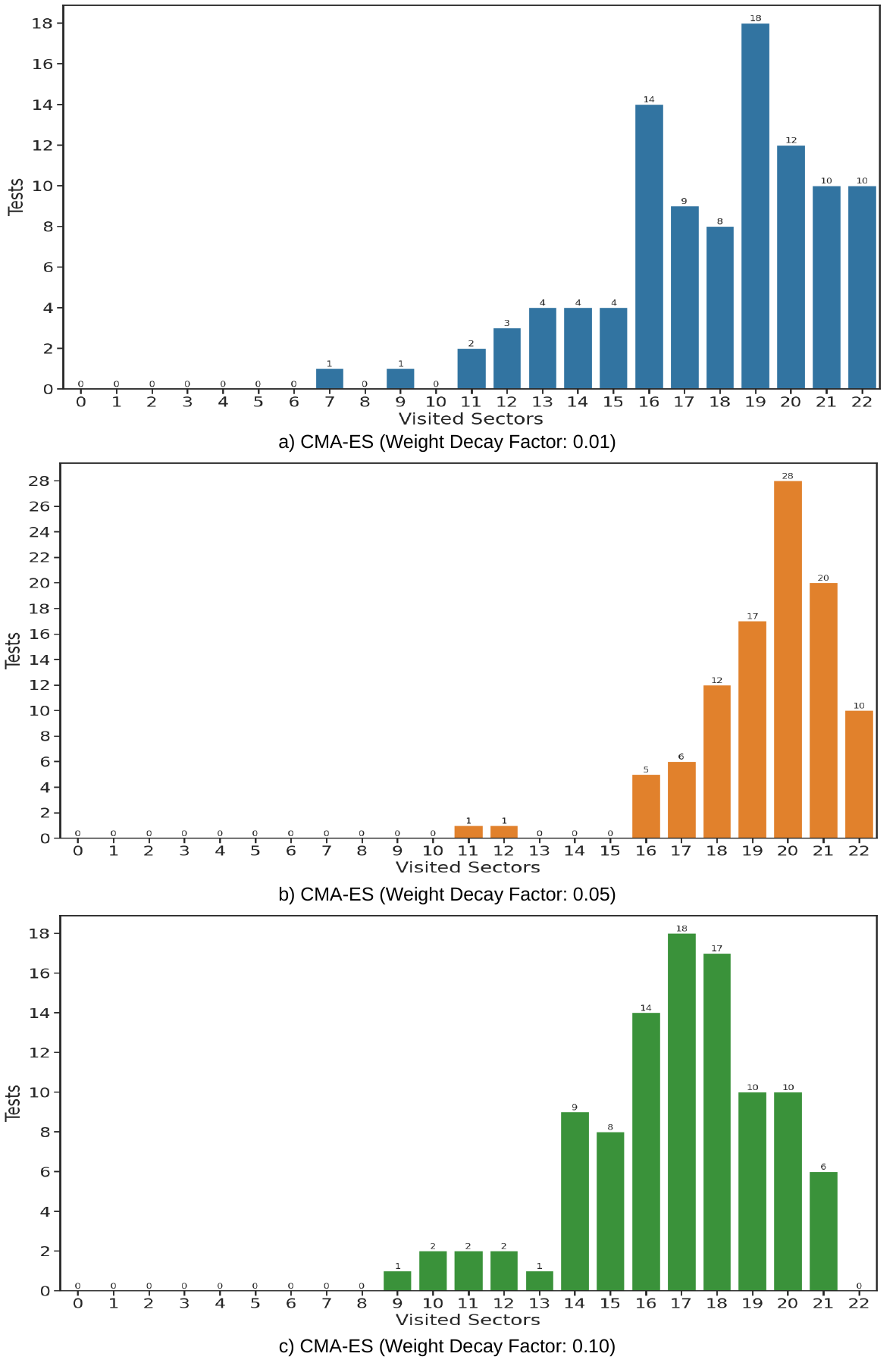}
	\caption{Number of tests achieving specific numbers of visited sectors. The diagrams compare the ability of our three CMA-ES hyperparameter configurations to explore the indoor corridor across 100 validation trials.}\label{fig:cma_results}
\end{figure*}

As depicted in the respective training curves, the CMA-ES algorithm exhibits a characteristically smoother fitness growth compared to the cross-entropy method, attributable to its second-order optimization nature and dynamic covariance matrix adaptation. Configuration (5), utilizing a moderate weight decay factor of 0.05, achieves the most robust performance, reaching a peak mean reward of 725.58 points at generation 1508. As observed in Fig. \ref{fig:cma_training}, a lower decay factor (\textit{i.e.}, Configuration (4)) results in more erratic fitness growth and suboptimal rewards, whereas excessive regularization (\textit{i.e.}, Configuration (6)) overly restricts the neural controller's expressive capacity, capping the mean reward at 674.95 points by generation 1080.

The validation phase unequivocally justifies the selection of Configuration (5), hereafter denoted as $\pi_{collective}^{CMA-ES}$. As shown in Table \ref{tab:cma_results}, it demonstrates the most balanced performance, achieving 19.44 mean sectors explored and a robust average reward of 4587.92 points across 100 trials, signifying a policy that is both spatially effective and highly safe.

\begin{table}[h!]
	\begin{minipage}{\columnwidth}
		\begin{center}
			\caption{Quantitative performance comparison of CMA-ES configurations across 100 validation trials.}\label{tab:cma_results}
			\begin{tabular}{@{}ccc@{}}
				\hline
				\textbf{Configuration} & \textbf{Sectors} & \textbf{Reward}\\
				\hline
				\hline
				(4) & 17.75 & 1195.83\\
				(5) & \textbf{19.44} & \textbf{4587.92}\\
				(6) & 16.86 & 890.18\\
				\hline
			\end{tabular}
		\end{center}
	\end{minipage}
\end{table}

As reflected in Fig. \ref{fig:cma_results}, Configuration (5) exhibits a more robust navigation profile characterized by highly consistent performance. This is statistically supported by the standard deviation of the visited sectors across the 100 validation trials, which is 1.94 for Configuration (5), compared to 3.15 and 2.60 for Configurations (4) and (6), respectively. This dispersion metric explicitly validates that the selected moderate weight decay factor effectively minimizes performance variance, confirming its optimal suitability as the final $\pi_{collective}^{CMA-ES}$ policy.

\subsubsection{Multi-Agent Reinforcement Learning Parameters}\label{subsubsec:marl_parameters}

Regarding the multi-agent reinforcement learning mechanics, each fitness evaluation dictates the simulation of a swarm comprising four differential-drive ground robots operating within our synthetic corridor environment. To pragmatically address the high computational demands inherent to online MARL processes, the fitness value (\textit{i.e.}, the accumulated reward) of any given candidate controller was computed through a single episode of interaction with the environment. We must acknowledge that calculating fitness across multiple independent episodes for each candidate would theoretically yield a more stable estimation and reduce generational variance. However, adopting such an approach would render our online training phase computationally prohibitive.

This single-episode evaluation methodology inherently introduces a degree of training instability and high variance across generations, which justifies the presentation of filtered or smoothed progression curves in our training performance analysis to clearly illustrate the underlying improvement trends without the noise of episodic stochasticity. As observed in Fig. \ref{fig:cem_training} and Fig. \ref{fig:cma_training}, we employ a simplified representation that only updates the reward curve when a configuration achieves a strictly superior mean fitness compared to its previous best. This\text{ }\textquotedblleft{best-so-far}\textquotedblright\text{ }approach filters the inherent stochastic noise of multi-agent interactions and highlights the authentic optimization milestones.

A critical element of our training framework is the implementation of a highly strict termination condition based on collisions. During the training phase, an episode is immediately terminated if any single member of the robotic swarm collides with either a static obstacle or a dynamic one (\textit{i.e.}, another mobile agent). This severe constraint is intentionally designed to force the evolutionary algorithms to prioritize truly effective, collision-free emergent behaviors from the very first generation.

This strict penalization synergizes directly with the aforementioned curriculum learning strategy. If early generations were evaluated over excessively long episodes, the probability of encountering a collision-inducing edge case would approach absolute certainty, thereby flattening the fitness landscape and stalling the exploration of the solution space. Conversely, initiating the training with shorter episodes (\textit{e.g.}, 500 steps) establishes an achievable survival milestone. It increases the probability that early, unrefined candidate policies can successfully complete a full episode without triggering the severe collision penalty, providing a clear and stable selection signal for the evolutionary algorithm before progressively increasing the temporal complexity of the navigation task.

Furthermore, to thoroughly evaluate the generalization capabilities of the evolved controllers and actively prevent overfitting to the specific training environment, we established a rigorous methodological dichotomy between the training and evaluation conditions. During training, the ambient illumination of the environment undergoes continuous variations to build basic perceptual robustness in the visual feature extraction. Concurrently, while the initial position and orientation of each robot within the swarm are randomized, the swarm as a whole is constrained to adopt a single initial formation centralized within the inner corridor. This controlled initialization was deliberately selected to guarantee a faster, more stable convergence towards satisfactory collective navigation policies, preventing excessive early-stage complexity from completely disrupting the emergence of cooperative behaviors.

In stark contrast, the evaluation phase was designed to rigorously test the boundaries and robustness of the learned policies under unforeseen conditions. During the validation trials, the swarm is randomly assigned one of six distinct initial formations. This deliberate divergence from the training conditions ensures that the macroscopic performance is not merely a product of memorizing a specific starting configuration. Additionally, during evaluation, the lighting conditions are held constant throughout the entirety of each episode, with the specific lighting level randomly determined at the onset. This methodological choice allows the simulation to emulate specific environmental conditions, such as varying levels of ambient daylight at different specific hours, thereby verifying the true adaptability of the vision-based policies to prolonged environmental shifts.

The duration of our evaluation trials is deliberately set to 5000 simulation steps, which exceeds the maximum curriculum training horizon of 4500 steps. This temporal extension serves as a rigorous stress test to verify the stability and generalization capabilities of the learned policies. By pushing the swarm beyond the temporal bounds encountered during the evolutionary phase, we ensure that the agents can reliably sustain decentralized, collision-free operation over prolonged periods without suffering from unforeseen behavioral degradation. Finally, unlike the training phase, agents are permitted to recover and continue exploring after a collision during evaluation, albeit with the severe reward penalty specified in Eq. (\ref{eqn:reward}).

\subsection{Experimental Validation and Behavioral Analysis}\label{subsec:performance}

In this subsection, we present and discuss the quantitative and qualitative results derived from the validation phase of the two most promising visual exploration policies, $\pi_{collective}^{CEM}$ and $\pi_{collective}^{CMA-ES}$, in comparison with the LiDAR-based benchmark. Furthermore, to directly address the sensitivity of these policies to environmental constraints, this subsection culminates in a dedicated ablation study isolating the impact of collision termination rules on the emergent exploratory behaviors.

\subsubsection{Validation Metrics}\label{subsubsec:metrics}

To holistically measure the performance, behavioral stability, and energy efficiency of the collective policies across 100 independent validation trials, we monitored the mean cumulative reward and the number of explored sectors, both initially introduced in Subection \ref{subsec:hyperparameters} for the hyperparameter sensitivity study. In addition to these primary task-oriented indicators, we incorporated the following three core validation metrics:

\begin{itemize}
	
	\item \textbf{Inter-Robot Collisions Percentage}: The percentage of total episode time during which one or more swarm members experienced a collision with another individual agent. We focused on collisions with dynamic obstacles (\textit{i.e.}, other robots) as the training goal was to evolve policies to avoid moving objects.
	
	\item \textbf{Traveled Distance}: The cumulative sum of individual displacements, quantified in meters (m), executed by all mobile agents in the system. Considering a swarm composed of $J$ robots, the total distance $\delta_{total}$ traveled during $T$ time instants is formally expressed by the equation:
	\begin{align}
		\delta_{total} &= \sum\limits_{j = 1}^{J}\sum\limits_{t = 1}^{T} \delta_{t}^{j},\label{eqn:distance}\\
		\delta_{t}^{j} &= \sqrt{(\chi_{t + 1}^{j} - \chi_{t}^{j})^{2} + (\psi_{t + 1}^{j} - \psi_{t}^{j})^{2}}.\nonumber
	\end{align}

	In Eq. (\ref{eqn:distance}), $\delta_{t}^{j}$ refers to the Euclidean distance between two points, represented by the $(\chi, \psi)$ coordinates of two environment positions traversed by the individual agent $e_{j}$ in consecutive simulation steps, $t$ and $t + 1$.

	\item \textbf{Energy Consumption}: The total energy consumed by the entire swarm per episode, quantified in watt-seconds (Ws), providing a critical quantitative metric for future hardware deployment efficiency. This is calculated by summing the instantaneous motor and sensor energy consumption \citep{Hausler_2016}, $\zeta_{motors, t}^{j}$ and $\zeta_{sensor, t}^{j}$, for all agents $e_{j}$, as shown in Eq. (\ref{eqn:energy}).
	\begin{align}
		\zeta_{total} &= \sum\limits_{j = 1}^{J}\sum\limits_{t = 1}^{T} \zeta_{t}^{j},\label{eqn:energy}\\
		\zeta_{t}^{j} &= \zeta_{motors, t}^{j} + \zeta_{sensor, t}^{j}.\nonumber
	\end{align}

	We estimated the instantaneous energy consumption corresponding to each robot's motors, $\zeta_{motors, t}^{j}$, by implementing the following quadratic model:
	\begin{align}
		\zeta_{motors, t}^{j} &= (2 * P_{base} + K_{1} * \nu_{t}^{j} + K_{2} * {\nu_{t}^{j}}^{2}) * \Delta t,\label{eqn:movement}\\
		\nu_{t}^{j} &= \frac{\delta_{t}^{j}}{\Delta t}.\nonumber
	\end{align}

	The quadratic model from Eq. (\ref{eqn:movement}) accounts for the base power $P_{base}$ in steady state of each of the two motors of our differential-drive ground vehicles, the velocity $\nu_{t}^{j}$ in meters per second (m/s) at which each individual agent moves in each simulation step $\Delta t$, and two empirical coefficients, $ K_{1}$ and $ K_{2}$, obtained through experimental characterization related to the motors. The motor model used was the 180 Optical Encoder, typically integrated into low-cost Makeblock mBot Ranger platforms. Based on its technical specifications\footnote{https://www.application-datasheet.com/pdf/makeblock/81340.pdf}, we determined that $P_{base} = 0.5$ W.
	
	We calculated the instantaneous energy consumption associated with the sensor embedded in each robot by multiplying the nominal power $P_{sensor}$ of the input device by the value assigned to $\Delta t$ (\textit{i.e.}, $1/3$ of a second), which is formally expressed as:
	\begin{equation}
		\zeta_{sensor, t}^{j} = P_{sensor} * \Delta t.\label{eqn:sensing}
	\end{equation}
	
	For the conventional LiDAR-based approach, after consulting the datasheet\footnote{http://bucket.download.slamtec.com/b90ae0a89feba3756bc5aaa0654c296dc76ba3ff/\\LD108\_SLAMTEC\_rplidar\_datasheet\_A1M8\_v2.2\_en.pdf} of the Slamtec RPLIDAR A1 laser scanner, we established that $P_{sensor} = 2$ W. In the case of the monocular RGB camera used by the proposed method, this value was significantly lower, specifically $P_{sensor} = 0.3$ W. To determine this second nominal power value, we started from the technical specifications\footnote{https://www.opensourceinstruments.com/Electronics/Data/IMX219PQ.pdf} of the Sony IMX219 CMOS image sensor used by the Raspberry Pi Camera Module 2.
	
	It is critical to clarify that the sensors are modeled in a continuous active state to ensure gapless reactive collision avoidance. Furthermore, this macroscopic metric strictly isolates the mechanical and sensory energy footprint; the computational overhead of the VAE inference is intentionally excluded at this simulation stage, as deep learning inference power consumption is highly variable and heavily dependent on the final embedded edge AI hardware selected for physical deployment.

\end{itemize}

\subsubsection{Quantitative Macroscopic Performance}\label{subsubsec:quantitative}

\begin{figure*}[h!]
	\centering
	\includegraphics[width=\textwidth]{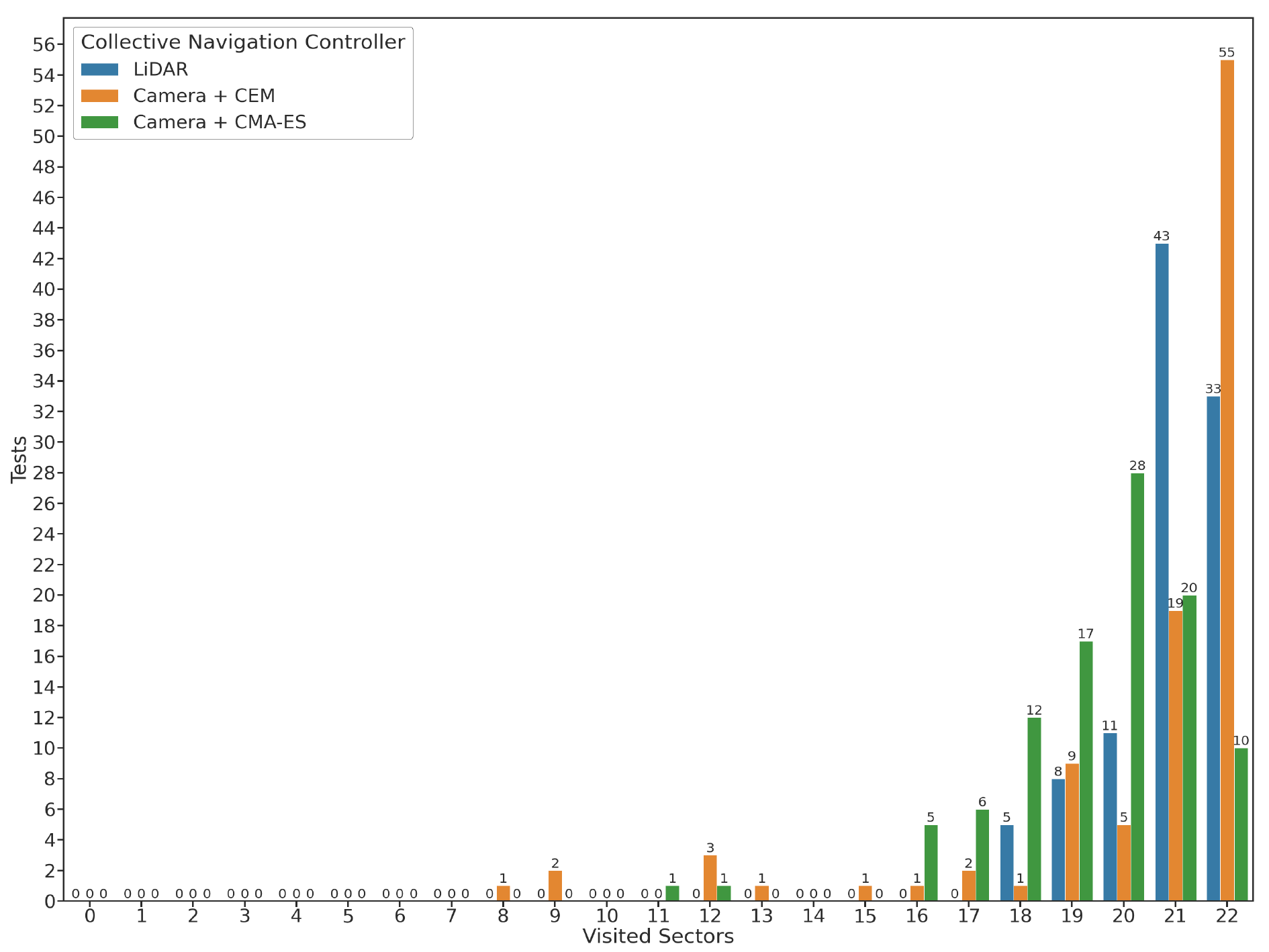}
	\caption{Number of tests achieving specific numbers of visited sectors. This bar chart compares the collective exploration performance of the LiDAR baseline (in blue), the proposed vision-based controller trained with our CEM algorithm (in orange), and the proposed vision-based controller trained with our CMA-ES algorithm (in green) over 100 multi-agent reinforcement learning episodes.}\label{fig:quantitative}
\end{figure*}

Fig. \ref{fig:quantitative} presents the distribution of visited sectors across the 100 validation tests for all three evaluated approaches. The baseline LiDAR approach, exploiting its high-precision distance measurement capabilities, demonstrates consistently effective environment exploration, successfully visiting all 22 sectors on 33 occasions. Our $\pi_{collective}^{CEM}$ controller significantly surpasses this milestone, achieving full 22-sector coverage in 55 of the 100 tests. However, the CEM controller registers a slightly lower average number of visited sectors compared to the LiDAR baseline: 20.38 versus 20.91. As evidenced in Table \ref{tab:quantitative}, the $\pi_{collective}^{CMA-ES}$ controller exhibits a more conservative performance profile, visiting an average of 19.44 sectors and achieving full coverage only 10 times.

\begin{table*}[h!]
	\begin{minipage}{\textwidth}
		\begin{center}
			\caption{Average test results for sectors explored, reward points, percentage of time with inter-robot collisions, traveled distance, and energy consumption.}\label{tab:quantitative}
			\begin{tabular}{@{}ccccccc@{}}
				\hline
				\textbf{Method} & \textbf{Sectors} & \textbf{Reward} & \textbf{Collisions (\%)} & \textbf{Distance (m)} & \textbf{Consumption (Ws)} \\
				\hline
				\hline
				LiDAR & \textbf{20.91} & \textbf{5827.51} & 4.68 & 162.05 & 3341.99\\
				CEM & 20.38 & 4673.10 & 2.69 & \textbf{166.39} & 2316.93\\
				CMA-ES & 19.44 & 4587.92 & \textbf{2.41} & 157.58 & \textbf{1987.14}\\
				\hline
			\end{tabular}
		\end{center}
	\end{minipage}
\end{table*}

In terms of accumulated reward, both vision-based controllers register lower mean values than the LiDAR benchmark: 4673.10 for CEM and 4587.92 for CMA-ES versus 5827.51 for LiDAR. However, consistent with our architectural objectives, the vision-based approaches exhibit significantly superior energy efficiency. The $\pi_{collective}^{CEM}$ controller consumes an average of 2316.93 Ws, and the $\pi_{collective}^{CMA-ES}$ controller consumes 1987.14 Ws; both metrics are substantially lower than the LiDAR approach's 3341.99 Ws. This energy advantage is crucial for real-world swarm robotics deployment.

Using the $\pi_{collective}^{CMA-ES}$ controller, the robots' visual sensors are responsible for 10.02\% of the total energy consumption per episode. In contrast, in the case of the baseline method, the consumption of LiDAR devices represents 39.47\% of the total. Additionally, with the $\pi_{collective}^{CMA-ES}$ controller, the motors of each mobile agent consume on average a lower percentage of the total compared to the CEM-trained controller: 89.98\% versus 91.40\%. This latter aspect helps to explain why the $\pi_{collective}^{CEM}$ controller facilitates the swarm to travel the greatest average distance, 166.39 meters. Thus, it marginally surpasses the LiDAR approach's 162.05 meters and clearly exceeds the CMA-ES controller, which allows our four robots to move an average of 157.58 meters per episode.

Both vision-based controllers demonstrate superior safety performance compared to the baseline, achieving markedly lower average inter-agent collision rates: 2.69\% for CEM and 2.41\% for CMA-ES, compared to 4.68\% for the LiDAR system. The higher inter-agent collision rate observed in the LiDAR system is likely attributable to inherent limitations in reliably identifying small, dynamically maneuvering objects due to scanning frequency constraints relative to the robots' velocity. This can generate temporarily unmonitored zones where collisions may occur.

While the mean values provide a primary indicator of exploration efficacy, assessing policy stability requires analyzing the variance in performance across independent runs. Derived from the distributions illustrated in Fig. \ref{fig:quantitative}, the standard deviation for the number of visited sectors is 3.05 for the $\pi_{collective}^{CEM}$ controller and 1.94 for the $\pi_{collective}^{CMA-ES}$ controller. In comparison, the classical LiDAR baseline exhibits a standard deviation of 1.10. 

As expected, the LiDAR-based navigation yields the highest stability due to its omnidirectional, high-precision geometric sensing. However, the $\pi_{collective}^{CEM}$ policy demonstrates highly competitive consistency, achieving nearly identical spatial coverage despite relying solely on low-cost, partially observable visual representations processed through a latent VAE bottleneck. This confirms that the learned evolutionary policies achieve robust macroscopic stability while significantly reducing the hardware complexity and energy footprint of the swarm.

\subsubsection{Qualitative Behavioral Analysis}\label{subsubsec:qualitative}

\begin{figure*}[h!]
	\centering
	\includegraphics[width=\textwidth]{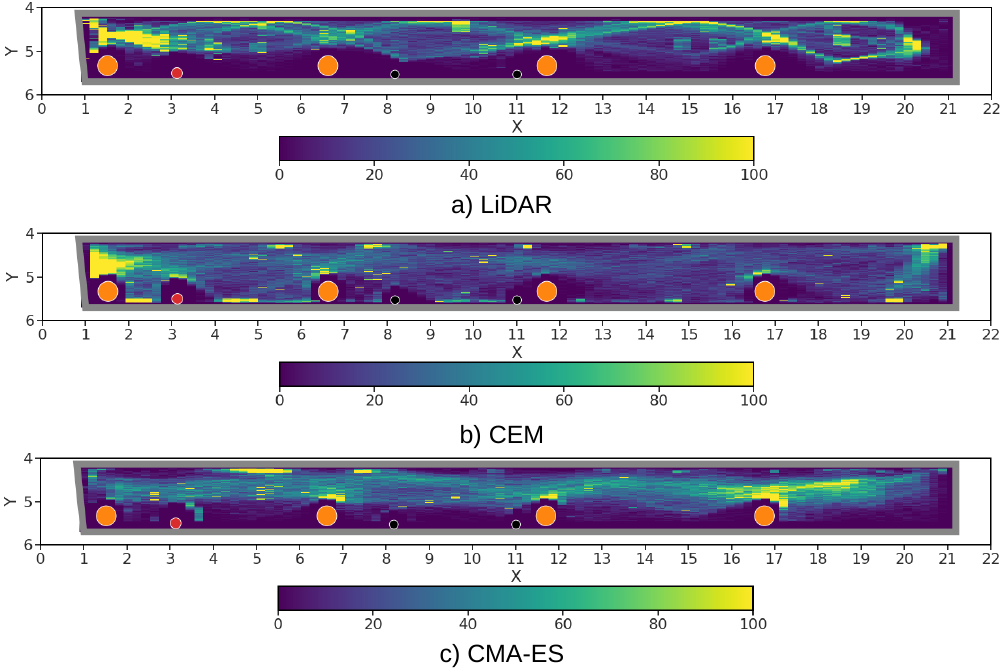}
	\caption{Navigational trajectory frequency heatmaps. This figure illustrates the number of times four robots traversed different positions in the simulated environment over 100 episodes, comparing the behaviors resulting from a) the LiDAR baseline, b) the policy obtained through our proposed cross-entropy method, and c) the policy generated using our proposed covariance matrix adaptation evolution strategy.}\label{fig:qualitative}
\end{figure*}

Fig. \ref{fig:qualitative} presents heatmaps illustrating the most frequently traversed positions in the corridor during the 100 validation tests, offering a qualitative assessment of the learned policies. By observing and representing these positions, we seek to identify specific collective behaviors, implicit coordination strategies between agents, emergent exploration patterns, and the system's adaptability to various conditions of the environment. The visualization of the swarm's movement patterns provides us with relevant information about the effectiveness of our policies, thus complementing the quantitative metrics.

When interpreting these spatial distributions, it is essential to consider the continuous wandering nature of the evaluated task. Given the finite navigable area of the simulated environment and the extended 5000-step duration of the validation episodes, the policies naturally and inevitably produce redundant exploration, where regions are visited repeatedly by the swarm members. Consequently, high-density traversal areas should not be misconstrued as an algorithmic flaw or an inability to escape local optima. Rather, this redundancy is the expected macroscopic outcome of a robust survival task, where the underlying priority is to maintain sustained, collision-free forward motion over prolonged periods.

In our heatmaps, environment positions traversed 100 or more times by the mobile robots are represented in yellow color at its maximum intensity. We established this specific limit to eliminate possible outliers from the point distribution shown in each heatmap, originated by a large number of collisions in certain areas of the environment. In contrast, the absolute absence of the yellow color indicates the corridor positions that were not traversed by the swarm on any occasion.

As observed in Fig. \ref{fig:qualitative}a), the LiDAR-based policy exhibits a clearly identifiable helical movement pattern. It also shows a pronounced tendency to follow the upper lateral wall and avoid the lower edge, where static obstacles are concentrated, suggesting effective contour following and obstacle avoidance behavior. These two particular behaviors are characteristic of systems based on LiDAR devices.

In contrast, the $\pi_{collective}^{CEM}$ controller, corresponding to Fig. \ref{fig:qualitative}b), promotes a significantly more dispersed exploration pattern. This policy strongly favors forward actions, which, while inducing a slightly increased collision risk, proves highly effective for maximizing rapid environmental coverage. This predisposition for advancement, along with the recovery of robots after suffering a collision, explains why this policy yielded the highest total distance traveled among all three evaluated approaches.

The $\pi_{collective}^{CMA-ES}$ controller, associated with Fig. \ref{fig:qualitative}c), exhibits an exploration pattern characterized by a lower concentration of robots at the corridor extremes. The visually apparent increase in turning actions directly contributes to its superior safety performance but concurrently results in a lower total distance traveled. The CMA-ES algorithm's dynamic covariance matrix adaptation mechanism, which is particularly efficient at exploring the solution space \citep{Yu_2015}, effectively optimized the pre-trained collision-free navigation weights for conservative safety without heavily biasing toward raw forward velocity.

\subsubsection{Ablation Study: Impact of Collision Termination Rules on Exploration}\label{subsubsec:ablation}

To rigorously evaluate the sensitivity of the learned policies to environmental penalties, an ablation study was conducted. Specifically, we analyzed how the relaxation of the strict collision termination rule during the evaluation phase impacts the final exploration metrics.

During the evolutionary training phase, any collision resulted in the immediate termination of the episode. However, during standard evaluation, agents were permitted to recover and continue exploring after a collision, albeit with a severe discrete reward penalty. To isolate the behavioral impact of the termination rules, the neural controllers were not retrained for this ablation study. The evaluations were conducted using the exact policies that had already converged during the initial training phase. Table \ref{tab:ablation} compares the macroscopic performance of both vision-based policies, $\pi_{collective}^{CEM}$ and $\pi_{collective}^{CMA-ES}$, under evaluation conditions with and without the ability to recover from collisions.

\begin{table}[h!]
	\begin{minipage}{\columnwidth}
		\begin{center}
			\caption{Average test results for sectors explored and reward points obtained by the vision-based controllers with and without the ability to recover and continue navigation after a collision.}\label{tab:ablation}
			\begin{tabular}{@{}cccc@{}}
				\hline
				\textbf{Evaluation Rule} & \textbf{Method} & \textbf{Sectors} & \textbf{Reward}\\
				\hline
				\hline
				With collision recovery & CEM & \textbf{20.38} & \textbf{4673.10}\\
				With collision recovery & CMA-ES & 19.44 & 4587.92\\
				\hline
				Without collision recovery & CEM & 11.48 & 752.35\\
				Without collision recovery & CMA-ES & \textbf{13.61} & \textbf{1427.84}\\
				\hline
			\end{tabular}
		\end{center}
	\end{minipage}
\end{table}

The empirical results from this ablation study reveal profound structural differences in the emergent behaviors generated by our two evolution strategies. The $\pi_{collective}^{CEM}$ controller's tendency to aggressively predict forward commands translates into extensive spatial coverage, but it concurrently assumes a higher navigational risk. When agents are permanently halted upon collision, the performance of the CEM policy degrades drastically, plummeting from 20.38 to a mere 11.48 mean sectors explored. The aggressive forward bias proves detrimental under strict zero-tolerance penalty conditions, as colliding robots become permanent static obstacles for the remainder of the episode, severing the swarm's exploratory momentum.

Conversely, the $\pi_{collective}^{CMA-ES}$ controller demonstrates significantly higher intrinsic robustness under strict evaluation rules. As indicated by its qualitative heatmap (\textit{i.e.}, Fig. \ref{fig:qualitative}c)), this policy favors a more conservative navigational profile, frequently executing turning commands to avoid close-quarters maneuvering near fixed objects. Because of this superior inherent obstacle avoidance, the CMA-ES policy suffers a markedly less severe degradation when collision recovery is disabled, achieving a mean of 13.61 explored sectors across 100 trials.

Consequently, under strict termination constraints, the $\pi_{collective}^{CMA-ES}$ controller actively outperforms the $\pi_{collective}^{CEM}$ controller. This highlights that while our cross-entropy method converged upon a high-yield, high-risk exploration strategy dependent on the system's tolerance for minor collisions, our covariance matrix adaptation evolution strategy successfully optimized for genuine, conservative safety. This confirms that the learned behaviors are highly sensitive to the reward architecture and that our hybrid evolutionary framework is capable of discovering diverse, valid solutions along the exploration-safety Pareto front depending on the applied constraints.

\subsection{Statistical Analysis of Exploration Effectiveness}\label{subsec:statistical}

To rigorously validate the quantitative results and provide a robust assessment of overall exploration efficacy, we conducted a statistical analysis focusing on the number of sectors visited, which serves as the primary indicator of spatial coverage. We used Analysis of Variance (ANOVA) followed by Tukey's Honestly Significant Difference (HSD) \textit{post hoc} test \citep{Nanda_2021}. The Tukey HSD test is appropriate for simultaneous pairwise comparisons and determines the statistical significance of differences between multiple group means using an adjusted p-value. This value incorporates a correction that controls the family-wise error rate at a specified significance level $\alpha$, which is essential when performing multiple simultaneous comparisons.

We set the significance level $\alpha = 0.05$ and a 95\% Confidence Interval (CI) for the difference between the means of each pair, represented by lower and upper bounds. If the confidence interval does not include zero, one can determine that the difference between the compared means is statistically significant. The same conclusion can be reached if the adjusted p-value obtained is below the established significance level \citep{Nanda_2021}. The results of our Tukey HSD test are presented in Table \ref{tab:tukey}.
	
\begin{table*}[h!]
	\begin{minipage}{\textwidth}
		\begin{center}
			\caption{Results of the Tukey HSD test for comparing means of sectors explored by the three collective autonomous navigation controllers evaluated.}\label{tab:tukey}
			\begin{tabular}{@{}ccccc@{}}
				\hline
				\textbf{Comparison} & \textbf{Difference} & \textbf{p-Value} & \textbf{Confidence Interval (95\%)} & \textbf{Significant} \\
				\hline
				\hline
				CEM vs LiDAR & -0.53 & 0.2008 & [-1.2575, 0.1975] & No\\
				CMA-ES vs LiDAR & -1.47 & 0.0 & [-2.1975, -0.7425] & Yes\\
				CMA-ES vs CEM & -0.94 & 0.0072 & [-1.6675, -0.2125] & Yes\\
				\hline
			\end{tabular}
		\end{center}
	\end{minipage}
\end{table*}

The analysis yielded the following critical findings:

\begin{itemize}
	
	\item \textbf{CEM vs LiDAR}: The adjusted p-value of $0.2008$ exceeds the threshold $\alpha = 0.05$, and the confidence interval $[-1.2575, 0.1975]$ includes zero. This confirms the impossibility of rejecting the null hypothesis of equality between means and indicates no statistically significant difference between both approaches in terms of exploratory capacity.
	
	\item \textbf{CMA-ES vs LiDAR}: The adjusted p-value is below $0.001$, and the confidence interval $[-2.1975, -0.7425]$ lies entirely in the negative domain. This demonstrates a statistically significant difference and confirms that the performance of the LiDAR approach significantly exceeds the $\pi_{collective}^{CMA-ES}$ controller by a mean of $1.47$ sectors.
	
	\item \textbf{CMA-ES vs CEM}: Similarly, a statistically significant difference exists between the $\pi_{collective}^{CEM}$ and $\pi_{collective}^{CMA-ES}$ controllers, as demonstrated by the adjusted p-value of $0.0072$. The confidence interval $[-1.6675, -0.2125]$ confirms that the $\pi_{collective}^{CEM}$ controller's performance is significantly superior, with a mean difference of $0.94$ sectors explored.
	
\end{itemize}

These statistical findings corroborate the qualitative and quantitative observations: the $\pi_{collective}^{CEM}$ controller achieves performance statistically comparable to the conventional LiDAR-based system in terms of exploratory capacity. This result is highly relevant as it validates the viability of using low-cost, energy-efficient monocular vision systems for complex autonomous navigation tasks, offering a critical advantage over the heavier and more expensive distance sensors typically employed.

\section{Analysis of Swarm Robotics Properties}\label{sec:analysis}

This section undertakes a systematic analysis of three essential properties inherent to the collective behaviors generated by the proposed swarm robotic controllers, $\pi_{collective}^{CEM}$ and $\pi_{collective}^{CMA-ES}$: robustness, flexibility, and scalability. The rigorous evaluation of these properties is paramount, as they determine the viability and effectiveness of deploying such controllers within physical swarm robotic systems operating in real-world conditions.

It is imperative to emphasize that the final policies evaluated throughout this experimental phase are deployed exactly as they were obtained at the culmination of the evolutionary training process detailed in Section \ref{sec:experiments}. To strictly evaluate the true generalization bounds and inherent properties of the macroscopic model, these controllers undergo a rigorous zero-shot transfer across all experimental scenarios. Specifically, the policies are not retrained, fine-tuned, or adapted in any capacity to accommodate the novel evaluation conditions. This direct deployment demonstrates that the learned behavioral parameters are intrinsically robust, flexible, and scalable, validating the efficiency of the proposed approach for practical implementations without requiring computationally expensive retraining phases.

Furthermore, a methodological distinction must be clarified regarding the reporting of the evaluation metrics in this section compared to the previous one. Having established the baseline policy stability and variance profiles in Section \ref{sec:experiments}, this section shifts the analytical focus toward macroscopic performance trends under environmental stress. Specifically, the following evaluations of robustness, flexibility, and scalability utilize mean performance metrics across 100 trials to quantify the expected systemic degradation or scaling capacity when the swarm is subjected to novel conditions.

Consistent with the comparative analysis presented in Section \ref{sec:experiments}, we utilize the previously established LiDAR-based collective autonomous navigation controller as a benchmark reference. The primary experimental environment remains the synthetic indoor corridor detailed in Fig. \ref{fig:environment}. However, to rigorously assess the controllers' adaptability, we introduce three additional synthetic environments, modeled using our high-fidelity physics simulator. Each new environment incorporates significant structural modifications designed to challenge the swarm's adaptive capabilities. These environments, representing distinct interior spaces commonly found in real buildings, introduce progressively complex variations from the original training corridor and feature varying numbers of navigable sectors, as outlined in Table \ref{tab:environments}.

\begin{table}[h!]
	\begin{minipage}{\columnwidth}
		\begin{center}
			\caption{Synthetic environments used to evaluate the robustness, flexibility, and scalability of the developed robotic swarm.}\label{tab:environments}
			\begin{tabular}{@{}ccc@{}}
				\hline
				\textbf{Identifier} & \textbf{Name} & \textbf{Sectors} \\
				\hline
				\hline
				(0) & Original corridor & 22\\
				(1) & Corridor with corner & 26\\
				(2) & Corridor with intersection & 30\\
				(3) & Corridor with passage & 33\\
				\hline
			\end{tabular}
		\end{center}
	\end{minipage}
\end{table}

Regarding the topological complexity of the evaluation scenarios, the original training environment and the three novel alternative layouts introduced for flexibility testing systematically challenge the swarm's navigation capabilities. These environments inherently feature narrow passages, structural bottlenecks, and severe visual occlusions (\textit{e.g.}, sharp corners and intersecting corridors). Successfully navigating these features validates the robustness and generalization capability of the proposed exploration method under restricted spatial conditions.

The validation of robustness, flexibility, and scalability necessitates extensive experimentation under diverse conditions, involving multiple agents and environmental variations. Conducting such large-scale experiments with physical multi-robot systems presents significant practical challenges. These challenges include substantial costs in terms of time and resources, inherent risks of equipment damage and environmental harm, difficulties in ensuring exact reproducibility due to uncontrolled environmental factors, and logistical complexities associated with managing and coordinating numerous physical agents, particularly when testing scalability with large swarm sizes.

Consequently, employing realistic physics simulations offers a highly advantageous alternative for this phase of evaluation. Simulation allows for risk-free exploration of potentially hazardous behaviors, provides significant temporal and economic efficiencies enabling a far greater number of tests, and ensures high reproducibility through precise control over experimental conditions. Furthermore, it facilitates the rapid generation and modification of diverse environmental scenarios, enables straightforward scalability testing with varying numbers of agents, and permits the systematic examination of system behavior under a breadth of configurations and controlled perturbations that would be impracticable in physical settings. This simulation-based validation is particularly valuable for rigorously assessing the fundamental properties required for successful real-world deployment in swarm robotics.

Our analysis is structured into three distinct evaluation processes:

\begin{itemize}
	
	\item A robustness evaluation process, detailed in Subsection \ref{subsec:robustness}, assessing the system's resilience to sensory perturbations, simulating common real-world sensor noise and variability.
	
	\item A flexibility evaluation process, described in Subsection \ref{subsec:flexibility}, examining the system's capacity to adapt and perform effectively across various novel and unknown environments.
	
	\item A scalability evaluation process, presented in Subsection \ref{subsec:scalability}, analyzing the system's performance characteristics as the number of constituent agents is varied.
	
\end{itemize}

The insights gained from these evaluations are crucial for understanding the controllers' potential and limitations regarding practical, real-world applicability.

\subsection{Robustness}\label{subsec:robustness}

The evaluation of robustness is a cornerstone in the development of swarm robotic models intended for operation in inherently uncertain and dynamic real-world environments. Robustness, within the context of swarm robotics, refers to the system's inherent capacity to maintain stable and effective performance despite internal failures (\textit{e.g.}, agent loss) or external perturbations \citep{Sahin_2005}. Various methodologies can be employed to assess this critical property, including the simulation of agent failures, the modification of environmental parameters, or, as adopted in this study, the introduction of noise into the agents' sensory observations.

In physical robotic swarm deployments based on low-cost hardware, sensory perception is inherently flawed. Consequently, evaluating the robustness of the learned navigation policies is not merely a test of basic collision avoidance, but a strict assessment of the framework's resilience against the intrinsic uncertainty of reinforcement learning under partial observability. Operating exclusively on highly compressed visual latent representations, the agents lack global state information.

To systematically evaluate how noise in visual observations affects final performance, we subjected the zero-shot transferred controllers to varying levels of Gaussian noise applied directly to their perceptual inputs. This controlled injection of sensory noise acts as a preliminary domain randomization technique \citep{Muratore_2022}. By deliberately corrupting the visual data, we emulate the perceptual uncertainty typical of physical camera sensors. This approach quantifies the impact of partial observability on policy optimality, demonstrating whether the agents can sustain their emergent collective behaviors when the reliability of their environmental perception is systematically degraded.

Furthermore, the injection of Gaussian noise holds particular relevance for controllers reliant on monocular vision, as it effectively simulates the types of sensory variability commonly encountered in real-world conditions. Such variability can arise from fluctuations in ambient illumination, partial occlusions of the visual field, or inherent imperfections and limitations of the camera sensors themselves. In the broader context of robotic visual navigation, uncertainty originating from sensor data frequently manifests as noise, potentially caused by factors including random environmental perturbations, inadequate light levels, suboptimal device operating temperatures \citep{Yogesh_2017}, electromagnetic interference, or even minor vibrations and unintentional movements of the robotic platform. A controller's ability to maintain performance despite such noisy input is a strong indicator of its potential for reliable real-world operation.

In our robustness evaluation protocol, we systematically introduce additive Gaussian noise to the input sensory data at three distinct levels: 5\%, 10\%, and 15\% of the signal range. For the vision-based controllers, $\pi_{collective}^{CEM}$ and $\pi_{collective}^{CMA-ES}$, this noise is applied directly to the pixels of the input RGB images prior to processing by the VAE encoder. For the LiDAR-based baseline controller, equivalent noise percentages are injected into the distance measurements provided by the simulated laser scanner. This graduated introduction of noise allows us to meticulously assess whether, and to what extent, the performance of each controller degrades as the level of sensory perturbation increases, providing critical insights into their resilience and suitability for deployment in unpredictable real-world settings.

\begin{figure*}[h!]
	\centering
	\includegraphics[width=\textwidth]{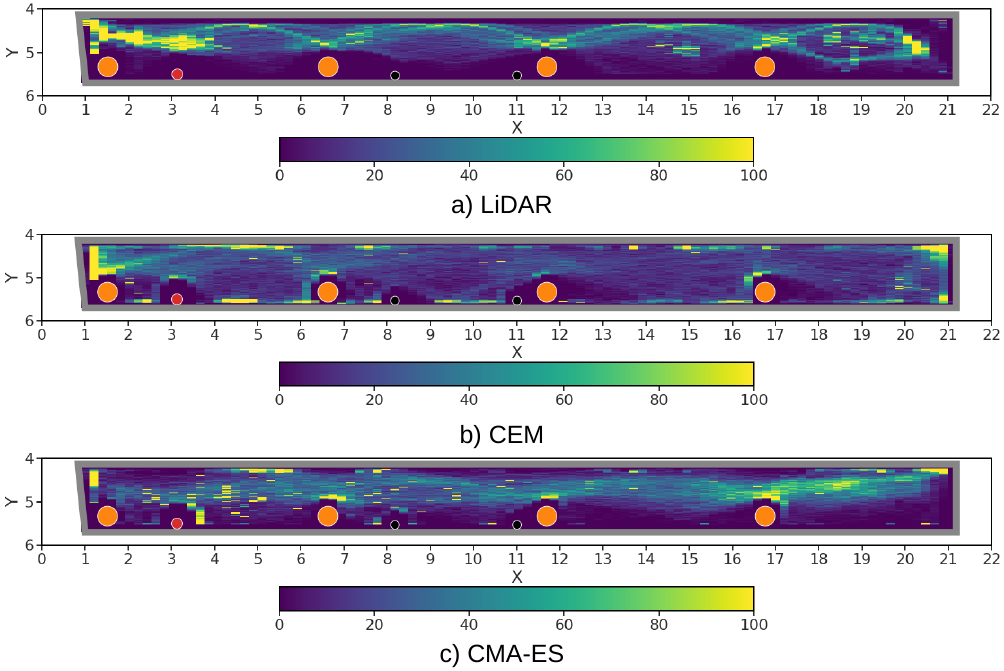}
	\caption{Navigational trajectory frequency heatmaps with 5\% noise. This figure illustrates the number of times four robots traversed different positions in the simulated environment (0) over 100 episodes with 5\% Gaussian noise added to sensory inputs, comparing behaviors from a) the LiDAR baseline, b) the proposed CEM-trained policy, and c) the proposed CMA-ES-trained policy.}\label{fig:robustness_5}
\end{figure*}

\begin{figure*}[h!]
	\centering
	\includegraphics[width=\textwidth]{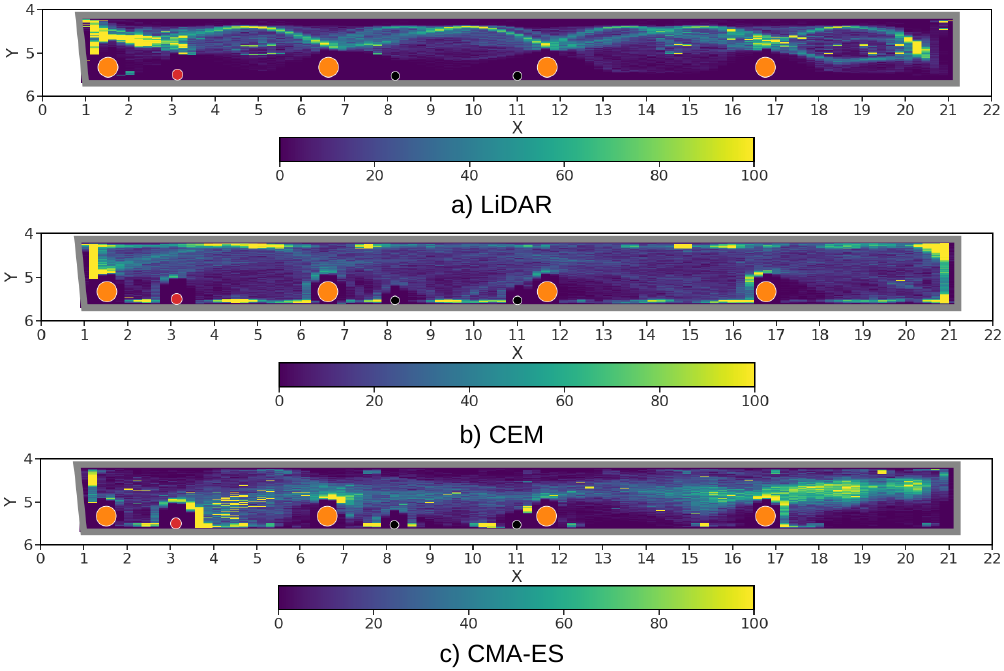}
	\caption{Navigational trajectory frequency heatmaps with 10\% noise. This figure illustrates the number of times four robots traversed different positions in the simulated environment (0) over 100 episodes with 10\% Gaussian noise added to sensory inputs, comparing behaviors from a) the LiDAR baseline, b) the proposed CEM-trained policy, and c) the proposed CMA-ES-trained policy.}\label{fig:robustness_10}
\end{figure*}

\begin{figure*}[h!]
	\centering
	\includegraphics[width=\textwidth]{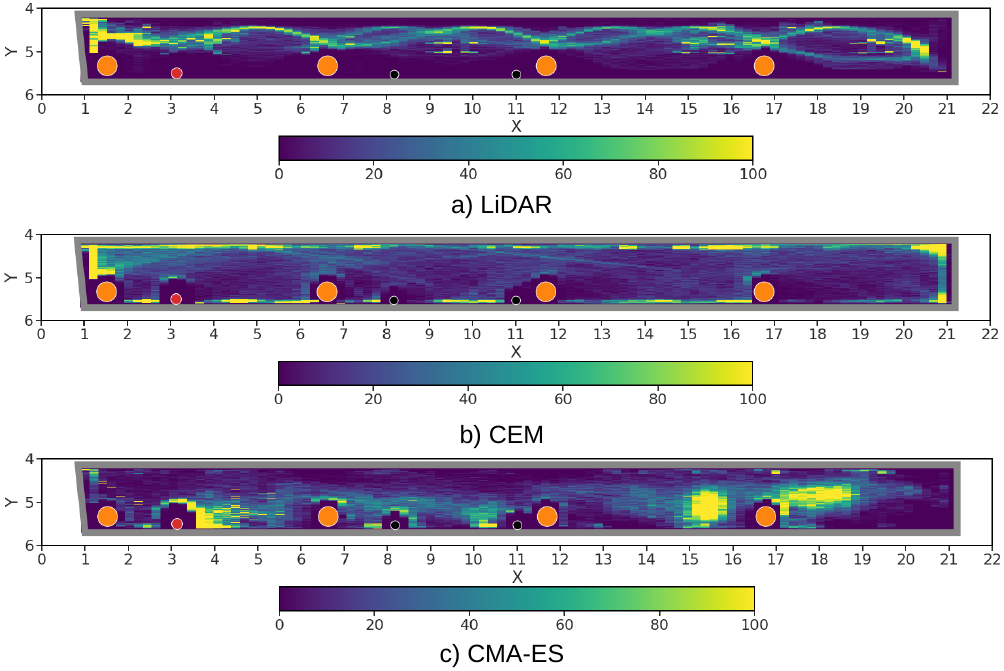}
	\caption{Navigational trajectory frequency heatmaps with 15\% noise. This figure illustrates the number of times four robots traversed different positions in the simulated environment (0) over 100 episodes with 15\% Gaussian noise added to sensory inputs, comparing behaviors from a) the LiDAR baseline, b) the proposed CEM-trained policy, and c) the proposed CMA-ES-trained policy.}\label{fig:robustness_15}
\end{figure*}

The qualitative impact of noise injection is visualized through heatmaps representing the positions traversed by the swarm robots over 100 consecutive MARL episodes for each noise level. A comparative analysis reveals distinct responses among the controllers. In the case of the CEM-trained controller, the degradation in performance is marginal, even at the highest noise level. As illustrated in Fig. \ref{fig:robustness_5}b), with 5\% noise, the exploration pattern remains remarkably similar to the noise-free baseline shown in Fig. \ref{fig:qualitative}b), exhibiting relatively uniform coverage of the corridor. Even with 10\% and 15\% noise, while subtle changes like a slightly higher concentration along the upper edge emerge, the overall exploration pattern remains consistent and effective, as evidenced in Fig. \ref{fig:robustness_10}b) and Fig. \ref{fig:robustness_15}b). Similarly, the LiDAR-based controller demonstrates considerable robustness, maintaining its characteristic helical movement pattern and contour-following behavior with minimal perceptible degradation across the tested noise levels.

In stark contrast, the performance degradation is notably evident for the controller trained via our CMA-ES algorithm. As observed in Fig. \ref{fig:robustness_5}c), at 5\% noise, the pattern resembles the noise-free case shown in Fig. \ref{fig:qualitative}c), suggesting tolerance to low-level perturbations. However, at 10\% noise, corresponding to Fig. \ref{fig:robustness_10}c), discernible differences appear, with agents showing a tendency to traverse positions nearer the corridor extremes and the lower edge, a significant deviation from the noise-free behavior. As evidenced in Fig. \ref{fig:robustness_15}c), this tendency becomes more pronounced at 15\% noise, indicating a substantial alteration of the emergent behavioral pattern. This shift towards navigating regions with higher concentrations of fixed obstacles inherently increases the probability of collisions and signifies a considerable loss of performance effectiveness under higher sensory uncertainty.

The quantitative results, summarized in Table \ref{tab:robustness}, align coherently with these qualitative observations. The $\pi_{collective}^{CEM}$ controller demonstrates remarkable resilience. At 5\% noise, it achieves an average accumulated reward of 4221.22 points and explores a mean of 20.61 sectors, results highly comparable to its noise-free performance: 4673.10 points and 20.38 sectors, from Table \ref{tab:quantitative}. Even at higher noise levels of 10\% and 15\%, while there is a discernible decrease in average reward and a slight increase in inter-robot collision time percentage, the controller maintains strong exploration capability, visiting an average of 20.66 and 19.97 sectors, respectively. Notably, in these higher noise scenarios, it even surpasses the LiDAR baseline in terms of average sectors explored. 

Additionally, the $\pi_{collective}^{CEM}$ controller exhibits consistent stability in the total distance traveled, maintaining mean values around 160 meters across all noise levels, consistently outperforming the baseline approach in this metric. This stability likely stems from its inherent tendency to favor forward actions, a bias that remains effective for exploration even when visual input contains inaccuracies, without leading to a catastrophic increase in collisions with dynamic obstacles.

Conversely, the $\pi_{collective}^{CMA-ES}$ controller shows a progressive and significant decline in performance as noise levels increase. Its average accumulated reward drops from 3142.97 at 5\% noise to a mere 535.67 at 15\% noise, with the average number of explored sectors decreasing from 18.65 to 9.81 over the same range. This controller exhibits particularly pronounced sensitivity in terms of distance traveled, which plummets from 136.91 meters at 5\% noise to 89.68 meters at 15\% noise. This behavior reinforces the qualitative observation that sensory uncertainty amplifies its natural inclination towards turning actions, leading to less effective, sometimes circular or serpentine, trajectories that severely limit spatial exploration. Concurrently, the average inter-robot collision rate deteriorates, rising from 1.99\% at 5\% noise to 5.26\% at 15\% noise.

While the $\pi_{collective}^{CMA-ES}$ controller consistently maintains the highest energy efficiency across all scenarios, which is directly correlated with its lower average traveled distance, its overall effectiveness is substantially inferior to the $\pi_{collective}^{CEM}$ controller. This is notably evident under high noise conditions where excessive turning can lead to robots rotating in place. This suggests a perturbation threshold beyond which its navigational capabilities become significantly compromised.

The LiDAR-based approach, while robust in maintaining its general exploration pattern, proves to be the least efficient and demonstrates increasing vulnerability regarding safety as noise intensifies. Its average inter-robot collision rate escalates from 4.10\% at 5\% noise to 6.25\% at 15\% noise. This susceptibility highlights the critical dependence of LiDAR on precise distance measurements for avoiding obstacles, particularly dynamic ones. Sensory noise appears to exacerbate its inherent limitations in reliably detecting other small, potentially fast-moving swarm members, especially those moving in trajectories perpendicular to the sensor beam or maneuvering abruptly.

\begin{table*}[h!]
	\begin{minipage}{\textwidth}
		\begin{center}
			\caption{Average robustness test results across different noise levels for sectors explored, reward points, percentage of time with inter-robot collisions, traveled distance, and energy consumption.}\label{tab:robustness}
			\setlength{\tabcolsep}{3pt}
			\begin{tabular}{@{}ccccccc@{}}
				\hline
				\textbf{Noise} & \textbf{Method} & \textbf{Sectors} & \textbf{Reward} & \textbf{Collisions (\%)} & \textbf{Distance (m)} & \textbf{Consumption (Ws)} \\
				\hline
				\hline
				5\% & LiDAR & \textbf{20.76} & \textbf{5305.83} & 4.10 & 154.82 & 3367.04\\
				5\% & CEM & 20.61 & 4221.22 & 2.42 & \textbf{165.15} & 2862.62\\
				5\% & CMA-ES & 18.65 & 3142.97 & \textbf{1.99} & 136.91 & \textbf{2113.32}\\
				\hline
				10\% & LiDAR & 19.76 & \textbf{4174.42} & 5.26 & 133.09 & 3614.75\\
				10\% & CEM & \textbf{20.66} & 3723.24 & \textbf{3.01} & \textbf{159.79} & 2367.03\\
				10\% & CMA-ES & 15.58 & 1631.54 & 3.09 & 109.89 & \textbf{1979.73}\\
				\hline
				15\% & LiDAR & 19.63 & 4016.93 & 6.25 & 125.57 & 3320.41\\
				15\% & CEM & \textbf{19.97} & \textbf{4099.51} & \textbf{2.94} & \textbf{166.78} & 2525.79\\
				15\% & CMA-ES & 9.81 & 535.67 & 5.26 & 89.68 & \textbf{1780.48}\\
				\hline
			\end{tabular}
		\end{center}
	\end{minipage}
\end{table*}

In summary, the robustness analysis underscores the superior resilience of the $\pi_{collective}^{CEM}$ controller to sensory noise compared to both the $\pi_{collective}^{CMA-ES}$ controller and, significantly, the LiDAR baseline in terms of safety, specifically regarding inter-robot collisions, and distance traveled. Its ability to maintain effective exploration and relatively low collision rates despite substantial input perturbations makes it a particularly promising candidate for real-world deployment where such conditions are commonplace.

Beyond the empirical resilience to visual perturbations, ensuring successful sim-to-real transferability requires addressing hardware-specific challenges, primarily onboard inference latency and actuation uncertainty. The proposed architecture mitigates inference latency through its extreme dimensionality reduction; by compressing high-dimensional pixel data into a 32-element latent vector via the VAE, the subsequent control inference relies solely on an ultra-compact, lightweight MLP. This enables high-frequency control loops even on resource-constrained microprocessors without specialized AI accelerators.

Moreover, the system's robustness to actuation uncertainty, such as unmodeled motor friction, asymmetrical wheel slip, or kinematic drift, is intrinsically supported by the memoryless, reactive nature of the MLP controller. Because the policy maps instantaneous visual feedback directly to kinematic commands frame-by-frame, without relying on complex historical states or dead-reckoning odometry, the robotic agents can dynamically and continuously correct for mechanical execution errors in real time.

\subsection{Flexibility}\label{subsec:flexibility}

The evaluation of flexibility in swarm robotic models is a critical step toward developing collective, adaptable, and effective systems suitable for real-world deployment. This property determines a swarm's capacity to operate effectively in dynamic and unknown environments, a fundamental prerequisite for practical applications where scenarios are often highly variable and unpredictable. One of the most effective methodologies for assessing flexibility involves exposing the swarm to diverse and challenging environments that differ significantly from those encountered during the training phase.

\begin{figure*}[h!]
	\centering
	\includegraphics[height=170mm, width=\textwidth]{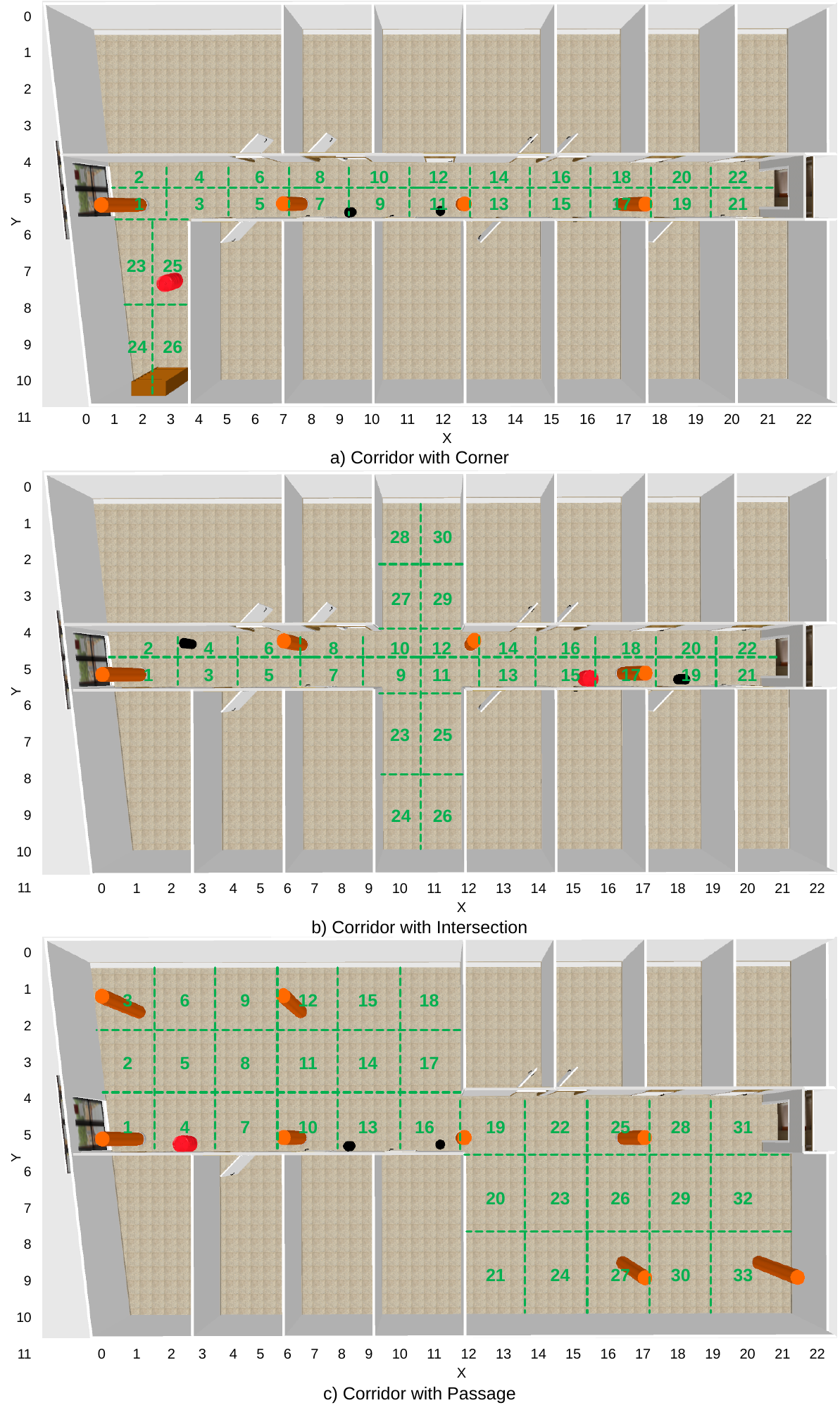}
	\caption{New synthetic environments for flexibility analysis. This figure illustrates the division of the three novel simulated environments into discrete sectors (in green) with their corresponding indices for quantitative evaluation. The coordinate axes correspond to three distinct interior spaces: a) a corridor with a corner, b) a corridor with an intersection, and c) a corridor with a narrow passage.}\label{fig:environments}
\end{figure*}

\begin{figure*}[h!]
	\centering
	\includegraphics[height=150mm, width=\textwidth]{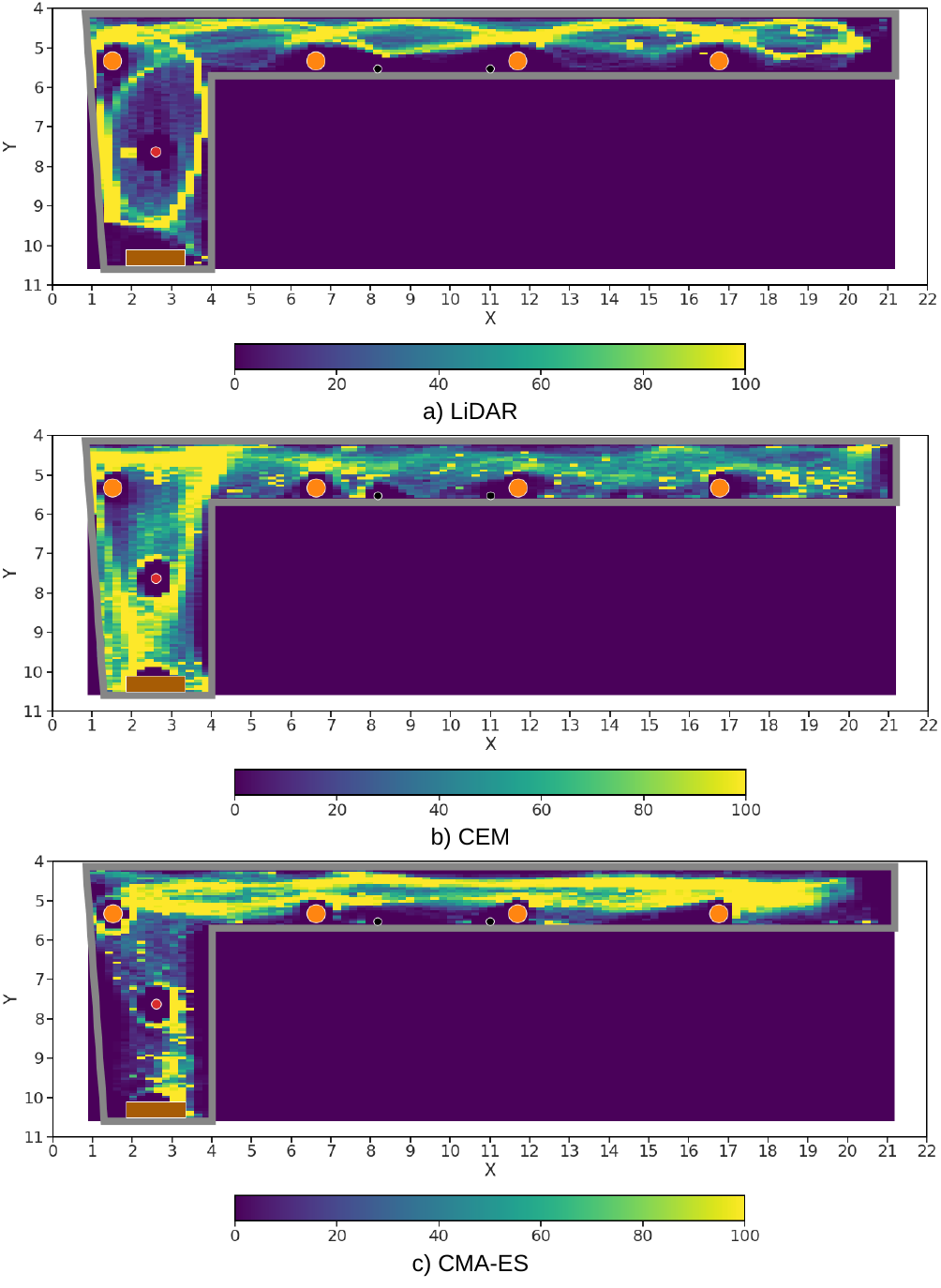}
	\caption{Navigational trajectory frequency heatmaps in environment (1). This figure illustrates the number of times four robots traversed different positions in the simulated corridor with a corner over 100 episodes. This flexibility assessment compares behaviors from: a) the LiDAR baseline, b) the proposed CEM-trained policy, and c) the proposed CMA-ES-trained policy.}\label{fig:flexibility_1}
\end{figure*}

Realistic physics simulators provide an invaluable and indispensable tool for this purpose. They facilitate the rapid generation and modification of a wide variety of environmental conditions, enabling a more exhaustive and systematic evaluation of controller flexibility than would be logistically feasible or economically viable with physical hardware. For vision-based controllers specifically, this simulation-based approach permits a rigorous examination of the system's adaptability to substantial structural and compositional changes (\textit{e.g.}, new layouts and novel obstacles) without altering the fundamental nature of the sensory input data. When the structure, number, or position of objects diverge from the training scenario, the swarm must demonstrate high flexibility, forcing individuals to adopt strategies that may differ from those initially learned.

In our investigation, we modeled three distinct synthetic environments to validate the flexibility of the controllers, which were previously trained and evaluated exclusively in the original corridor, shown in Fig. \ref{fig:environment}. These new testbeds correspond to indoor corridors (1), (2), and (3), as presented in Table \ref{tab:environments} and graphically represented in Fig. \ref{fig:environments}.

\begin{figure*}[h!]
	\centering
	\includegraphics[height=175mm, width=\textwidth]{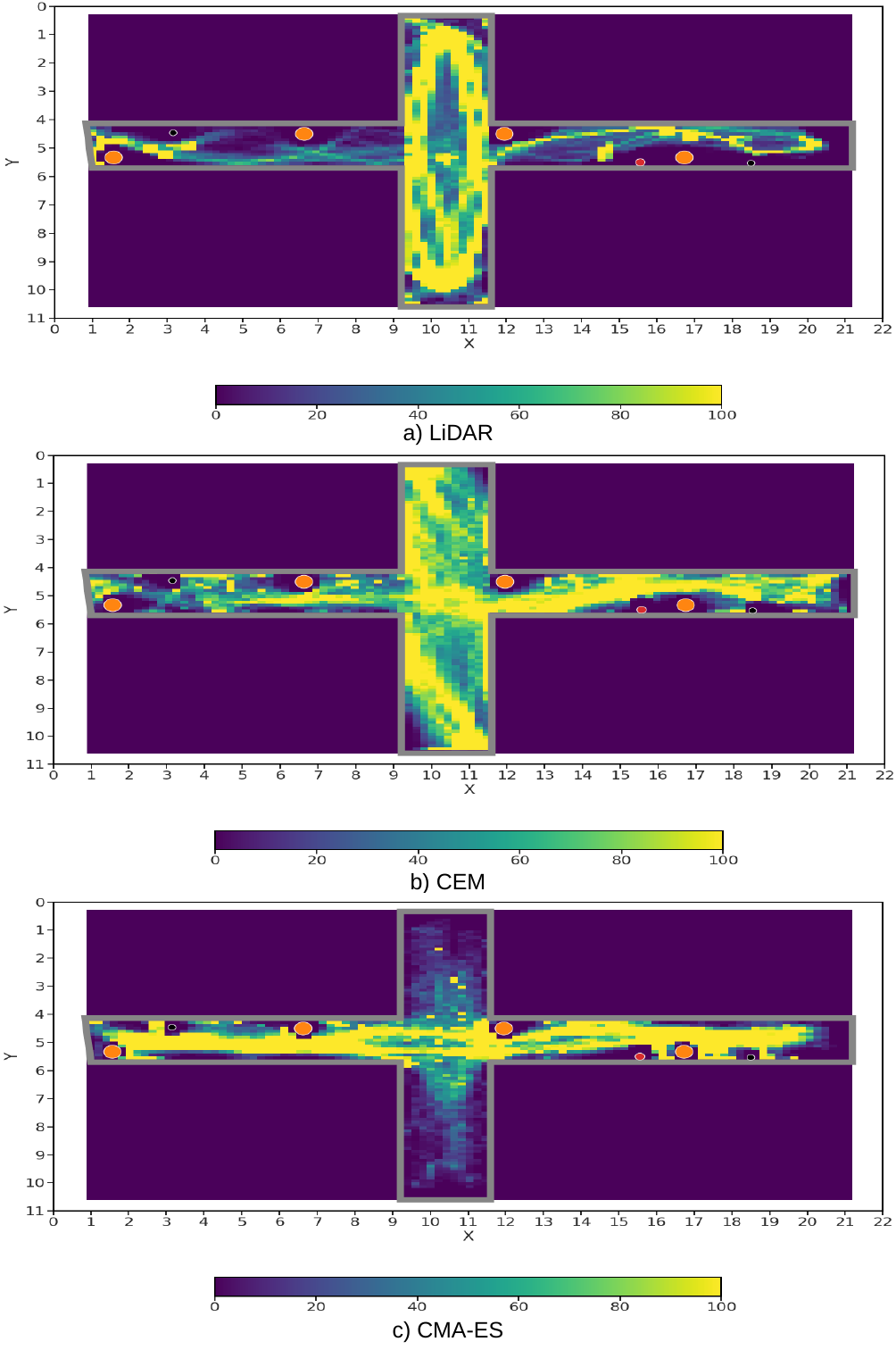}
	\caption{Navigational trajectory frequency heatmaps in environment (2). This figure illustrates the number of times four robots traversed different positions in the simulated corridor with an intersection over 100 episodes. This flexibility assessment compares behaviors from: a) the LiDAR baseline, b) the proposed CEM-trained policy, and c) the proposed CMA-ES-trained policy.}\label{fig:flexibility_2}
\end{figure*}

\begin{figure*}[h!]
	\centering
	\includegraphics[height=175mm, width=\textwidth]{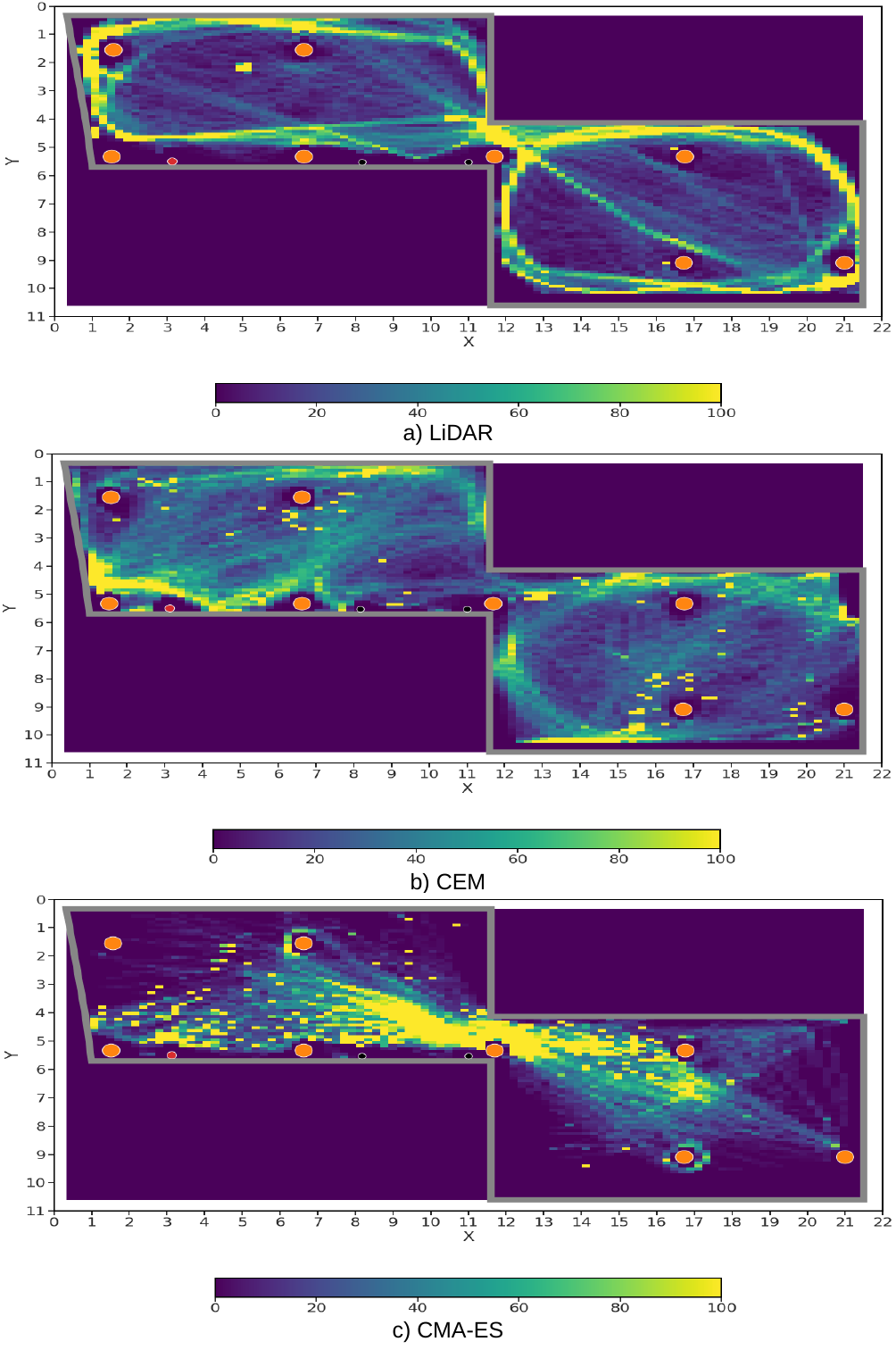}
	\caption{Navigational trajectory frequency heatmaps in environment (3). This figure illustrates the number of times four robots traversed different positions in the simulated corridor with a passage over 100 episodes. This flexibility assessment compares behaviors from: a) the LiDAR baseline, b) the proposed CEM-trained policy, and c) the proposed CMA-ES-trained policy.}\label{fig:flexibility_3}
\end{figure*}

Analysis of the navigational trajectory heatmaps for each environment reveals distinctive patterns in the exploration conducted by both the LiDAR-based multi-robot system and our vision-based swarm robotic controllers, $\pi_{collective}^{CEM}$ and $\pi_{collective}^{CMA-ES}$. In the corridor with a corner, as shown in Fig. \ref{fig:flexibility_1}, all three tested controllers promote relatively broad coverage of the environment. However, the $\pi_{collective}^{CEM}$ controller exhibits a distinct advantage in exploring peripheral areas. This advantage is particularly pronounced in the novel secondary corridor (\textit{i.e.}, the navigable sectors not present in the training environment), especially in the positions surrounding a brown bookshelf, an object category not encountered during the training phase. In contrast, the positions most frequently traversed by robots using the $\pi_{collective}^{CMA-ES}$ controller tend to concentrate in the central areas of the main corridor.

Fig. \ref{fig:flexibility_2} illustrates more marked differences between the behaviors deployed by the three controllers in the corridor with an intersection. The $\pi_{collective}^{CMA-ES}$ controller favors navigation along the main corridor, as evidenced by the higher quantity of traversals represented by maximum-intensity yellow in these regions. Conversely, the $\pi_{collective}^{CEM}$ controller tends to concentrate its navigational activity in the secondary, intersecting corridor, resulting in a less exhaustive exploration of the main corridor. This tendency is even more pronounced with the LiDAR-based controller, which promotes much more exhaustive coverage of the secondary corridor, often at the expense of the main one. Notably, this controller continues to exhibit the helical movement pattern that was clearly identifiable in the initial tests conducted in the original environment (0).

The corridor with a narrow central passage presents a significant navigational challenge for all three controllers. As demonstrated in Fig. \ref{fig:flexibility_3}, the $\pi_{collective}^{CEM}$ controller promotes a notably more homogeneous spatial coverage, facilitating the exploration of more distant and hard-to-access areas. The LiDAR-based controller shows similar behavior while additionally enabling highly effective contour-tracking of the interior space. In contrast, the $\pi_{collective}^{CMA-ES}$ controller concentrates its most frequently traversed positions around the central passage itself. It is noteworthy that all three deployed behaviors demonstrate effective collision-free navigation through this restrictive passage, validating their capability to handle narrow chokepoints.

From a quantitative perspective, the results presented in Table \ref{tab:flexibility} show that the controller trained with the CEM algorithm accumulates significantly higher average rewards compared to the $\pi_{collective}^{CMA-ES}$ controller across all three novel environments. The $\pi_{collective}^{CEM}$ controller's tendency to predict forward commands, which provide the highest reward points, makes it particularly well-suited for environments (2) and (3) with their larger navigable surfaces and open spaces. This predisposition is also responsible for the increased average distance traveled by the swarm in these two environments. However, this \textit{forward} bias can be counterproductive in areas with a high concentration of complex obstacles. This limitation becomes apparent in the four-cornered intersection of environment (2), where robots controlled by the $\pi_{collective}^{CEM}$ policy experience difficulties. This specific structural peculiarity is responsible for the increase in the percentage of time with inter-agent collisions per episode observed for all three collective navigation policies in this environment.

During the 100 MARL episodes conducted across all three new corridors, the $\pi_{collective}^{CEM}$ controller achieves superior average results in terms of explored sectors compared to the $\pi_{collective}^{CMA-ES}$ controller. This success stems from its forward action tendency combined with the evaluation phase's allowance for recovery after collisions. This advantage proves decisive in the corridor with a narrow passage, whose structure presents the greatest divergence from the training environment. In this scenario, the $\pi_{collective}^{CEM}$ controller obtains its most notable advantage over the $\pi_{collective}^{CMA-ES}$ controller. Additionally, the controller trained through our CEM algorithm stands out for its overall balance between safety and exploration, recording the lowest mean rates of collision with moving obstacles in all three scenarios while maintaining competitive traveled distances. Its behavior in corridor (3) is particularly notable, where only 0.78\% of each episode's duration involves inter-agent collisions.

\begin{table*}[h!]
	\begin{minipage}{\textwidth}
		\begin{center}
			\caption{Average flexibility test results across different environments for sectors explored, reward points, percentage of time with inter-robot collisions, traveled distance, and energy consumption.}\label{tab:flexibility}
			\setlength{\tabcolsep}{1.75pt}
			\begin{tabular}{@{}ccccccc@{}}
				\hline
				\textbf{Corridor} & \textbf{Method} & \textbf{Sectors} & \textbf{Reward} & \textbf{Collisions (\%)} & \textbf{Distance (m)} & \textbf{Consumption (Ws)} \\
				\hline
				\hline
				(1) & LiDAR & \textbf{24.83} & \textbf{7177.43} & 1.92 & \textbf{193.27} & 3252.71\\
				(1) & CEM & 21.54 & 4493.92 & \textbf{1.64} & 156.60 & 2248.05\\
				(1) & CMA-ES & 18.42 & 2335.72 & 2.20 & 118.79 & \textbf{2087.76}\\
				\hline
				(2) & LiDAR & 20.41 & \textbf{5509.19} & 3.76 & 154.49 & 2887.14\\
				(2) & CEM & \textbf{21.61} & 5177.19 & \textbf{1.76} & \textbf{165.69} & \textbf{1781.02}\\
				(2) & CMA-ES & 18.49 & 1842.17 & 2.55 & 113.78 & 1809.04\\
				\hline
				(3) & LiDAR & \textbf{31.19} & \textbf{7911.72} & 1.86 & \textbf{211.46} & 3596.15\\
				(3) & CEM & 22.53 & 6133.70 & \textbf{0.78} & 186.27 & 2370.74\\
				(3) & CMA-ES & 13.21 & 703.56 & 2.06 & 79.89 & \textbf{1815.67}\\
				\hline
			\end{tabular}
		\end{center}
	\end{minipage}
\end{table*}

The controller trained using our cross-entropy method even surpasses the LiDAR-based controller in the flexibility experiment within the intersecting corridor (2), achieving an average of 21.61 explored regions and 165.69 meters traveled compared to the reference approach's 20.41 sectors and 154.49 meters. This superiority primarily results from the $\pi_{collective}^{CEM}$ controller's greater spatial coverage of the main corridor in that specific environment. Nevertheless, the LiDAR-based controller outperforms both vision-based controllers in the remaining experiments in terms of visited sectors, accumulated reward, and traveled distance. This superior performance, which comes at the cost of significantly higher energy consumption, stems from the sensor's high precision in measuring distances to static obstacles, typically resulting in more fluid navigation.

Finally, the controller obtained through our CMA-ES algorithm stands out for preserving its characteristic energy efficiency, as shown in Table \ref{tab:flexibility}, but at the cost of a substantial reduction in the distance traveled per episode. This is especially critical in the corridor with a narrow passage, where the swarm travels an average of only 79.89 meters. This behavior demonstrates how its predisposition to turning actions proves ineffective in spatial configurations, such as the aforementioned narrow passage, where maneuverability is severely limited and predominantly rectilinear progression is required.

\subsection{Scalability}\label{subsec:scalability}

Scalability evaluation is a vital component in the validation of multi-robot systems, including artificial swarms. This property determines the system's capacity to maintain or enhance its performance as the number of constituent agents changes. The importance of scalability is widely recognized in scientific literature \citep{Asiain_2020}, as it is a fundamental prerequisite for deploying swarms in large-scale, real-world applications. Testing a vision-only controller with varying swarm sizes provides valuable insights into its adaptability and coordination capabilities under different agent densities.

Conducting such scalability evaluations with large numbers of physical robots is often prohibitively expensive and logistically complex. Simulation-based validation is, therefore, particularly valuable in swarm robotics, as it permits the systematic examination of system behavior under a wide range of configurations and agent densities. This approach enables an assessment of performance from a few agents to many, a task that would be impracticable in physical environments due to constraints of time, cost, and logistics.

\begin{table*}[h!]
	\begin{minipage}{\textwidth}
		\begin{center}
			\caption{Average scalability test results across different swarm sizes for sectors explored, reward points, percentage of time with inter-robot collisions, traveled distance, and energy consumption.}\label{tab:scalability}
			\setlength{\tabcolsep}{2.75pt}
			\begin{tabular}{@{}ccccccc@{}}
				\hline
				\textbf{Robots} & \textbf{Method} & \textbf{Sectors} & \textbf{Reward} & \textbf{Collisions (\%)} & \textbf{Distance (m)} & \textbf{Consumption (Ws)} \\
				\hline
				\hline
				2 & LiDAR & \textbf{28.64} & \textbf{4198.51} & 0.37 & \textbf{111.96} & 1803.83\\
				2 & CEM & 18.24 & 2976.72 & \textbf{0.13} & 91.25 & 1144.42\\
				2 & CMA-ES & 9.33 & 337.84 & 0.45 & 38.14 & \textbf{843.78}\\
				\hline
				8 & LiDAR & \textbf{32.71} & \textbf{15540.27} & 6.81 & \textbf{419.95} & 7348.64\\
				8 & CEM & 25.09 & 12147.43 & \textbf{4.35} & 373.93 & 4827.89\\
				8 & CMA-ES & 16.71 & 2124.36 & 11.29 & 177.44 & \textbf{3654.71}\\
				\hline
				16 & LiDAR & \textbf{33.00} & \textbf{29306.89} & 20.25 & \textbf{796.61} & 14539.18\\
				16 & CEM & 27.77 & 23719.90 & \textbf{15.95} & 735.47 & 9129.75\\
				16 & CMA-ES & 22.55 & 4810.51 & 34.57 & 381.75 & \textbf{7533.37}\\
				\hline
			\end{tabular}
		\end{center}
	\end{minipage}
\end{table*}

\begin{figure*}[h!]
	\centering
	\includegraphics[height=175mm, width=\textwidth]{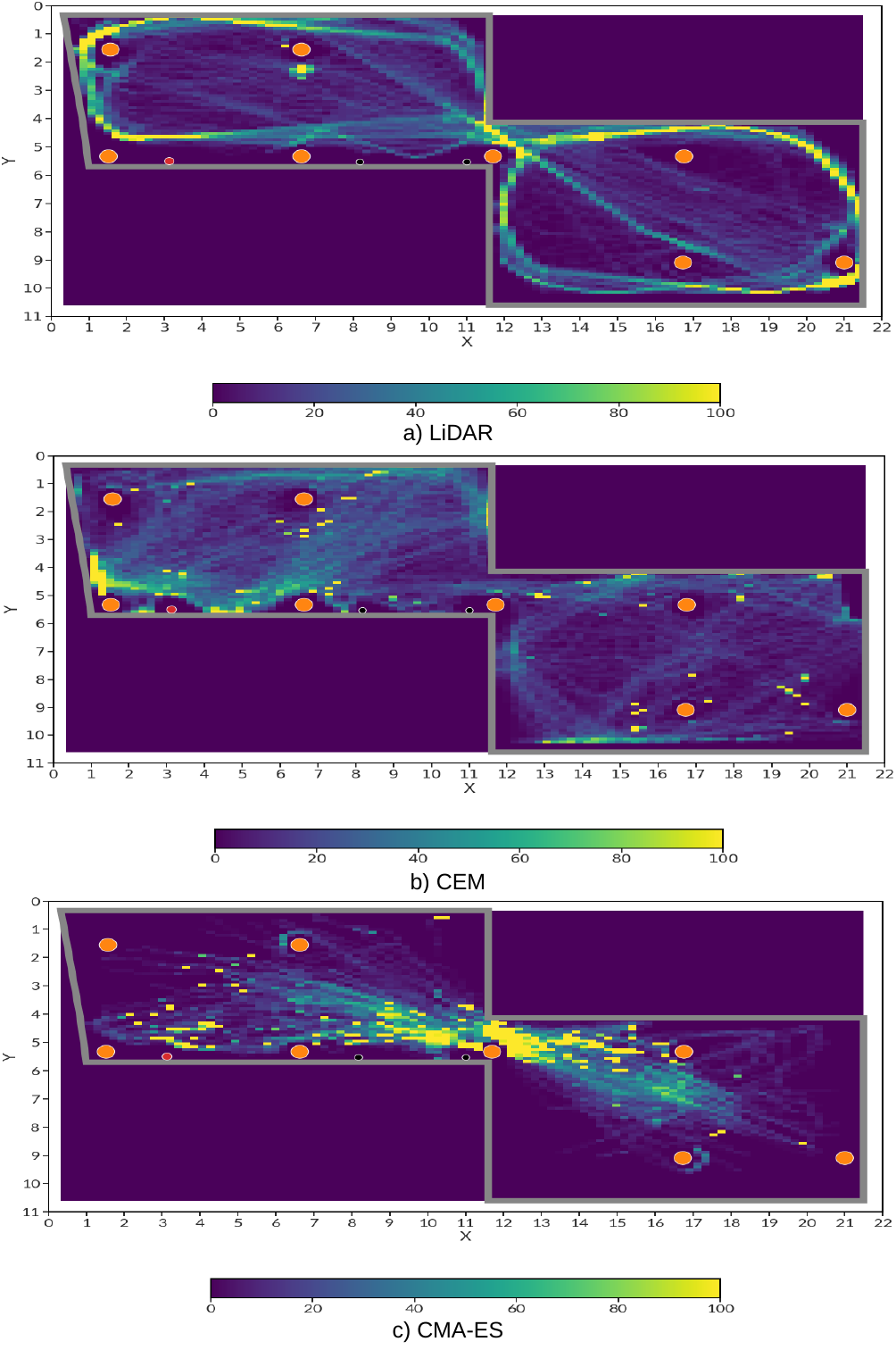}
	\caption{Navigational trajectory frequency heatmaps with 2 robots. This figure illustrates the number of times two robots traversed different positions in the simulated environment (3) over 100 episodes. This scalability assessment compares behaviors from: a) the LiDAR baseline, b) the proposed CEM-trained policy, and c) the proposed CMA-ES-trained policy.}\label{fig:scalability_2}
\end{figure*}

In our evaluation, we analyze system performance with swarms of 2, 8, and 16 robots, providing insights into both reduction and significant increase scenarios compared to the original 4-robot training configuration. It must be explicitly reiterated that neither $\pi_{collective}^{CEM}$ nor $\pi_{collective}^{CMA-ES}$ were retrained, fine-tuned, or adapted in any capacity for these new configurations. A truly scalable system should maintain consistent and effective performance across these variations. We selected synthetic environment (3), the corridor with a narrow passage, for this scalability analysis. This environment is particularly appropriate due to its larger navigable surface area and complex structure, which offers a more diverse exploration space suitable for larger swarms while minimizing artificial congestion. Furthermore, its narrow central passage provides an additional challenge to test coordinated navigation in high-density situations.

\begin{figure*}[h!]
	\centering
	\includegraphics[height=175mm, width=\textwidth]{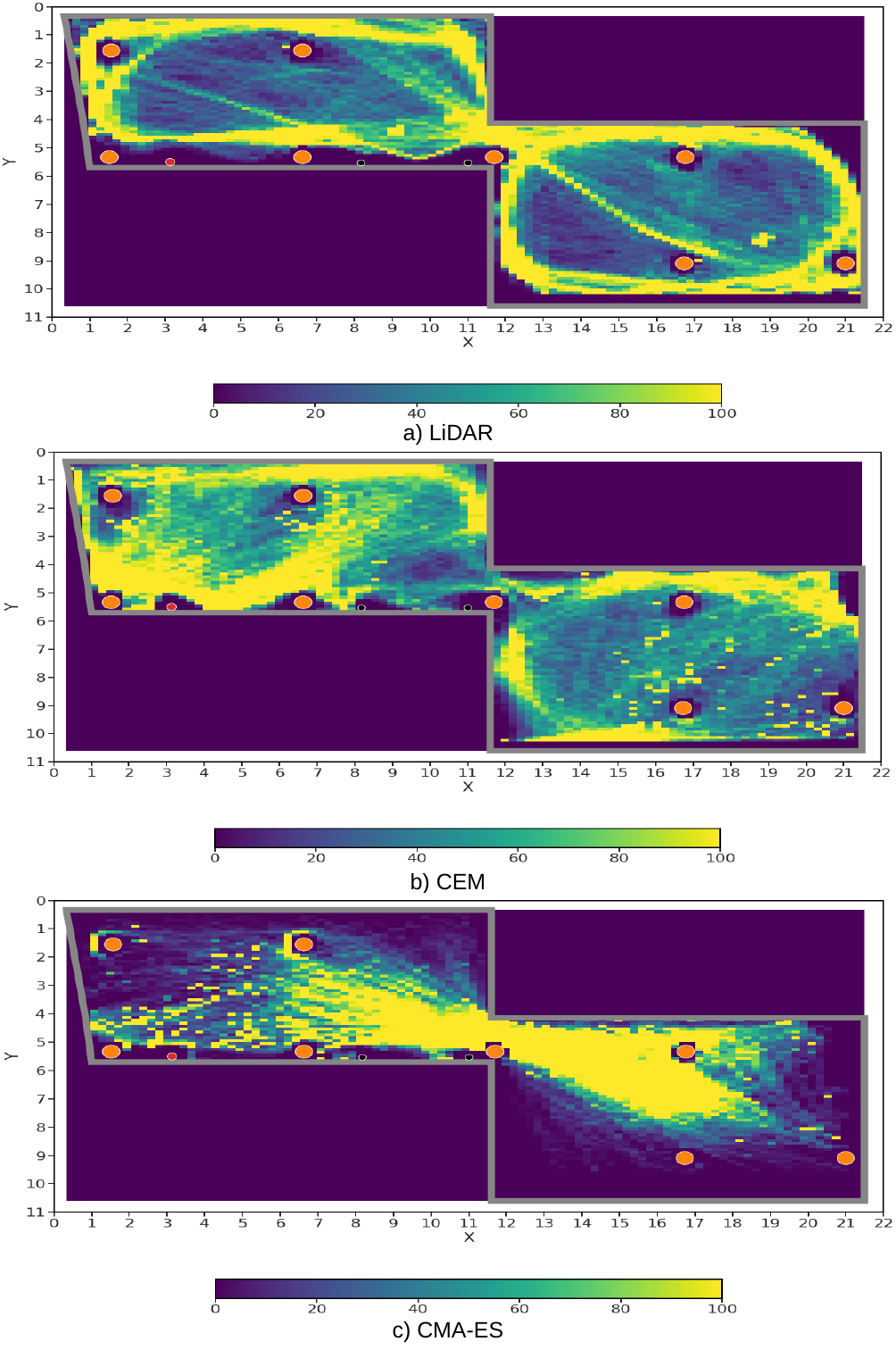}
	\caption{Navigational trajectory frequency heatmaps with 8 robots. This figure illustrates the number of times eight robots traversed different positions in the simulated environment (3) over 100 episodes. This scalability assessment compares behaviors from: a) the LiDAR baseline, b) the proposed CEM-trained policy, and c) the proposed CMA-ES-trained policy.}\label{fig:scalability_8}
\end{figure*}

\begin{figure*}[h!]
	\centering
	\includegraphics[height=175mm, width=\textwidth]{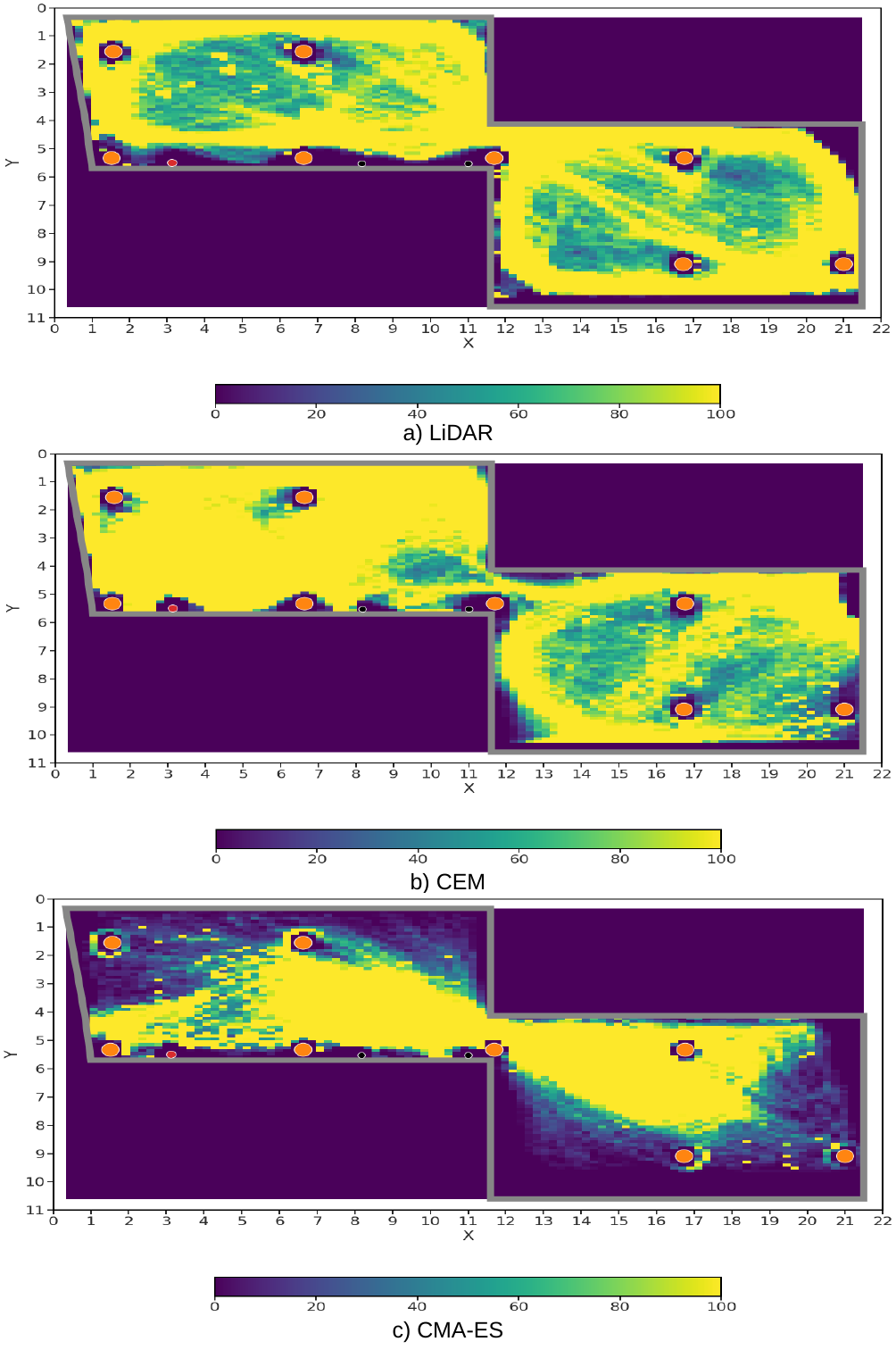}
	\caption{Navigational trajectory frequency heatmaps with 16 robots. This figure illustrates the number of times sixteen robots traversed different positions in the simulated environment (3) over 100 episodes. This scalability assessment compares behaviors from: a) the LiDAR baseline, b) the proposed CEM-trained policy, and c) the proposed CMA-ES-trained policy.}\label{fig:scalability_16}
\end{figure*}

The test results reveal notable effects on exploration as the swarm size changes, as evidenced in Table \ref{tab:scalability}. With a 2-robot swarm, the $\pi_{collective}^{CEM}$ controller shows an expected reduction in average accumulated reward, explored sectors, and traveled distance compared to the 4-robot configuration. Specifically, the 2-robot swarm achieves 2976.72 reward points, 18.24 explored sectors, and 91.25 meters traveled whereas the 4-robot swarm reaches 6133.70 points, 22.53 sectors, and 186.27 meters. Notably, halving the number of robots does not halve the exploratory coverage, indicating positive scalability characteristics even in reduction. The $\pi_{collective}^{CMA-ES}$ controller exhibits a more pronounced, though expected, performance decrease in this scenario. Qualitatively, as shown in Fig. \ref{fig:scalability_2}, both controllers maintain exploration patterns similar to their 4-robot counterparts. The $\pi_{collective}^{CEM}$ policy promotes a more dispersed exploration, reaching peripheral positions, while the $\pi_{collective}^{CMA-ES}$ policy favors curved trajectories, resulting in exhaustive coverage of central areas.

Increasing the swarm size to eight robots yields substantial improvements in accumulated reward and total distance traveled for both controllers, alongside moderate increases in the average number of explored sectors. The $\pi_{collective}^{CEM}$ controller's coverage increases from 22.53 to 25.09 sectors, while the $\pi_{collective}^{CMA-ES}$ controller's rises from 13.21 to 16.71. It is critical to note that each additional robot also acts as a dynamic obstacle, significantly increasing the probability of inter-agent collisions. This factor likely explains the modest improvement in explored regions despite a doubled swarm size. Qualitatively, the heatmaps in Fig. \ref{fig:scalability_8} show broad environment coverage, which becomes nearly complete for the $\pi_{collective}^{CEM}$ controller. This policy's pattern is visibly more uniform, whereas the $\pi_{collective}^{CMA-ES}$ controller maintains a more focused area coverage.

With 16-robot configurations, both approaches demonstrate considerable further performance improvements in exploration. The $\pi_{collective}^{CEM}$ controller achieves a remarkable average reward of 23719.90 points, travels 735.47 meters, and explores 27.77 sectors, while the $\pi_{collective}^{CMA-ES}$ controller attains 4810.51 points, 381.75 meters, and 22.55 sectors. This latter value represents a significant improvement from its 16.71-sector performance with eight robots. These results suggest the 16-agent configuration may be optimal for our vision-based controller in corridor (3), as larger swarms can better exploit inherent redundancy to counteract potential collisions.

However, as expected, the mean rate of inter-robot collisions increases substantially, reaching 34.57\% for the $\pi_{collective}^{CMA-ES}$ controller. Its inherent tendency to select turning actions becomes counterproductive in restricted spaces with high agent density, which generates unpredictable trajectories. Total energy consumption also scales significantly, exceeding 14500 Ws for the LiDAR-based method. Qualitatively, as observed in Fig. \ref{fig:scalability_16}, the 16-robot swarms achieve near-complete environment coverage. The $\pi_{collective}^{CEM}$ controller successfully explores all 33 sectors in 32 of the 100 episodes, and all agents in both policies successfully navigate the central narrow passage.

As evidenced by the quantitative results detailed above, the system exhibits a fundamental trade-off inherent to decentralized swarm robotics: the balance between spatial coverage effectiveness and multi-agent interference. While the raw exploration capacity naturally scales with the addition of agents, the marginal utility per robot (\textit{i.e.}, the net increase in spatial coverage contributed by each supplementary agent) is bounded by the increased frequency of inter-agent encounters in confined spaces \citep{Asiain_2020}, a dynamic clearly reflected in the performance variations between the 8 and 16-robot swarms.

Despite this anticipated interference, a highly relevant qualitative emergent property observed alongside these quantitative metrics is the spontaneous resolution of multi-robot gridlocks in high-density configurations. In densely populated narrow corridors, reactive homogeneous swarms can fall into perpetual deadlocks due to symmetrical spatial conflicts. However, the proposed framework naturally breaks these symmetries without the injection of artificial sensory noise. Even though all agents execute identical deterministic policies during a given trial, the continuous nature of the simulation physics and the microscopic spatial asymmetries in their initial trajectories prevent perfect symmetrical lock-ups.

Because the visual perceptions are inherently dependent on the continuous coordinate space, each individual agent experiences a uniquely angled, sub-pixel viewpoint of the congestion. This microscopic heterogeneity in their momentary latent representations leads to divergent collision-avoidance maneuvers. Consequently, the swarm fluidly dissipates congestion without requiring explicit inter-agent communication, demonstrating that the behavioral parameters learned by the macroscopic model are intrinsically scalable and structurally robust against severe density fluctuations.

In summary, the LiDAR-based controller demonstrates exceptional scalability in terms of raw exploration, achieving 100\% coverage (\textit{i.e.}, 33 sectors) in all 100 tests with 16 robots. This highlights its superior precision in navigating static obstacles. The $\pi_{collective}^{CEM}$ controller achieves the closest, though still inferior, exploration results. However, a critical distinction emerges in safety: our vision-based CEM-trained controller presents the lowest average collision rates with dynamic obstacles across all scenarios. This robust resistance to performance degradation, even in high density, suggests the vision-based policy can process rich contextual information, distinguish between other robots and fixed obstacles, and adapt its forward-biased behavior accordingly. This scalability in safety and efficiency is a crucial finding for real-world swarm deployment.

\subsection{Comparative Statistical Analysis and Discussion}\label{subsec:discussion}

This subsection presents a comprehensive statistical analysis and comparative discussion of the swarm robotic properties evaluated in this section. We examine the experimental results from Subsections \ref{subsec:robustness}, \ref{subsec:flexibility}, and \ref{subsec:scalability} using a rigorous statistical methodology to formally assess the relative robustness, flexibility, and scalability of our generated controllers. The following discussion provides critical insights into the performance characteristics of the $\pi_{collective}^{CEM}$ and $\pi_{collective}^{CMA-ES}$ policies, benchmarked against the conventional LiDAR-based approach.

The experiments have confirmed that our macroscopic model, trained via neuro-evolution, successfully addresses the challenges of collective visual exploration. Both the CEM and CMA-ES algorithms demonstrated the capacity to generate effective swarm robotic controllers for navigation in uncertain environments, although their resultant performance characteristics were markedly distinct.

Following the statistical methodology established in Subsection \ref{subsec:statistical}, we first analyzed the robustness of the proposed navigation approaches under different sensory noise conditions. The Tukey HSD test results, presented in Table \ref{tab:robustness_hsd}, yield critical conclusions regarding the exploration effectiveness of our vision-based controllers.

\begin{table*}[h!]
	\begin{minipage}{\textwidth}
		\begin{center}
			\caption{Results of the Tukey HSD test for comparing means of sectors explored by the three collective autonomous navigation controllers evaluated under different Gaussian noise levels.}\label{tab:robustness_hsd}
			\setlength{\tabcolsep}{1pt}
			\begin{tabular}{@{}cccccc@{}}
				\hline
				\textbf{Noise} & \textbf{Comparison} & \textbf{Difference} & \textbf{p-Value} & \textbf{Confidence Interval (95\%)} & \textbf{Significant} \\
				\hline
				\hline
				5\% & CEM vs LiDAR & -0.15 & 0.8726 & [-0.8604, 0.5604] & No\\
				5\% & CMA-ES vs LiDAR & -2.11 & 0.0 & [-2.8204, -1.3996] & Yes\\
				5\% & CMA-ES vs CEM & -1.96 & 0.0 & [-2.6704, -1.2496] & Yes\\
				\hline
				10\% & CEM vs LiDAR & 0.9 & 0.0329 & [0.0581, 1.7419] & Yes\\
				10\% & CMA-ES vs LiDAR & -4.18 & 0.0 & [-5.0219, -3.3381] & Yes\\
				10\% & CMA-ES vs CEM & -5.08 & 0.0 & [-5.9219, -4.2381] & Yes\\
				\hline
				15\% & CEM vs LiDAR & 0.34 & 0.6149 & [-0.5113, 1.1913] & No\\
				15\% & CMA-ES vs LiDAR & -9.82 & 0.0 & [-10.6713, -8.9687] & Yes\\
				15\% & CMA-ES vs CEM & -10.16 & 0.0 & [-11.0113, -9.3087] & Yes\\
				\hline
			\end{tabular}
		\end{center}
	\end{minipage}
\end{table*}

The statistical analysis consistently demonstrated the $\pi_{collective}^{CEM}$ controller's superior exploration capability compared to the $\pi_{collective}^{CMA-ES}$ controller across all noise levels. This performance gap became increasingly pronounced as the sensory noise intensified: the $\pi_{collective}^{CEM}$ controller outperformed the $\pi_{collective}^{CMA-ES}$ controller by a mean of $1.96$ sectors at 5\% noise, $5.08$ sectors at 10\% noise, and a substantial $10.16$ sectors at 15\% noise. More importantly, when compared to the LiDAR benchmark, our CEM-trained controller showed no statistically significant difference at 5\% and 15\% noise levels. In a notable finding, with 10\% noise, this vision-based controller demonstrated statistically superior performance to the LiDAR approach by an average of $0.9$ sectors.

These results highlight the remarkable robustness of the CEM-trained controller. Its ability to maintain, and in one case exceed, the exploration effectiveness of the resource-intensive LiDAR system despite significant sensory perturbation is a crucial validation. This resilience is paramount for practical deployment in real-world environments, which are characterized by unpredictable sensory variability.

The flexibility analysis, which employed the same statistical framework, assessed the controllers' adaptability to novel environmental configurations. The Tukey HSD test results, detailed in Table \ref{tab:flexibility_hsd}, reveal clear patterns.

\begin{table*}[h!]
	\begin{minipage}{\textwidth}
		\begin{center}
			\caption{Results of the Tukey HSD test for comparing means of sectors explored by the three collective autonomous navigation controllers evaluated in novel, unknown environments.}\label{tab:flexibility_hsd}
			\setlength{\tabcolsep}{1pt}
			\begin{tabular}{@{}cccccc@{}}
				\hline
				\textbf{Corridor} & \textbf{Comparison} & \textbf{Difference} & \textbf{p-Value} & \textbf{Confidence Interval (95\%)} & \textbf{Significant} \\
				\hline
				\hline
				(1) & CEM vs LiDAR & -3.29 & 0.0 & [-4.4961, -2.0839] & Yes\\
				(1) & CMA-ES vs LiDAR & -6.41 & 0.0 & [-7.6161, -5.2039] & Yes\\
				(1) & CMA-ES vs CEM & -3.12 & 0.0 & [-4.3261, -1.9139] & Yes\\
				\hline
				(2) & CEM vs LiDAR & 1.2 & 0.2998 & [-0.704, 3.104] & No\\
				(2) & CMA-ES vs LiDAR & -1.92 & 0.0476 & [-3.824, -0.016] & Yes\\
				(2) & CMA-ES vs CEM & -3.12 & 0.0004 & [-5.024, -1.216] & Yes\\
				\hline
				(3) & CEM vs LiDAR & -8.66 & 0.0 & [-10.0988, -7.2212] & Yes\\
				(3) & CMA-ES vs LiDAR & -17.98 & 0.0 & [-19.4188, -16.5412] & Yes\\
				(3) & CMA-ES vs CEM & -9.32 & 0.0 & [-10.7588, -7.8812] & Yes\\
				\hline
			\end{tabular}
		\end{center}
	\end{minipage}
\end{table*}

The $\pi_{collective}^{CEM}$ controller's superior exploration capability over the $\pi_{collective}^{CMA-ES}$ controller was consistent across all three new environments. This advantage was most pronounced in environment (3) (\textit{i.e.}, the corridor with a narrow passage), where it explored an average of $9.32$ more sectors. In comparison to the reference approach, the LiDAR benchmark performed significantly better in environments (1) (\textit{i.e.}, the corridor with a corner) and (3). However, in environment (2) (\textit{i.e.}, the corridor with an intersection), no statistically significant difference was found between the $\pi_{collective}^{CEM}$ controller and the LiDAR approach. The adjusted p-value of $0.2998$ exceeds the threshold $\alpha = 0.05$, which confirms the impossibility of rejecting the null hypothesis of equality between means.

These findings underline the CEM-generated controller's remarkable adaptability to different environmental conditions, especially in our second flexibility experiment, where it achieves exploration efficacy statistically equivalent to the specialized LiDAR system in a structurally complex, unknown environment. This confirmed adaptability is essential for real-world applications where systems must be deployed in unmapped or dynamically changing spaces.

Finally, the scalability analysis, alongside the corresponding Tukey HSD test results presented in Table \ref{tab:scalability_hsd}, provided clear evidence of each controller's ability to maintain exploration efficacy with varying agent counts.

\begin{table*}[h!]
	\begin{minipage}{\textwidth}
		\begin{center}
			\caption{Results of the Tukey HSD test for comparing means of sectors explored by the three collective autonomous navigation controllers evaluated across different swarm sizes.}\label{tab:scalability_hsd}
			\setlength{\tabcolsep}{1pt}
			\begin{tabular}{@{}cccccc@{}}
				\hline
				\textbf{Robots} & \textbf{Comparison} & \textbf{Difference} & \textbf{p-Value} & \textbf{Confidence Interval (95\%)} & \textbf{Significant} \\
				\hline
				\hline
				2 & CEM vs LiDAR & -10.4 & 0.0 & [-11.8718, -8.9282] & Yes\\
				2 & CMA-ES vs LiDAR & -19.31 & 0.0 & [-20.7818, -17.8382] & Yes\\
				2 & CMA-ES vs CEM & -8.91 & 0.0 & [-10.3818, -7.4382] & Yes\\
				\hline
				8 & CEM vs LiDAR & -7.62 & 0.0 & [-9.2031, -6.0369] & Yes\\
				8 & CMA-ES vs LiDAR & -16.0 & 0.0 & [-17.5831, -14.4169] & Yes\\
				8 & CMA-ES vs CEM & -8.38 & 0.0 & [-9.9631, -6.7969] & Yes\\
				\hline
				16 & CEM vs LiDAR & -5.23 & 0.0 & [-6.6737, -3.7863] & Yes\\
				16 & CMA-ES vs LiDAR & -10.45 & 0.0 & [-11.8937, -9.0063] & Yes\\
				16 & CMA-ES vs CEM & -5.22 & 0.0 & [-6.6637, -3.7763] & Yes\\
				\hline
			\end{tabular}
		\end{center}
	\end{minipage}
\end{table*}

The statistical analysis consistently demonstrated the CEM-trained controller's superior exploration capability compared to the controller obtained through our CMA-ES algorithm across all swarm sizes. This advantage remained robust, with the $\pi_{collective}^{CEM}$ controller outperforming the $\pi_{collective}^{CMA-ES}$ controller by an average of $8.91$ sectors with 2 robots, $8.38$ sectors with 8 robots, and $5.22$ sectors with 16 robots. When compared to the baseline approach, the LiDAR-based controller demonstrated statistically superior exploration in all scalability tests. However, a notable and promising trend emerged: the performance gap between the $\pi_{collective}^{CEM}$ controller and the LiDAR approach progressively narrowed as the swarm size increased, shrinking from a mean difference of $10.40$ sectors with 2 robots to $5.23$ sectors with 16 robots.

These results underscore the $\pi_{collective}^{CEM}$ controller's effective scalability. Its improved relative performance in larger swarms suggests that our vision-based method becomes increasingly competitive with traditional, resource-intensive sensor approaches in high-density, multi-agent deployments, strongly supporting its potential for large-scale collective navigation applications in the real world.

\section{Conclusions}\label{sec:conclusions}

This study has presented specialized implementations of the cross-entropy method and the covariance matrix adaptation evolution strategy tailored for multi-agent reinforcement learning. Through these evolutionary techniques, we have successfully developed two distinct autonomous navigation policies, establishing a novel hybrid method for the automatic design of collective behaviors for low-cost, reactive robotic swarms. The developed system, founded on advanced machine learning and an efficient neural architecture, demonstrates the ability to learn and deploy adaptive collective behaviors for visual exploration in challenging, dynamic environments.

Our investigation has successfully addressed the five fundamental research questions posed in Section \ref{sec:introduction}. The proposed hybrid automatic design method demonstrated its effectiveness in generating functional swarm robotic controllers for visual navigation tasks. The resulting controllers exhibited remarkable robustness against sensory perturbations and considerable flexibility in adapting to diverse environmental configurations not encountered during training. Scalability experiments confirmed that our approach remains effective as the number of agents varies, with performance improvements observed in larger swarms. Finally, our best vision-based controller achieved performance statistically comparable to traditional navigation methods employing more expensive sensors, validating the viability of our low-cost, monocular-vision approach.

Across all experiments, our CEM-generated controller demonstrated superior efficacy for collective exploration, enabling the swarm to visit an average of 36.20\% more regions than the controller obtained via the CMA-ES algorithm. Furthermore, while achieving exploration capabilities statistically comparable to a conventional LiDAR-based method, our best vision-based controller yielded a significant 31.40\% average reduction in energy consumption. This energy efficiency is a critical finding, as it addresses the strict computational and power constraints inherent to low-cost swarm robotics. In an era increasingly dominated by large-scale, resource-intensive AI models such as transformers, this research validates that architectural efficiency is a practical and effective alternative for hardware-constrained domains like swarm robotics, establishing a viable path for developing low-cost, autonomous collective intelligence.

The positive results from this comprehensive analysis underscore the potential for deploying the proposed swarm model in real-world scenarios. The demonstrated robustness, flexibility, and scalability are fundamental properties for practical applicability in complex and hazardous environments where multi-agent systems offer significant advantages over single-agent systems. This opens viable pathways for applications in fields such as search and rescue in emergency situations, as well as environmental monitoring and precision agriculture.

\subsection{Limitations}\label{subsec:limitations}

Despite the robust empirical performance of our hybrid neuro-evolutionary framework, several limitations must be acknowledged. First, due to the substantial computational cost associated with multi-agent macroscopic simulations, comprehensive ablation studies isolating the exact marginal contributions of specific architectural components, such as the individual terms of the reward function or the substitution of the perception module, were not performed. 

Second, while the system successfully demonstrated flexibility by generalizing to unseen corridor topologies without requiring computationally expensive retraining phases, as evidenced in Subsection \ref{subsec:flexibility}, these test environments did not encompass extreme navigational challenges. Consequently, the generalization capability of the reactive controllers in drastically more complex scenarios featuring extreme occlusions, highly restrictive narrow passages, or dead ends requires further empirical validation.

Finally, the current methodology relies exclusively on simulation-based training prior to physical deployment. The lack of online adaptation or incremental continual learning mechanisms constitutes a significant constraint. Addressing the feasibility of real-time evolutionary optimization on physical swarms presents profound safety challenges; the stochastic weight mutations inherent to algorithms like CEM and CMA-ES introduce severe risks of catastrophic physical collisions if executed outside a secure simulated sandbox.

\subsection{Future Work}\label{subsec:future}

To address the aforementioned limitations and expand the capabilities of the proposed framework, future research will prioritize several key directions. The immediate priority is the deployment of the synthesized controllers on physical multi-robot systems. While the proposed architecture is designed to mitigate the sim-to-real domain gap by addressing onboard inference latency via the VAE bottleneck and actuation uncertainty through high-frequency reactive control, empirical validation on physical hardware remains essential.

To overcome the constraints of purely simulation-based optimization, we intend to explore safe reinforcement learning paradigms to enable lightweight, incremental fine-tuning. Developing secure sim-to-real-to-sim pipelines will be crucial to facilitate real-time adaptation without compromising the physical integrity of the swarm, aligning with recent paradigm shifts in robot learning and evolution \citep{Kuckling_2023}.

Furthermore, to further improve training stability, continuous control precision, and sample efficiency, future iterations of the framework will explore hybridizing the current evolutionary approach with modern deep reinforcement learning algorithms. Recent studies have demonstrated that integrating deterministic policy gradient methods, such as the Deep Deterministic Policy Gradient (DDPG) and Twin Delayed Deep Deterministic Policy Gradient (TD3) algorithms, with baseline stabilizing controllers yields exceptional performance and disturbance rejection in continuous robotic control tasks \citep{Hazem_2025a, Hazem_2025c, Hazem_2025b}. Adapting these hybrid deep RL paradigms to decentralized multi-agent navigation presents a highly promising avenue for scaling the resilience and adaptability of swarm robotics.

\section*{CRediT author statement}

\textbf{\'{A}lvaro D\'{i}ez:} Conceptualization, Methodology, Software, Validation, Formal analysis, Investigation, Data curation, Writing - Original draft, Writing - Review and editing, Visualization. \textbf{Fidel Aznar:} Conceptualization, Methodology, Software, Validation, Resources, Supervision, Project administration.

\section*{Declaration of interests}

The authors declare that they have no known competing financial interests or personal relationships that could have appeared to influence the work reported in this paper.

\section*{Declaration of generative AI and AI-assisted technologies in the writing process}

During the preparation of this work, the authors used Google Gemini 2.5 in order to improve the quality and readability of the written text. After using this tool, the authors reviewed and edited the content as needed and take full responsibility for the content of the published article.

\section*{Acknowledgments}

This work was supported by the CIAICO/2026/169 project from the Generalitat Valenciana.


\begin{thebibliography}{61}
	\expandafter\ifx\csname natexlab\endcsname\relax\def\natexlab#1{#1}\fi
	\providecommand{\url}[1]{\texttt{#1}}
	\providecommand{\href}[2]{#2}
	\providecommand{\path}[1]{#1}
	\providecommand{\DOIprefix}{doi:}
	\providecommand{\ArXivprefix}{arXiv:}
	\providecommand{\URLprefix}{URL: }
	\providecommand{\Pubmedprefix}{pmid:}
	\providecommand{\doi}[1]{\href{http://dx.doi.org/#1}{\path{#1}}}
	\providecommand{\Pubmed}[1]{\href{pmid:#1}{\path{#1}}}
	\providecommand{\bibinfo}[2]{#2}
	\ifx\xfnm\relax \def\xfnm[#1]{\unskip,\space#1}\fi
	\bibitem[{Amato et~al.(2013)Amato, Chowdhary, Geramifard, Ure and
		Kochenderfer}]{Amato_2013}
	\bibinfo{author}{Amato, C.}, \bibinfo{author}{Chowdhary, G.},
	\bibinfo{author}{Geramifard, A.}, \bibinfo{author}{Ure, N.K.},
	\bibinfo{author}{Kochenderfer, M.J.}, \bibinfo{year}{2013}.
	\newblock \bibinfo{title}{Decentralized control of partially observable
		{M}arkov decision processes}, in: \bibinfo{booktitle}{2013 52nd IEEE
		Conference on Decision and Control (CDC)}, pp. \bibinfo{pages}{2398--2405}.
	\newblock \bibinfo{note}{IEEE, Florence, Italy.
		\url{https://doi.org/10.1109/CDC.2013.6760239}}.
	\bibitem[{Asiain and Godoy(2020)}]{Asiain_2020}
	\bibinfo{author}{Asiain, J.}, \bibinfo{author}{Godoy, J.},
	\bibinfo{year}{2020}.
	\newblock \bibinfo{title}{{N}avigation in {L}arge {G}roups of {R}obots}.
	\newblock \bibinfo{journal}{Current Robotics Reports} \bibinfo{volume}{1},
	\bibinfo{pages}{203--213}.
	\newblock \bibinfo{note}{\url{https://doi.org/10.1007/s43154-020-00017-2}}.
	\bibitem[{B\"{a}ck and Schwefel(1996)}]{Back_1996}
	\bibinfo{author}{B\"{a}ck, T.}, \bibinfo{author}{Schwefel, H.P.},
	\bibinfo{year}{1996}.
	\newblock \bibinfo{title}{Evolutionary computation: an overview}, in:
	\bibinfo{booktitle}{1996 IEEE International Conference on Evolutionary
		Computation (ICEC)}, pp. \bibinfo{pages}{20--29}.
	\newblock \bibinfo{note}{IEEE, Nagoya, Japan.
		\url{https://doi.org/10.1109/ICEC.1996.542329}}.
	\bibitem[{Bank et~al.(2023)Bank, Koenigstein and Giryes}]{Bank_2023}
	\bibinfo{author}{Bank, D.}, \bibinfo{author}{Koenigstein, N.},
	\bibinfo{author}{Giryes, R.}, \bibinfo{year}{2023}.
	\newblock \bibinfo{title}{Autoencoders}, in: \bibinfo{editor}{Rokach, L.},
	\bibinfo{editor}{Maimon, O.}, \bibinfo{editor}{Shmueli, E.} (Eds.),
	\bibinfo{booktitle}{Machine Learning for Data Science Handbook}, pp.
	\bibinfo{pages}{353--374}.
	\newblock \bibinfo{note}{Springer, Cham.
		\url{https://doi.org/10.1007/978-3-031-24628-9_16}}.
	\bibitem[{Batra et~al.(2021)Batra, Huang, Petrenko, Kumar, Molchanov and
		Sukhatme}]{Batra_2021}
	\bibinfo{author}{Batra, S.}, \bibinfo{author}{Huang, Z.},
	\bibinfo{author}{Petrenko, A.}, \bibinfo{author}{Kumar, T.},
	\bibinfo{author}{Molchanov, A.}, \bibinfo{author}{Sukhatme, G.S.},
	\bibinfo{year}{2021}.
	\newblock \bibinfo{title}{{D}ecentralized {C}ontrol of {Q}uadrotor {S}warms
		with {E}nd-to-end {D}eep {R}einforcement {L}earning}, in:
	\bibinfo{booktitle}{2021 5th Conference on Robot Learning (CoRL)}, pp.
	\bibinfo{pages}{576--586}.
	\newblock \bibinfo{note}{PMLR, London, United Kingdom.
		\url{https://proceedings.mlr.press/v164/batra22a/batra22a.pdf}}.
	\bibitem[{Bezerra et~al.(2025)Bezerra, dos Santos and Park}]{Bezerra_2025}
	\bibinfo{author}{Bezerra, L.C.D.}, \bibinfo{author}{dos Santos, A.M.G.},
	\bibinfo{author}{Park, S.}, \bibinfo{year}{2025}.
	\newblock \bibinfo{title}{{L}earning {P}olicies for {D}ynamic {C}oalition
		{F}ormation in {M}ulti-{R}obot {T}ask {A}llocation}.
	\newblock \bibinfo{journal}{IEEE Robotics and Automation Letters}
	\bibinfo{volume}{10}, \bibinfo{pages}{9216--9223}.
	\newblock \bibinfo{note}{\url{https://doi.org/10.1109/LRA.2025.3592080}}.
	\bibitem[{Bloom et~al.(2022)Bloom, Mukherjee and Pinciroli}]{Bloom_2022}
	\bibinfo{author}{Bloom, J.}, \bibinfo{author}{Mukherjee, A.},
	\bibinfo{author}{Pinciroli, C.}, \bibinfo{year}{2022}.
	\newblock \bibinfo{title}{{A} {S}tudy of {R}einforcement {L}earning
		{A}lgorithms for {A}ggregates of {M}inimalistic {R}obots}.
	\newblock \bibinfo{note}{Preprint at
		\url{https://doi.org/10.48550/arXiv.2203.15129}}.
	\bibitem[{Boige et~al.(2023)Boige, Flet-Berliac, Flajolet, Richard and
		Pierrot}]{Boige_2023}
	\bibinfo{author}{Boige, R.}, \bibinfo{author}{Flet-Berliac, Y.},
	\bibinfo{author}{Flajolet, A.}, \bibinfo{author}{Richard, G.},
	\bibinfo{author}{Pierrot, T.}, \bibinfo{year}{2023}.
	\newblock \bibinfo{title}{{PASTA}: {P}retrained {A}ction-{S}tate {T}ransformer
		{A}gents}.
	\newblock \bibinfo{note}{Preprint at
		\url{https://doi.org/10.48550/arXiv.2307.10936}}.
	\bibitem[{Botev and Kroese(2004)}]{Botev_2004}
	\bibinfo{author}{Botev, Z.}, \bibinfo{author}{Kroese, D.P.},
	\bibinfo{year}{2004}.
	\newblock \bibinfo{title}{{G}lobal {L}ikelihood {O}ptimization {V}ia the
		{C}ross-{E}ntropy {M}ethod, with an {A}pplication to {M}ixture {M}odels}, in:
	\bibinfo{booktitle}{2004 Winter Simulation Conference (WSC)}, pp.
	\bibinfo{pages}{535--541}.
	\newblock \bibinfo{note}{IEEE, Washington D.C., United States.
		\url{https://doi.org/10.1109/WSC.2004.1371358}}.
	\bibitem[{Brambilla et~al.(2013)Brambilla, Ferrante, Birattari and
		Dorigo}]{Brambilla_2013}
	\bibinfo{author}{Brambilla, M.}, \bibinfo{author}{Ferrante, E.},
	\bibinfo{author}{Birattari, M.}, \bibinfo{author}{Dorigo, M.},
	\bibinfo{year}{2013}.
	\newblock \bibinfo{title}{Swarm robotics: a review from the swarm engineering
		perspective}.
	\newblock \bibinfo{journal}{Swarm Intelligence} \bibinfo{volume}{7},
	\bibinfo{pages}{1--41}.
	\newblock \bibinfo{note}{\url{https://doi.org/10.1007/s11721-012-0075-2}}.
	\bibitem[{Cheraghi et~al.(2021)Cheraghi, Shahzad and Graffi}]{Cheraghi_2021}
	\bibinfo{author}{Cheraghi, A.R.}, \bibinfo{author}{Shahzad, S.},
	\bibinfo{author}{Graffi, K.}, \bibinfo{year}{2021}.
	\newblock \bibinfo{title}{{P}ast, {P}resent, and {F}uture of {S}warm
		{R}obotics}.
	\newblock \bibinfo{note}{Preprint at
		\url{https://doi.org/10.48550/arXiv.2101.00671}}.
	\bibitem[{\c{S}ahin(2005)}]{Sahin_2005}
	\bibinfo{author}{\c{S}ahin, E.}, \bibinfo{year}{2005}.
	\newblock \bibinfo{title}{{S}warm {R}obotics: {F}rom {S}ources of {I}nspiration
		to {D}omains of {A}pplication}, in: \bibinfo{editor}{\c{S}ahin, E.},
	\bibinfo{editor}{Spears, W.M.} (Eds.), \bibinfo{booktitle}{Swarm Robotics}.
	volume \bibinfo{volume}{3342} of \textit{\bibinfo{series}{Lecture Notes in
			Computer Science}}, pp. \bibinfo{pages}{10--20}.
	\newblock \bibinfo{note}{Springer, Berlin, Heidelberg.
		\url{https://doi.org/10.1007/978-3-540-30552-1_2}}.
	\bibitem[{Dahouda and Joe(2021)}]{Dahouda_2021}
	\bibinfo{author}{Dahouda, M.K.}, \bibinfo{author}{Joe, I.},
	\bibinfo{year}{2021}.
	\newblock \bibinfo{title}{{A} {D}eep-{L}earned {E}mbedding {T}echnique for
		{C}ategorical {F}eatures {E}ncoding}.
	\newblock \bibinfo{journal}{IEEE Access} \bibinfo{volume}{9},
	\bibinfo{pages}{114381--114391}.
	\newblock \bibinfo{note}{\url{https://doi.org/10.1109/ACCESS.2021.3104357}}.
	\bibitem[{Dufourq and Bassett(2017)}]{Dufourq_2017}
	\bibinfo{author}{Dufourq, E.}, \bibinfo{author}{Bassett, B.A.},
	\bibinfo{year}{2017}.
	\newblock \bibinfo{title}{{A}utomated {P}roblem {I}dentification: {R}egression
		vs {C}lassification via {E}volutionary {D}eep {N}etworks}.
	\newblock \bibinfo{note}{Preprint at
		\url{https://doi.org/10.48550/arXiv.1707.00703}}.
	\bibitem[{Fan et~al.(2018)Fan, Theodorou and Reeder}]{Fan_2018}
	\bibinfo{author}{Fan, D.D.}, \bibinfo{author}{Theodorou, E.A.},
	\bibinfo{author}{Reeder, J.}, \bibinfo{year}{2018}.
	\newblock \bibinfo{title}{{M}odel-{B}ased {S}tochastic {S}earch for {L}arge
		{S}cale {O}ptimization of {M}ulti-{A}gent {UAV} {S}warms}, in:
	\bibinfo{booktitle}{2018 IEEE Symposium Series on Computational Intelligence
		(SSCI)}, pp. \bibinfo{pages}{2216--2222}.
	\newblock \bibinfo{note}{IEEE, Bangalore, India.
		\url{https://doi.org/10.1109/SSCI.2018.8628677}}.
	\bibitem[{Feuerriegel et~al.(2023)Feuerriegel, Hartmann, Janiesch and
		Zschech}]{Feuerriegel_2023}
	\bibinfo{author}{Feuerriegel, S.}, \bibinfo{author}{Hartmann, J.},
	\bibinfo{author}{Janiesch, C.}, \bibinfo{author}{Zschech, P.},
	\bibinfo{year}{2023}.
	\newblock \bibinfo{title}{{G}enerative {AI}}.
	\newblock \bibinfo{journal}{Business \& Information Systems Engineering}
	\bibinfo{volume}{66}, \bibinfo{pages}{111--126}.
	\newblock \bibinfo{note}{\url{https://doi.org/10.1007/s12599-023-00834-7}}.
	\bibitem[{Francesca et~al.(2014)Francesca, Brambilla, Brutschy, Trianni and
		Birattari}]{Francesca_2014}
	\bibinfo{author}{Francesca, G.}, \bibinfo{author}{Brambilla, M.},
	\bibinfo{author}{Brutschy, A.}, \bibinfo{author}{Trianni, V.},
	\bibinfo{author}{Birattari, M.}, \bibinfo{year}{2014}.
	\newblock \bibinfo{title}{{A}uto{M}o{D}e: A novel approach to the automatic
		design of control software for robot swarms}.
	\newblock \bibinfo{journal}{Swarm Intelligence} \bibinfo{volume}{8},
	\bibinfo{pages}{89--112}.
	\newblock \bibinfo{note}{\url{https://doi.org/10.1007/s11721-014-0092-4}}.
	\bibitem[{Glasmachers et~al.(2010)Glasmachers, Schaul, Sun, Wierstra and
		Schmidhuber}]{Glasmachers_2010}
	\bibinfo{author}{Glasmachers, T.}, \bibinfo{author}{Schaul, T.},
	\bibinfo{author}{Sun, Y.}, \bibinfo{author}{Wierstra, D.},
	\bibinfo{author}{Schmidhuber, J.}, \bibinfo{year}{2010}.
	\newblock \bibinfo{title}{Exponential natural evolution strategies}, in:
	\bibinfo{booktitle}{2010 12th Annual Conference on Genetic and Evolutionary
		Computation (GECCO)}, pp. \bibinfo{pages}{393--400}.
	\newblock \bibinfo{note}{ACM, Portland, United States.
		\url{https://doi.org/10.1145/1830483.1830557}}.
	\bibitem[{Gupta et~al.(2017)Gupta, Egorov and Kochenderfer}]{Gupta_2017}
	\bibinfo{author}{Gupta, J.K.}, \bibinfo{author}{Egorov, M.},
	\bibinfo{author}{Kochenderfer, M.}, \bibinfo{year}{2017}.
	\newblock \bibinfo{title}{{C}ooperative {M}ulti-agent {C}ontrol {U}sing {D}eep
		{R}einforcement {L}earning}, in: \bibinfo{editor}{Sukthankar, G.},
	\bibinfo{editor}{Rodriguez-Aguilar, J.A.} (Eds.),
	\bibinfo{booktitle}{Autonomous Agents and Multiagent Systems}. volume
	\bibinfo{volume}{10642} of \textit{\bibinfo{series}{Lecture Notes in Computer
			Science}}, pp. \bibinfo{pages}{66--83}.
	\newblock \bibinfo{note}{Springer, Cham.
		\url{https://doi.org/10.1007/978-3-319-71682-4_5}}.
	\bibitem[{G\"{u}zel(2013)}]{Guzel_2013}
	\bibinfo{author}{G\"{u}zel, M.S.}, \bibinfo{year}{2013}.
	\newblock \bibinfo{title}{{A}utonomous {V}ehicle {N}avigation {U}sing {V}ision
		and {M}apless {S}trategies: {A} {S}urvey}.
	\newblock \bibinfo{journal}{Advances in Mechanical Engineering}
	\bibinfo{volume}{5}, \bibinfo{pages}{234747}.
	\newblock \bibinfo{note}{\url{https://doi.org/10.1155/2013/234747}}.
	\bibitem[{Ha(2017)}]{Ha_2017}
	\bibinfo{author}{Ha, D.}, \bibinfo{year}{2017}.
	\newblock \bibinfo{title}{{E}volving {S}table {S}trategies}.
	\newblock \bibinfo{note}{Blog.otoro.net.
		\url{https://blog.otoro.net/2017/11/12/evolving-stable-strategies/} (accessed
		12 October 2025)}.
	\bibitem[{Ha and Schmidhuber(2018)}]{Ha_2018}
	\bibinfo{author}{Ha, D.}, \bibinfo{author}{Schmidhuber, J.},
	\bibinfo{year}{2018}.
	\newblock \bibinfo{title}{World models}.
	\newblock \bibinfo{note}{Preprint at
		\url{https://doi.org/10.48550/arXiv.1803.10122}}.
	\bibitem[{Ha and Tang(2022)}]{Ha_2022}
	\bibinfo{author}{Ha, D.}, \bibinfo{author}{Tang, Y.}, \bibinfo{year}{2022}.
	\newblock \bibinfo{title}{Collective intelligence for deep learning: {A} survey
		of recent developments}.
	\newblock \bibinfo{journal}{Collective Intelligence} \bibinfo{volume}{1},
	\bibinfo{pages}{26339137221114874}.
	\newblock \bibinfo{note}{\url{https://doi.org/10.1177/26339137221114874}}.
	\bibitem[{Hansen and Ostermeier(2001)}]{Hansen_2001}
	\bibinfo{author}{Hansen, N.}, \bibinfo{author}{Ostermeier, A.},
	\bibinfo{year}{2001}.
	\newblock \bibinfo{title}{{C}ompletely {D}erandomized {S}elf-{A}daptation in
		{E}volution {S}trategies}.
	\newblock \bibinfo{journal}{Evolutionary Computation} \bibinfo{volume}{9},
	\bibinfo{pages}{159--195}.
	\newblock \bibinfo{note}{\url{https://doi.org/10.1162/106365601750190398}}.
	\bibitem[{Hasselmann et~al.(2021)Hasselmann, Ligot, Ruddick and
		Birattari}]{Hasselmann_2021}
	\bibinfo{author}{Hasselmann, K.}, \bibinfo{author}{Ligot, A.},
	\bibinfo{author}{Ruddick, J.}, \bibinfo{author}{Birattari, M.},
	\bibinfo{year}{2021}.
	\newblock \bibinfo{title}{Empirical assessment and comparison of
		neuro-evolutionary methods for the automatic off-line design of robot
		swarms}.
	\newblock \bibinfo{journal}{Nature Communications} \bibinfo{volume}{12},
	\bibinfo{pages}{4345}.
	\newblock \bibinfo{note}{\url{https://doi.org/10.1038/s41467-021-24642-3}}.
	\bibitem[{Hausler et~al.(2016)Hausler, Saccon, Aguiar, Hauser and
		Pascoal}]{Hausler_2016}
	\bibinfo{author}{Hausler, A.J.}, \bibinfo{author}{Saccon, A.},
	\bibinfo{author}{Aguiar, A.P.}, \bibinfo{author}{Hauser, J.},
	\bibinfo{author}{Pascoal, A.M.}, \bibinfo{year}{2016}.
	\newblock \bibinfo{title}{{E}nergy-{O}ptimal {M}otion {P}lanning for {M}ultiple
		{R}obotic {V}ehicles {W}ith {C}ollision {A}voidance}.
	\newblock \bibinfo{journal}{IEEE Transactions on Control Systems Technology}
	\bibinfo{volume}{24}, \bibinfo{pages}{867--883}.
	\newblock \bibinfo{note}{\url{https://doi.org/10.1109/TCST.2015.2475399}}.
	\bibitem[{Hazem et~al.(2025a)Hazem, Guler and Altaif}]{Hazem_2025a}
	\bibinfo{author}{Hazem, Z.B.}, \bibinfo{author}{Guler, N.},
	\bibinfo{author}{Altaif, A.H.}, \bibinfo{year}{2025}a.
	\newblock \bibinfo{title}{Model-free trajectory tracking control of a 5-{DOF}
		{M}itsubishi robotic arm using deep deterministic policy gradient algorithm}.
	\newblock \bibinfo{journal}{Discover Robotics} \bibinfo{volume}{1},
	\bibinfo{pages}{4}.
	\newblock \bibinfo{note}{\url{https://doi.org/10.1007/s44430-025-00004-2}}.
	\bibitem[{Hazem et~al.(2025b)Hazem, Saidi, Guler and Altaif}]{Hazem_2025c}
	\bibinfo{author}{Hazem, Z.B.}, \bibinfo{author}{Saidi, F.},
	\bibinfo{author}{Guler, N.}, \bibinfo{author}{Altaif, A.H.},
	\bibinfo{year}{2025}b.
	\newblock \bibinfo{title}{{A} {H}ybrid {R}einforcement {L}earning {F}ramework
		{C}ombining {TD3} and {PID} {C}ontrol for {R}obust {T}rajectory {T}racking of
		a 5-{DOF} {R}obotic {A}rm}.
	\newblock \bibinfo{journal}{Automation} \bibinfo{volume}{6},
	\bibinfo{pages}{56}.
	\newblock \bibinfo{note}{\url{https://doi.org/10.3390/automation6040056}}.
	\bibitem[{Hazem et~al.(2025c)Hazem, Saidi, Guler and Altaif}]{Hazem_2025b}
	\bibinfo{author}{Hazem, Z.B.}, \bibinfo{author}{Saidi, F.},
	\bibinfo{author}{Guler, N.}, \bibinfo{author}{Altaif, A.H.},
	\bibinfo{year}{2025}c.
	\newblock \bibinfo{title}{Reinforcement learning-based intelligent trajectory
		tracking for a 5-{DOF} {M}itsubishi robotic arm: comparative evaluation of
		{DDPG}, {LC}-{DDPG}, and {TD3}-{ADX}}.
	\newblock \bibinfo{journal}{International Journal of Intelligent Robotics and
		Applications} \bibinfo{volume}{9}, \bibinfo{pages}{1982--2002}.
	\newblock \bibinfo{note}{\url{https://doi.org/10.1007/s41315-025-00475-x}}.
	\bibitem[{Hiraga(2026)}]{Hiraga_2026}
	\bibinfo{author}{Hiraga, M.}, \bibinfo{year}{2026}.
	\newblock \bibinfo{title}{Topology and weight neuroevolution for social
		learning in robotic swarms}.
	\newblock \bibinfo{journal}{Artificial Life and Robotics} \bibinfo{volume}{31},
	\bibinfo{pages}{202--212}.
	\newblock \bibinfo{note}{\url{https://doi.org/10.1007/s10015-025-01078-2}}.
	\bibitem[{H\"{u}ttenrauch et~al.(2019)H\"{u}ttenrauch, \v{S}o\v{s}i\'c and
		Neumann}]{Huttenrauch_2019}
	\bibinfo{author}{H\"{u}ttenrauch, M.}, \bibinfo{author}{\v{S}o\v{s}i\'c, A.},
	\bibinfo{author}{Neumann, G.}, \bibinfo{year}{2019}.
	\newblock \bibinfo{title}{{D}eep {R}einforcement {L}earning for {S}warm
		{S}ystems}.
	\newblock \bibinfo{journal}{Journal of Machine Learning Research}
	\bibinfo{volume}{20}, \bibinfo{pages}{1--31}.
	\newblock
	\bibinfo{note}{\url{https://www.jmlr.org/papers/volume20/18-476/18-476.pdf}}.
	\bibitem[{Kaelbling et~al.(1996)Kaelbling, Littman and Moore}]{Kaelbling_1996}
	\bibinfo{author}{Kaelbling, L.P.}, \bibinfo{author}{Littman, M.L.},
	\bibinfo{author}{Moore, A.W.}, \bibinfo{year}{1996}.
	\newblock \bibinfo{title}{{R}einforcement {L}earning: {A} {S}urvey}.
	\newblock \bibinfo{journal}{Journal of Artificial Intelligence Research}
	\bibinfo{volume}{4}, \bibinfo{pages}{237--285}.
	\newblock \bibinfo{note}{\url{https://doi.org/10.1613/jair.301}}.
	\bibitem[{Kingma and Welling(2013)}]{Kingma_2013}
	\bibinfo{author}{Kingma, D.P.}, \bibinfo{author}{Welling, M.},
	\bibinfo{year}{2013}.
	\newblock \bibinfo{title}{{A}uto-{E}ncoding {V}ariational {B}ayes}.
	\newblock \bibinfo{note}{Preprint at
		\url{https://doi.org/10.48550/arXiv.1312.6114}}.
	\bibitem[{Kolar et~al.(2020)Kolar, Benavidez and Jamshidi}]{Kolar_2020}
	\bibinfo{author}{Kolar, P.}, \bibinfo{author}{Benavidez, P.},
	\bibinfo{author}{Jamshidi, M.}, \bibinfo{year}{2020}.
	\newblock \bibinfo{title}{{S}urvey of {D}atafusion {T}echniques for {L}aser and
		{V}ision {B}ased {S}ensor {I}ntegration for {A}utonomous {N}avigation}.
	\newblock \bibinfo{journal}{Sensors} \bibinfo{volume}{20},
	\bibinfo{pages}{2180}.
	\newblock \bibinfo{note}{\url{https://doi.org/10.3390/s20082180}}.
	\bibitem[{Kuckling(2023)}]{Kuckling_2023}
	\bibinfo{author}{Kuckling, J.}, \bibinfo{year}{2023}.
	\newblock \bibinfo{title}{Recent trends in robot learning and evolution for
		swarm robotics}.
	\newblock \bibinfo{journal}{Frontiers in Robotics and AI} \bibinfo{volume}{10},
	\bibinfo{pages}{1134841}.
	\newblock \bibinfo{note}{\url{https://doi.org/10.3389/frobt.2023.1134841}}.
	\bibitem[{Lawson and Qureshi(2023)}]{Lawson_2023}
	\bibinfo{author}{Lawson, D.}, \bibinfo{author}{Qureshi, A.H.},
	\bibinfo{year}{2023}.
	\newblock \bibinfo{title}{{C}ontrol {T}ransformer: {R}obot {N}avigation in
		{U}nknown {E}nvironments {T}hrough {PRM}-{G}uided {R}eturn-{C}onditioned
		{S}equence {M}odeling}, in: \bibinfo{booktitle}{2023 IEEE/RSJ International
		Conference on Intelligent Robots and Systems (IROS)}, pp.
	\bibinfo{pages}{9324--9331}.
	\newblock \bibinfo{note}{IEEE, Detroit, United States.
		\url{https://doi.org/10.1109/IROS55552.2023.10341628}}.
	\bibitem[{LeCun et~al.(2015)LeCun, Bengio and Hinton}]{LeCun_2015}
	\bibinfo{author}{LeCun, Y.}, \bibinfo{author}{Bengio, Y.},
	\bibinfo{author}{Hinton, G.}, \bibinfo{year}{2015}.
	\newblock \bibinfo{title}{Deep learning}.
	\newblock \bibinfo{journal}{Nature} \bibinfo{volume}{521},
	\bibinfo{pages}{436--444}.
	\newblock \bibinfo{note}{\url{https://doi.org/10.1038/nature14539}}.
	\bibitem[{Liu et~al.(2025a)Liu, Long, Li, Yan, Li, Chen, Gu, Pu and
		Luo}]{LiuH_2025}
	\bibinfo{author}{Liu, H.}, \bibinfo{author}{Long, X.}, \bibinfo{author}{Li,
		Y.}, \bibinfo{author}{Yan, J.}, \bibinfo{author}{Li, M.},
	\bibinfo{author}{Chen, C.}, \bibinfo{author}{Gu, F.}, \bibinfo{author}{Pu,
		H.}, \bibinfo{author}{Luo, J.}, \bibinfo{year}{2025}a.
	\newblock \bibinfo{title}{Adaptive multi-{UAV} cooperative path planning based
		on novel rotation artificial potential fields}.
	\newblock \bibinfo{journal}{Knowledge-Based Systems} \bibinfo{volume}{317},
	\bibinfo{pages}{113429}.
	\newblock \bibinfo{note}{\url{https://doi.org/10.1016/j.knosys.2025.113429}}.
	\bibitem[{Liu et~al.(2026)Liu, Wen, Guo and Liu}]{Liu_2026}
	\bibinfo{author}{Liu, X.}, \bibinfo{author}{Wen, S.}, \bibinfo{author}{Guo,
		Z.}, \bibinfo{author}{Liu, H.}, \bibinfo{year}{2026}.
	\newblock \bibinfo{title}{Transfer learning-based sparse-reward
		meta-{Q}-learning algorithm for active {SLAM}}.
	\newblock \bibinfo{journal}{Expert Systems with Applications}
	\bibinfo{volume}{299}, \bibinfo{pages}{129799}.
	\newblock \bibinfo{note}{\url{https://doi.org/10.1016/j.eswa.2025.129799}}.
	\bibitem[{Liu et~al.(2025b)Liu, Wen, Liu and Yu}]{LiuX_2025}
	\bibinfo{author}{Liu, X.}, \bibinfo{author}{Wen, S.}, \bibinfo{author}{Liu,
		H.}, \bibinfo{author}{Yu, F.R.}, \bibinfo{year}{2025}b.
	\newblock \bibinfo{title}{{CPL}-{SLAM}: {C}entralized {C}ollaborative
		{M}ultirobot {V}isual-{I}nertial {SLAM} {U}sing {P}oint-and-{L}ine
		{F}eatures}.
	\newblock \bibinfo{journal}{IEEE Internet of Things Journal}
	\bibinfo{volume}{12}, \bibinfo{pages}{21866--21875}.
	\newblock \bibinfo{note}{\url{https://doi.org/10.1109/JIOT.2025.3548958}}.
	\bibitem[{Meng et~al.(2023)Meng, Wen, Le, Li, Xing, Zhang, Wen, Zhang, Wang,
		Yang and Xu}]{Meng_2023}
	\bibinfo{author}{Meng, L.}, \bibinfo{author}{Wen, M.}, \bibinfo{author}{Le,
		C.}, \bibinfo{author}{Li, X.}, \bibinfo{author}{Xing, D.},
	\bibinfo{author}{Zhang, W.}, \bibinfo{author}{Wen, Y.},
	\bibinfo{author}{Zhang, H.}, \bibinfo{author}{Wang, J.},
	\bibinfo{author}{Yang, Y.}, \bibinfo{author}{Xu, B.}, \bibinfo{year}{2023}.
	\newblock \bibinfo{title}{{O}ffline {P}re-trained {M}ulti-agent {D}ecision
		{T}ransformer}.
	\newblock \bibinfo{journal}{Machine Intelligence Research}
	\bibinfo{volume}{20}, \bibinfo{pages}{233--248}.
	\newblock \bibinfo{note}{\url{https://doi.org/10.1007/s11633-022-1383-7}}.
	\bibitem[{Moradi et~al.(2019)Moradi, Berangi and Minaei}]{Moradi_2019}
	\bibinfo{author}{Moradi, R.}, \bibinfo{author}{Berangi, R.},
	\bibinfo{author}{Minaei, B.}, \bibinfo{year}{2019}.
	\newblock \bibinfo{title}{A survey of regularization strategies for deep
		models}.
	\newblock \bibinfo{journal}{Artificial Intelligence Review}
	\bibinfo{volume}{53}, \bibinfo{pages}{3947--3986}.
	\newblock \bibinfo{note}{\url{https://doi.org/10.1007/s10462-019-09784-7}}.
	\bibitem[{Mukhlish et~al.(2018)Mukhlish, Page and Bain}]{Mukhlish_2018}
	\bibinfo{author}{Mukhlish, F.}, \bibinfo{author}{Page, J.},
	\bibinfo{author}{Bain, M.}, \bibinfo{year}{2018}.
	\newblock \bibinfo{title}{Evolutionary-learning framework: improving automatic
		swarm robotics design}.
	\newblock \bibinfo{journal}{International Journal of Intelligent Unmanned
		Systems} \bibinfo{volume}{6}, \bibinfo{pages}{197--215}.
	\newblock \bibinfo{note}{\url{https://doi.org/10.1108/IJIUS-06-2018-0016}}.
	\bibitem[{Muratore et~al.(2022)Muratore, Ramos, Turk, Yu, Gienger and
		Peters}]{Muratore_2022}
	\bibinfo{author}{Muratore, F.}, \bibinfo{author}{Ramos, F.},
	\bibinfo{author}{Turk, G.}, \bibinfo{author}{Yu, W.},
	\bibinfo{author}{Gienger, M.}, \bibinfo{author}{Peters, J.},
	\bibinfo{year}{2022}.
	\newblock \bibinfo{title}{{R}obot {L}earning {F}rom {R}andomized {S}imulations:
		{A} {R}eview}.
	\newblock \bibinfo{journal}{Frontiers in Robotics and AI} \bibinfo{volume}{9},
	\bibinfo{pages}{799893}.
	\newblock \bibinfo{note}{\url{https://doi.org/10.3389/frobt.2022.799893}}.
	\bibitem[{Nanda et~al.(2021)Nanda, Mohapatra, Mahapatra, Mahapatra and
		Mahapatra}]{Nanda_2021}
	\bibinfo{author}{Nanda, A.}, \bibinfo{author}{Mohapatra, D.B.B.},
	\bibinfo{author}{Mahapatra, A.P.K.}, \bibinfo{author}{Mahapatra, A.P.K.},
	\bibinfo{author}{Mahapatra, A.P.K.}, \bibinfo{year}{2021}.
	\newblock \bibinfo{title}{Multiple comparison test by {T}ukey's honestly
		significant difference ({HSD}): {D}o the confident level control type {I}
		error}.
	\newblock \bibinfo{journal}{International Journal of Statistics and Applied
		Mathematics} \bibinfo{volume}{6}, \bibinfo{pages}{59--65}.
	\newblock \bibinfo{note}{\url{https://doi.org/10.22271/maths.2021.v6.i1a.636}}.
	\bibitem[{Nguyen et~al.(2021)Nguyen, Tran and Luong}]{Nguyen_2021}
	\bibinfo{author}{Nguyen, H.T.}, \bibinfo{author}{Tran, K.},
	\bibinfo{author}{Luong, N.H.}, \bibinfo{year}{2021}.
	\newblock \bibinfo{title}{{E}nhancing {T}win {D}elayed {D}eep {D}eterministic
		{P}olicy {G}radient with {C}ross-{E}ntropy {M}ethod}, in:
	\bibinfo{booktitle}{2021 8th NAFOSTED Conference on Information and Computer
		Science (NICS)}, pp. \bibinfo{pages}{132--137}.
	\newblock \bibinfo{note}{IEEE, Hanoi, Vietnam.
		\url{https://doi.org/10.1109/NICS54270.2021.9701549}}.
	\bibitem[{Pierson and Gashler(2017)}]{Pierson_2017}
	\bibinfo{author}{Pierson, H.A.}, \bibinfo{author}{Gashler, M.S.},
	\bibinfo{year}{2017}.
	\newblock \bibinfo{title}{Deep learning in robotics: a review of recent
		research}.
	\newblock \bibinfo{journal}{Advanced Robotics} \bibinfo{volume}{31},
	\bibinfo{pages}{821--835}.
	\newblock \bibinfo{note}{\url{https://doi.org/10.1080/01691864.2017.1365009}}.
	\bibitem[{Rubinstein(1999)}]{Rubinstein_1999}
	\bibinfo{author}{Rubinstein, R.}, \bibinfo{year}{1999}.
	\newblock \bibinfo{title}{{T}he {C}ross-{E}ntropy {M}ethod for {C}ombinatorial
		and {C}ontinuous {O}ptimization}.
	\newblock \bibinfo{journal}{Methodology And Computing In Applied Probability}
	\bibinfo{volume}{1}, \bibinfo{pages}{127--190}.
	\newblock \bibinfo{note}{\url{https://doi.org/10.1023/A:1010091220143}}.
	\bibitem[{Salimans et~al.(2017)Salimans, Ho, Chen, Sidor and
		Sutskever}]{Salimans_2017}
	\bibinfo{author}{Salimans, T.}, \bibinfo{author}{Ho, J.},
	\bibinfo{author}{Chen, X.}, \bibinfo{author}{Sidor, S.},
	\bibinfo{author}{Sutskever, I.}, \bibinfo{year}{2017}.
	\newblock \bibinfo{title}{{E}volution {S}trategies as a {S}calable
		{A}lternative to {R}einforcement {L}earning}.
	\newblock \bibinfo{note}{Preprint at
		\url{https://doi.org/10.48550/arXiv.1703.03864}}.
	\bibitem[{Schulman et~al.(2015)Schulman, Levine, Abbeel, Jordan and
		Moritz}]{Schulman_2015}
	\bibinfo{author}{Schulman, J.}, \bibinfo{author}{Levine, S.},
	\bibinfo{author}{Abbeel, P.}, \bibinfo{author}{Jordan, M.},
	\bibinfo{author}{Moritz, P.}, \bibinfo{year}{2015}.
	\newblock \bibinfo{title}{{T}rust {R}egion {P}olicy {O}ptimization}, in:
	\bibinfo{booktitle}{2015 32nd International Conference on Machine Learning
		(ICML)}, pp. \bibinfo{pages}{1889--1897}.
	\newblock \bibinfo{note}{PMLR, Lille, France.
		\url{http://proceedings.mlr.press/v37/schulman15.pdf}}.
	\bibitem[{Schulman et~al.(2017)Schulman, Wolski, Dhariwal, Radford and
		Klimov}]{Schulman_2017}
	\bibinfo{author}{Schulman, J.}, \bibinfo{author}{Wolski, F.},
	\bibinfo{author}{Dhariwal, P.}, \bibinfo{author}{Radford, A.},
	\bibinfo{author}{Klimov, O.}, \bibinfo{year}{2017}.
	\newblock \bibinfo{title}{{P}roximal {P}olicy {O}ptimization {A}lgorithms}.
	\newblock \bibinfo{note}{Preprint at
		\url{https://doi.org/10.48550/arXiv.1707.06347}}.
	\bibitem[{Shaikh and Ra\~{n}\'{o}(2020)}]{Shaikh_2020}
	\bibinfo{author}{Shaikh, D.}, \bibinfo{author}{Ra\~{n}\'{o}, I.},
	\bibinfo{year}{2020}.
	\newblock \bibinfo{title}{{B}raitenberg {V}ehicles as {C}omputational {T}ools
		for {R}esearch in {N}euroscience}.
	\newblock \bibinfo{journal}{Frontiers in Bioengineering and Biotechnology}
	\bibinfo{volume}{8}, \bibinfo{pages}{565963}.
	\newblock \bibinfo{note}{\url{https://doi.org/10.3389/fbioe.2020.565963}}.
	\bibitem[{Shanthamallu et~al.(2017)Shanthamallu, Spanias, Tepedelenlioglu and
		Stanley}]{Shanthamallu_2017}
	\bibinfo{author}{Shanthamallu, U.S.}, \bibinfo{author}{Spanias, A.},
	\bibinfo{author}{Tepedelenlioglu, C.}, \bibinfo{author}{Stanley, M.},
	\bibinfo{year}{2017}.
	\newblock \bibinfo{title}{A brief survey of machine learning methods and their
		sensor and {I}o{T} applications}, in: \bibinfo{booktitle}{2017 8th
		International Conference on Information, Intelligence, Systems \&
		Applications (IISA)}, pp. \bibinfo{pages}{1--8}.
	\newblock \bibinfo{note}{IEEE, Larnaca, Cyprus.
		\url{https://doi.org/10.1109/IISA.2017.8316459}}.
	\bibitem[{Stanley and Miikkulainen(2002)}]{Stanley_2002}
	\bibinfo{author}{Stanley, K.O.}, \bibinfo{author}{Miikkulainen, R.},
	\bibinfo{year}{2002}.
	\newblock \bibinfo{title}{{E}volving {N}eural {N}etworks through {A}ugmenting
		{T}opologies}.
	\newblock \bibinfo{journal}{Evolutionary Computation} \bibinfo{volume}{10},
	\bibinfo{pages}{99--127}.
	\newblock \bibinfo{note}{\url{https://doi.org/10.1162/106365602320169811}}.
	\bibitem[{Stefek et~al.(2020)Stefek, Pham, Krivanek and Pham}]{Stefek_2020}
	\bibinfo{author}{Stefek, A.}, \bibinfo{author}{Pham, T.V.},
	\bibinfo{author}{Krivanek, V.}, \bibinfo{author}{Pham, K.L.},
	\bibinfo{year}{2020}.
	\newblock \bibinfo{title}{{E}nergy {C}omparison of {C}ontrollers {U}sed for a
		{D}ifferential {D}rive {W}heeled {M}obile {R}obot}.
	\newblock \bibinfo{journal}{IEEE Access} \bibinfo{volume}{8},
	\bibinfo{pages}{170915--170927}.
	\newblock \bibinfo{note}{\url{https://doi.org/10.1109/ACCESS.2020.3023345}}.
	\bibitem[{Wawrzy\'{n}ski(2009)}]{Wawrzynski_2009}
	\bibinfo{author}{Wawrzy\'{n}ski, P.}, \bibinfo{year}{2009}.
	\newblock \bibinfo{title}{{A} {C}at-{L}ike {R}obot {R}eal-{T}ime {L}earning to
		{R}un}, in: \bibinfo{editor}{Kolehmainen, M.}, \bibinfo{editor}{Toivanen,
		P.}, \bibinfo{editor}{Beliczynski, B.} (Eds.), \bibinfo{booktitle}{Adaptive
		and Natural Computing Algorithms}. volume \bibinfo{volume}{5495} of
	\textit{\bibinfo{series}{Lecture Notes in Computer Science}}, pp.
	\bibinfo{pages}{380--390}.
	\newblock \bibinfo{note}{Springer, Berlin, Heidelberg.
		\url{https://doi.org/10.1007/978-3-642-04921-7_39}}.
	\bibitem[{Wierstra et~al.(2014)Wierstra, Schaul, Glasmachers, Sun, Peters and
		Schmidhuber}]{Wierstra_2014}
	\bibinfo{author}{Wierstra, D.}, \bibinfo{author}{Schaul, T.},
	\bibinfo{author}{Glasmachers, T.}, \bibinfo{author}{Sun, Y.},
	\bibinfo{author}{Peters, J.}, \bibinfo{author}{Schmidhuber, J.},
	\bibinfo{year}{2014}.
	\newblock \bibinfo{title}{{N}atural {E}volution {S}trategies}.
	\newblock \bibinfo{journal}{Journal of Machine Learning Research}
	\bibinfo{volume}{15}, \bibinfo{pages}{949--980}.
	\newblock
	\bibinfo{note}{\url{https://www.jmlr.org/papers/volume15/wierstra14a/wierstra14a.pdf}}.
	\bibitem[{Xiao et~al.(2022)Xiao, Liu, Warnell and Stone}]{Xiao_2022}
	\bibinfo{author}{Xiao, X.}, \bibinfo{author}{Liu, B.},
	\bibinfo{author}{Warnell, G.}, \bibinfo{author}{Stone, P.},
	\bibinfo{year}{2022}.
	\newblock \bibinfo{title}{Motion planning and control for mobile robot
		navigation using machine learning: a survey}.
	\newblock \bibinfo{journal}{Autonomous Robots} \bibinfo{volume}{46},
	\bibinfo{pages}{569--597}.
	\newblock \bibinfo{note}{\url{https://doi.org/10.1007/s10514-022-10039-8}}.
	\bibitem[{Ye et~al.(2023)Ye, Pan, Gao, Wu and Ren}]{Ye_2023}
	\bibinfo{author}{Ye, X.}, \bibinfo{author}{Pan, Z.}, \bibinfo{author}{Gao, X.},
	\bibinfo{author}{Wu, K.}, \bibinfo{author}{Ren, B.}, \bibinfo{year}{2023}.
	\newblock \bibinfo{title}{{D}ifferentiable {L}earning of {S}calable
		{M}ulti-{A}gent {N}avigation {P}olicies}.
	\newblock \bibinfo{journal}{IEEE Robotics and Automation Letters}
	\bibinfo{volume}{8}, \bibinfo{pages}{2229--2236}.
	\newblock \bibinfo{note}{\url{https://doi.org/10.1109/LRA.2023.3248440}}.
	\bibitem[{Yogesh et~al.(2017)Yogesh, Dubey, Arora and Yadav}]{Yogesh_2017}
	\bibinfo{author}{Yogesh}, \bibinfo{author}{Dubey, A.K.},
	\bibinfo{author}{Arora, R.}, \bibinfo{author}{Yadav, S.},
	\bibinfo{year}{2017}.
	\newblock \bibinfo{title}{A review and comprehensive comparison of image
		de-noising techniques}, in: \bibinfo{booktitle}{2017 6th International
		Conference on Reliability, Infocom Technologies and Optimization (Trends and
		Future Directions) (ICRITO)}, pp. \bibinfo{pages}{624--629}.
	\newblock \bibinfo{note}{IEEE, Noida, India.
		\url{https://doi.org/10.1109/ICRITO.2017.8342503}}.
	\bibitem[{Yu et~al.(2015)Yu, Yasuda, Ohkura, Matsumura and Goka}]{Yu_2015}
	\bibinfo{author}{Yu, T.}, \bibinfo{author}{Yasuda, T.},
	\bibinfo{author}{Ohkura, K.}, \bibinfo{author}{Matsumura, Y.},
	\bibinfo{author}{Goka, M.}, \bibinfo{year}{2015}.
	\newblock \bibinfo{title}{{R}obust {S}warm {R}obotics {S}ystem using
		{CMA}-{N}euroes with {I}ncremental {E}volution}.
	\newblock \bibinfo{journal}{International Journal of Engineering Research \&
		Technology} \bibinfo{volume}{4}.
	\newblock
	\bibinfo{note}{\url{https://www.ijert.org/research/robust-swarm-robotics-system-using-cmaneuroes-with-incremental-evolution-IJERTV4IS110303.pdf}}.
	
\end{thebibliography}
\end{document}